\documentclass[accepted]{uai2023} 

\usepackage[american]{babel}

\usepackage{natbib} 
\usepackage{mathtools} 
\usepackage{booktabs} 
\usepackage{tikz} 
\usepackage{hyperref}
\usepackage{url}
\usepackage{booktabs} 
\usepackage{algorithm}
\usepackage{algorithmic} 
\usepackage{amssymb}
\usepackage{mathtools}
\usepackage{picinpar}
\usepackage{wrapfig}
\usepackage{enumitem}
\usepackage{subfigure}
\usepackage{multirow}
\usepackage{booktabs}
\usepackage{bbding}
\usepackage{subfigure}
\usepackage{graphicx}
\usepackage{xcolor}
\usepackage[most]{tcolorbox}
\usepackage{makecell}
\usepackage{caption}
\allowdisplaybreaks[4]

\title{Time-Conditioned Generative Modeling of Object-Centric Representations for Video Decomposition and Prediction}

\author[1]{Chengmin Gao}
\author[1]{Bin Li\thanks{Corresponding author (libin@fudan.edu.cn)}}

\affil[1]{%
    School of Computer Science, Fudan University
}

\begin{document}
\maketitle
\begin{abstract}
  When perceiving the world from multiple viewpoints, humans have the ability to reason about the complete objects in a compositional manner even when an object is completely occluded from certain viewpoints. Meanwhile, humans are able to imagine novel views after observing multiple viewpoints. Recent remarkable advances in multi-view object-centric learning still leaves some unresolved problems: 1) The shapes of partially or completely occluded objects can not be well reconstructed. 2) The novel viewpoint prediction depends on expensive viewpoint annotations rather than implicit rules in view representations. In this paper, we introduce a time-conditioned generative model for videos. To reconstruct the complete shape of an object accurately, we enhance the disentanglement between the latent representations of objects and views, where the latent representations of time-conditioned views are jointly inferred with a Transformer and then are input to a sequential extension of Slot Attention to learn object-centric representations. In addition, Gaussian processes are employed as priors of view latent variables for video generation and novel-view prediction without viewpoint annotations. Experiments on multiple datasets demonstrate that the proposed model can make object-centric video decomposition, reconstruct the complete shapes of occluded objects, and make novel-view predictions. 
\end{abstract}

\section{Introduction}
  Humans understand the multi-object world in a compositional manner that the representations of multiple objects are memorized separately and then combined into the perceived whole \cite[]{kahneman1992reviewing,spelke2007core,johnson2010infants}. When it comes to the multi-object scene with multiple viewpoints, humans exhibit higher-level intelligence in multiple aspects: On one hand, a certain object is endowed with a canonical representation that depicts its complete 3D shape and appearance through multi-view perception \cite[]{turnbull1997neuropsychology}. As a result, humans have the ability to reason about the complete object even when an object is completely occluded from certain viewpoints \cite[]{shepard1971mental}. On the other hand, scenes observed from novel viewpoints can be imagined on the basis of the learned implicit rules of perspective \cite[]{schacter2012future,beaty2016creative}. Such compositional modeling from multiple viewpoints is the fundamental ingredient for high-level cognitive intelligence. 

  Unsupervised object-centric learning that is dedicated to simulating human intelligence have recently achieved remarkable advances \cite[]{yuan2022compositional}, especially in single-view object-centric learning on both images \cite[]{burgess2018understanding,yuan2019generative, yuan2019spatial,engelcke2021genesis} and videos \cite[]{kosiorek2018sequential,jiang2019scalor,lin2020improving}. Meanwhile, multi-view object-centric learning \cite[]{li2020learning,chen2021roots,kabra2021simone,yuan2022unsupervised}, which aims to learn 3D object representations, also demonstrates a promising blueprint; however, it still leaves some unresolved problems: 1) The shapes of partially or completely occluded objects from some viewpoints cannot be reconstructed through 3D representations learned from other viewpoints. Although some models can theoretically restore occlusions, relatively poor restoration (e.g. inaccurate shadows, blurs and noises) is inevitably observed. 2) Despite using the query objective during training \cite[]{li2020learning}, the ability for novel viewpoint prediction depends on expensive viewpoint annotations, which provide strong location information and play a crucial role in update of object-centric representations; while the implicit rules of view representations are not fully explored to make prediction. 
  It is, therefore, crucial to develop a unified multi-view model to perform object-centric learning like humans. 

\begin{figure}[tb]
  \begin{center}
  \centering
  \includegraphics[width=\linewidth]{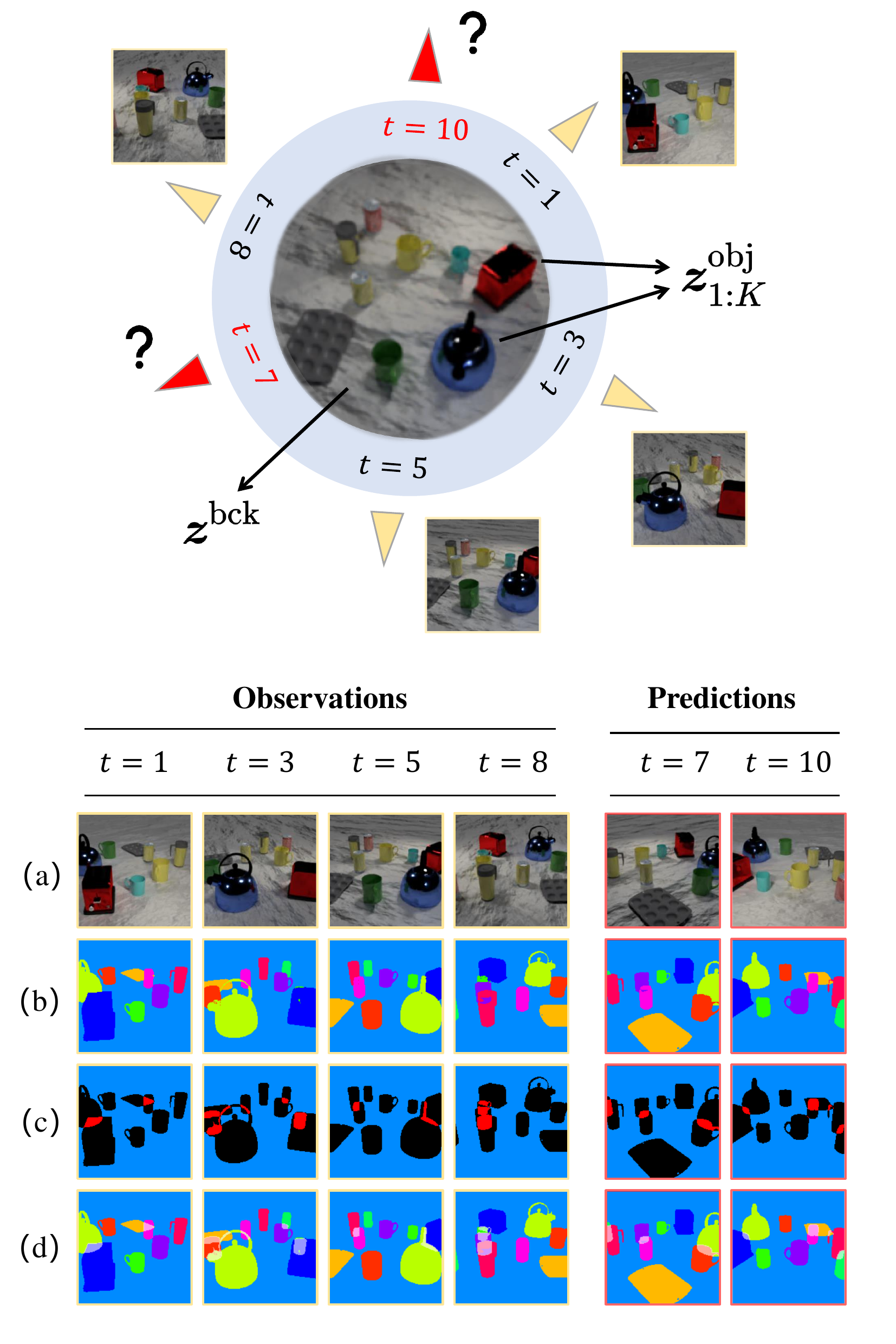}
  \caption{\textbf{Top}: Video decomposition and prediction with multiple observed time-conditioned viewpoints. The yellow and red triangles represent the observed frames and predicted frames, respectively. \textbf{Bottom}: The expected outputs: (a) reconstruction, (b) segmentation, (c) overlaps, and (d) complete segmentation. In our problem setting, only the observation set and time stamps are provided.}
  \label{fig:intro}
  \end{center}
\end{figure}

  In this paper, we focus on learning object-centric and viewpoint representations conditioned on time stamps from multi-view static scenes for video decomposition and unknown-viewpoint prediction. The problem setting and the expected outputs are illustrated in Figure \ref{fig:intro}. Under the setting that only the observation set and time stamps are provided, a generative model is developed to 1) make video decomposition based on object-centric representations; 2) reconstruct the complete shapes of partially or even completely occluded objects; and 3) predict 2D images from unknown viewpoints conditioned on known viewpoints.

  To enable the abovementioned abilities, we propose a time-conditioned generative model for video decomposition and prediction. The proposed model reconstructs the complete shape of an object accurately through enhancing the disentanglement between object-centric representations and viewpoint representations, where the latent representations of time-conditioned views are jointly inferred with a Transformer \cite[]{vaswani2017attention} and then are input to a sequential extension of Slot Attention \cite[]{locatello2020object} to learn viewpoint-invariant object-centric representations. In addition, the prediction from novel viewpoints without viewpoint annotations is enabled. Specifically, Gaussian processes are employed as priors of viewpoint latent variables for video generation and novel-view inference, based on the learned functions depicting the underlying implicit rules in view representations. 
  
  Experiments on multiple synthetic datasets demonstrate that the proposed model can 1) make object-centric video decomposition, 2) reconstruct the complete shapes of occluded objects, and 3) make novel-view predictions. Moreover, the proposed model outperforms the state-of-the-art methods in video decomposition and, compared with the method that uses viewpoint annotations, achieves competitive results on novel-view prediction.

\section{Related Work}

  \textbf{Single-View Object-Centric Learning.} Recent advances mainly focus on aggregating the input image into multiple slots based on the attention mechanism. AIR \cite[]{eslami2016attend} extracts a variable number of object representations based on the bounding-box attention \cite[]{jaderberg2015spatial}. SQAIR \cite[]{kosiorek2018sequential} further extends AIR to videos. Both SPACE \cite[]{lin2019space} and GMIOO \cite[]{yuan2019generative} model the background separately and model occlusions from different perspectives. SCALOR \cite[]{jiang2019scalor} implements object discovery and tracking in videos with dynamic backgrounds based on SPACE. G-SWM \cite[]{lin2020improving} integrates the advantages of current models on videos and further models the multimodal uncertainty. MONet \cite[]{burgess2019monet} adopts the attention network to iteratively infer masks and then extract object-centric representations based on masked features. GENESIS \cite[]{engelcke2020genesis} additionally models layouts of scenes based on MONet. GENESIS-V2 \cite[]{engelcke2021genesis} infers the attention masks inspired by instance coloring previously used in supervised instance segmentation. Slot Attention \cite[]{locatello2020object} and EfficientMORL \cite[]{emami2021efficient} randomly initialize the embeddings of objects in the slots to compute the similarities between the embeddings and local features. ADI \cite[]{yuan2021knowledge} proposes a continual learning strategy and makes pilot explorations in the acquisition and exploitation of knowledge.

  \textbf{Multi-View Object-Centric Learning.} We can coarsely categorize the recent advances in terms of viewpoint annotation. GQN \cite[]{eslami2018neural} uses viewpoint annotations to build single-object scenes. Based on novel-view annotations, single-object images from the given viewpoints can be generated. MulMON \cite[]{li2020learning} models the multi-object multi-view scenes according to viewpoint annotations. The double-level iterative inference is conducted to achieve both multi-object segmentation and prediction. ROOTS \cite[]{chen2021roots} divides the three-dimensional space into equal-spaced grids and discovers objects in different grids. ROOTS also considers occlusions and makes predictions with viewpoint annotations. SIMONe \cite[]{kabra2021simone} and OCLOC \cite[]{yuan2022unsupervised} are the most recent models without viewpoint annotations. They learn viewpoint representations and object-centric representations separately. The difference is that SIMONe learns representations from videos and can recompose representations to novel scenes, while OCLOC is capable of modeling scenes from unordered viewpoints.
    
  \textbf{Deep Learning with Stochastic Processes.} The Gaussian Process (GP) \cite[]{rasmussen2006gaussian} is a classical non-parametric model that regards the outputs of a function as a random variable of multivariate Gaussian distribution. The Neural Process (NP) \cite[]{garnelo2018neural,kim2019attentive} captures function stochasticity with a Gaussian distributed latent variable obtained from an inference network. To integrate stochastic processes into generative models, \cite[]{shi2021raven} employs GPs with deep kernels for Raven’s progressive matrices completion. CLAP-NP \cite[]{shi2022compositional} takes the first attempt in compositional law parsing with random functions based on NPs. In addition, a number of deep generative models \cite[]{deng2020modeling, norcliffe2021neural, song2021score} introduce ODEs or SDEs to learn diverse random functions on latent states.

\section{Background}
  In order to enable the abilities illustrated in Figure \ref{fig:intro}, in the following we list the treatments to consider in multi-view object-centric representation learning from videos without viewpoint annotations. 

  \textbf{Variable Number of Objects.} As the number of objects differs from one scene to another, it requires modeling and inference. A possible solution is to introduce a set of Bernoulli variables $\boldsymbol{z}^{\text{pres}} = \{ z_1^{\text{pres}}, ..., z_K^{\text{pres}}\}$ to model object presences in the $K$ slots for automatic counting, where $K$ denotes the maximum number of objects that may appear in a scene.

  \textbf{Separately Modeling of Background.} As foreground objects only occupy local regions while the background covers the entire image, the generation of 3D objects from multiple viewpoints tends to blur through a decoder shared with the background. We train two different decoders, a shared foreground object decoder and a separate background decoder. 

  \textbf{View-independent Object Representations.} We don't learn object representations from different viewpoints separately. As we can view representations of the same object inherently consistent independent of viewpoints, we consider $\{ \boldsymbol{z}^{\text{bck}}, \boldsymbol{z}_1^{\text{obj}},...,\boldsymbol{z}_K^{\text{obj}} \}$ as view-independent object-centric representations, learned from multiple observed viewpoints to represent viewpoint-invariant 3D objects.

  \textbf{Depth Estimation of Objects.} 
  We introduce a depth variable $o_{t,k} \in \big[0,1\big]$ of the $k$th object in the $t$th frame and its complete shape $\boldsymbol{s}_{t,k}^{\text{shp}} \in \big[0,1 \big]^{N}$ before being occluded in generative modeling. In this way, the pixels of an object with larger depth values will cover the pixels with smaller depth values. We can thus naturally obtain the observed shape of an possibly occluded object. It is worth noting that this treatment is also applicable to situations where an object is completely occluded.

  \textbf{Modeling of Viewpoints.} We explicitly learn the viewpoint representations according to modelling the correlations of viewpoints, instead of directly leveraging viewpoint annotations as previous works \cite[]{li2020learning,chen2021roots}. The view-correlation based modeling can also enable novel-view prediction given any time. To this end, we define $\boldsymbol{z}^{\text{view}} \in \mathbb{R}^{T \times D}$ and $\boldsymbol{\lambda} \in \mathbb{R}^{T\times D \times D_{\lambda}}$, where $T$ denotes the number of frames, $D$ denotes the dimensionality of viewpoint representations, and $\boldsymbol{z}^{\text{view}}$ follows the GPs w.r.t. $\boldsymbol{\lambda}$ that characterizes the position of the camera in different frames.

\section{Method}

  Our goal is to infer object-centric latent variables independent of viewpoints and correlated viewpoint latent variables dependent on time $t$. In the following, we introduce our time-conditioned generative model, the inference method and a two-stage training procedure to achieve the goal.

\begin{figure*}[tb]
  \centering
  \includegraphics[width=\textwidth]{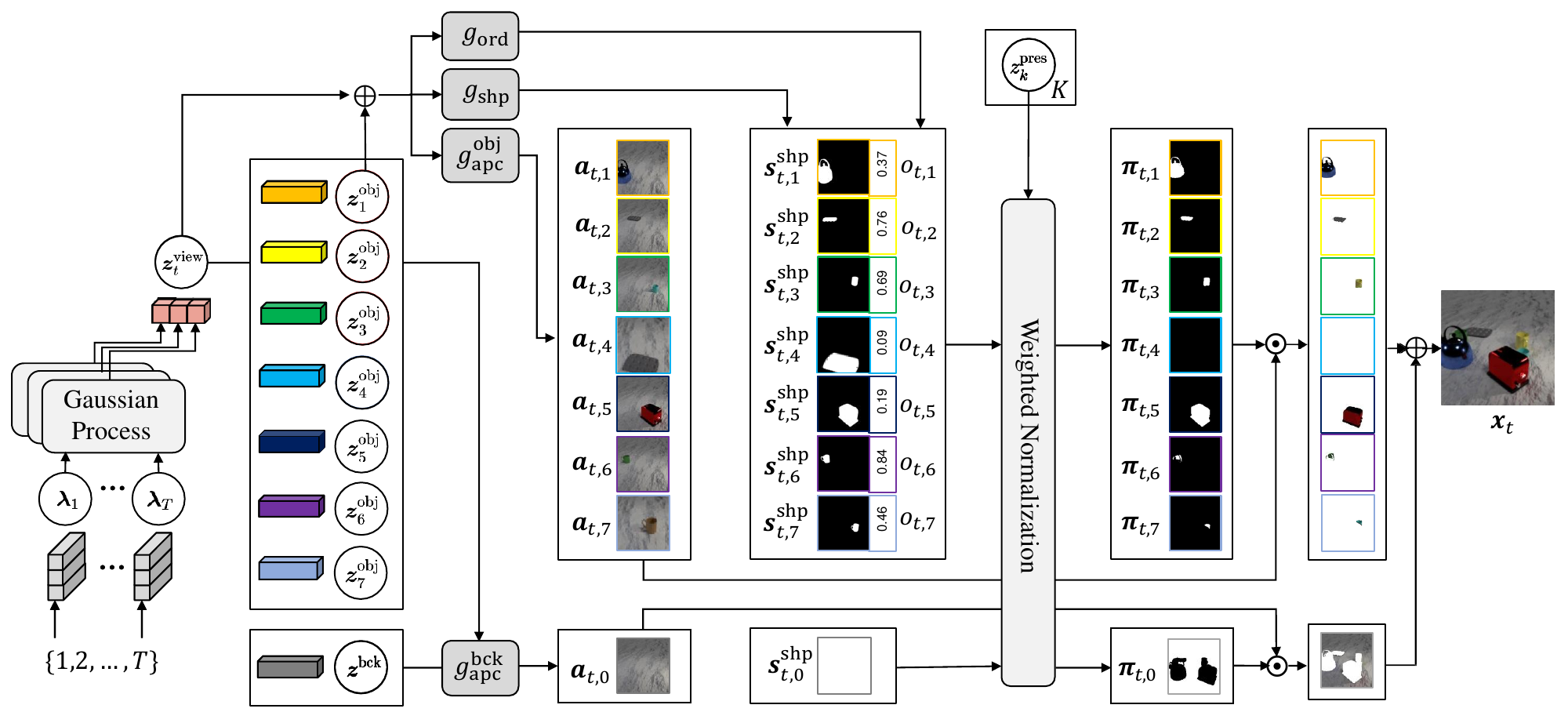}
  \caption{The proposed time-conditioned generative process for generating the $t$th frame in a video. The correlations between the viewpoint representations of $T$ frames are modeled dimension-wisely with GPs. The notations in circles denote latent variables; the notations in deep gray boxes denote neural networks.}
  \label{fig:generative}
\end{figure*}

\subsection{Generative Model}
\label{sec:generative_modeling}

  Let $\boldsymbol{x}_\mathcal{S} = \{\boldsymbol{x}_{1},...,\boldsymbol{x}_{T}\}$ be the $T$ frames in a video and $\boldsymbol{t}_{\mathcal{S}}$ be their timestamps. The frame set $\boldsymbol{x}_\mathcal{S}$ can be arbitrarily divided into an observation frame set $\boldsymbol{x}_{\mathcal{T}}$ and a prediction frame set $\boldsymbol{x}_{\mathcal{Q}}$, where $\boldsymbol{x}_{\mathcal{S}} = \boldsymbol{x}_{\mathcal{T}} \cup \boldsymbol{x}_{\mathcal{Q}}$.  For convenience, the elements in $\boldsymbol{x}_{\mathcal{T}} $ and $\boldsymbol{x}_{\mathcal{Q}}$ is sorted according to the time, e.g. $\boldsymbol{x}_{\mathcal{T}} = \big( \boldsymbol{x}_1 , \boldsymbol{x}_3, \boldsymbol{x}_7, \boldsymbol{x}_9 \big)$; similarly, $\boldsymbol{t}_{\mathcal{S}}$ can be divided into $\boldsymbol{t}_{\mathcal{T}}$ and $\boldsymbol{t}_{\mathcal{Q}}$ accordingly. Figure \ref{fig:generative} shows the flowchart of the generative process. The generative model conditioned on time $\boldsymbol{t}_{\mathcal{S}}$ can be expressed as:
\begin{align}
  \label{eq:lambda_prior}
  & \boldsymbol{\lambda}_{t,d}  \sim \mathcal{N}(\boldsymbol{A}\boldsymbol{w}_t,\sigma_{w}^2\boldsymbol{I}) \\
  &\kappa_{\boldsymbol{\eta}}^d (\boldsymbol{\lambda}_{t,d},\boldsymbol{\lambda}_{t',d}) = l^2 \exp \Big( \frac{\|g_{\boldsymbol{\eta}}^d (\boldsymbol{\lambda}_{t,d} )-g_{\boldsymbol{\eta}}^d (\boldsymbol{\lambda}_{t',d} )\|_2^2}{2\sigma^2} \Big) \\
  &\boldsymbol{z}_k^{\text{obj}} \sim \mathcal{N}(\boldsymbol{0},\boldsymbol{I}) \quad\quad \boldsymbol{z}^{\text{bck}}\sim \mathcal{N}(\boldsymbol{0},\boldsymbol{I}) \\
  &\boldsymbol{K}_{\boldsymbol{\eta}}^d = \left[\begin{array}{ccc}
  \kappa_{\boldsymbol{\eta}}^{d}\left(\boldsymbol{\lambda}_{1,d}, \boldsymbol{\lambda}_{1,d}\right) & \cdots & \kappa_{\boldsymbol{\eta}}^{d}\left(\boldsymbol{\lambda}_{1,d}, \boldsymbol{\lambda}_{T,d}\right) \\
  \vdots & \ddots & \vdots \\
  \kappa_{\boldsymbol{\eta}}^{d}\left(\boldsymbol{\lambda}_{T,d}, \boldsymbol{\lambda}_{1,d}\right) & \cdots & \kappa_{\boldsymbol{\eta}}^{d}\left(\boldsymbol{\lambda}_{T,d}, \boldsymbol{\lambda}_{T,d}\right)
  \end{array}\right] \\
  \label{eq:view_concat}
  &\boldsymbol{z}_{1:T,d}^{\text{view}} \sim \mathcal{N}(\boldsymbol{0},\boldsymbol{K}_{\boldsymbol{\eta}}^d)\\
  &\boldsymbol{z}_{1:T}^{\text{view}}=\text{concat}(\boldsymbol{z}_{\cdot,1}^{\text{view}},...,\boldsymbol{z}_{\cdot,D}^{\text{view}}) \\
  &z_k^{\text{pres}} \sim \text{Bernoulli}(\nu_k) \quad \quad \nu_k \sim \text{Beta}(\alpha / K, 1)  \\
  & s_{t,k,n}^{\text{shp}} = \text{Sigmoid}(g_{\text{shp}}(\boldsymbol{z}_k^{\text{obj}},\boldsymbol{z}_t^{\text{view}})_n)\\
  \label{eq:ord}
  &o_{t,k} = g_{\text{ord}}(\boldsymbol{z}_k^{\text{obj}} ,\boldsymbol{z}_t^{\text{view}}) \\
  \label{eq:occlusion}
  & \pi_{t,k,n}=\begin{cases}\prod_{k'=1}^{K}(1 - z_{k'}^{\text{pres}}s_{t,k',n}^{\text{shp}}), \quad \ k=0 \\ \frac{(1-\pi_{t,0,n})(1-z_{k}^{\text{pres}} s_{t,k,n}^{\text{shp}} o_{t,k})}{\sum_{k'=1}^{K} (1-z_{k'}^{\text{pres}} s_{t,k',n}^{\text{shp}} o_{t,k'})}, \ \ k\geq 1  \end{cases} \\
  \label{eq:apc_gen}
  & \boldsymbol{a}_{t,k,n} = \begin{cases} g_{\text{apc}}^{\text{bck}}(\boldsymbol{z}_t^{\text{view}},\boldsymbol{z}^{\text{bck}})_n,\quad \quad \quad \ \ \  k=0 \\ g_{\text{apc}}^{\text{obj}}(\boldsymbol{z}_t^{\text{view}},\boldsymbol{z}_k^{\text{obj}})_n , \quad \quad \quad \ \ \ \ k \geq 1 \end{cases} \\
  \label{eq:likelihood}
  & \boldsymbol{x}_{t,n} \sim \mathcal{N}\Big(\sum \nolimits_{k=0}^K \pi_{t,k,n}  \boldsymbol{a}_{t,k,n}, \sigma_{x}^2 \boldsymbol{I}\Big)
\end{align}
  In the above, the ranges of all indices ($1\leq t \leq T, 1 \leq d \leq D, 1\leq k \leq K, 1\leq n\leq N$) are omitted for simplicity. The way to time embedding $\boldsymbol{w}_t = \text{TimeEncoding}(t)$ can be diverse, e.g. $\boldsymbol{w}_t = \big[ \cos t, \sin t\big]$. $\boldsymbol{\lambda}_{t,d}$ follows a linear Gaussian distribution with a projection matrix $\boldsymbol{A}$, which can be either learned or provided, and $\sigma_w$ is a hyperparameter. $\kappa_{\boldsymbol{\eta}}^d$ is the kernel function corresponding to the $d$th dimension of $\boldsymbol{z}^{\text{view}}$ composed of a neural network $g_{\boldsymbol{\eta}}^d$ and an RBF kernel parameterized with $\boldsymbol{\eta}$, $l$ and $\sigma$ (\cite[]{wilson2016deep}). Each dimension of the viewpoint latent variable $\boldsymbol{z}^{\text{view}}_t$ is generated by a different GP in Eq.\ref{eq:view_concat}. The occlusions are treated in Eq.\ref{eq:occlusion} through sorting the depth values of objects to obtain the soft masks $\boldsymbol{\pi}_{t,k}$ of the background and objects. $\boldsymbol{a}_{t,k}$ in Eq.\ref{eq:apc_gen} denotes the complete appearance of the $k$th object or background in GRB values at time $t$. The likelihood of the $n$th observed pixel at time $t$ is a Gaussian distribution parameterized with $\boldsymbol{\pi}$ and $\boldsymbol{a}$ in Eq.\ref{eq:likelihood}.

  Let $\boldsymbol{\Omega} = \{ \boldsymbol{z}^{\text{obj}}, \boldsymbol{z}^{\text{bck}}, \boldsymbol{z}^{\text{pres}},  \boldsymbol{z}^{\text{view}}, \boldsymbol{\lambda}, \boldsymbol{\nu}\}$ denote the collection of all latent variables, the joint conditional probability of $\boldsymbol{x}_{\mathcal{S}}$ and $\boldsymbol{\Omega}$ can be written as:
\begin{align}
  p(\boldsymbol{x}_{\mathcal{S}},\boldsymbol{\Omega} \mid & \boldsymbol{t}_{\mathcal{S}}) =  \prod \nolimits_{t=1}^T \prod \nolimits_{n=1}^N p(\boldsymbol{x}_{t,n} \mid \boldsymbol{\Omega})  p(\boldsymbol{z}^{\text{bck}}) \notag\\
  & \cdot \prod \nolimits_{d=1}^{D} p(\boldsymbol{z}_{\mathcal{S},d}^{\text{view}}\mid \boldsymbol{\lambda}_{\mathcal{S},d}) \prod \nolimits_{t=1}^T p(\boldsymbol{\lambda}_{t,d}\mid \boldsymbol{t}_{\mathcal{S}} ) \notag \\
  & \cdot \prod \nolimits_{k=1}^K p(\boldsymbol{z}_k^{\text{obj}}) p(z_k^{\text{pres}}\mid \nu_k) p(\nu_k)
\end{align}

\subsection{Inference}
\label{sec:inference}
  Since we can hardly compute the likelihood through integrating out the latent variables $\boldsymbol{\Omega}$, the amortized variational inference approach is employed to approximate the posterior of $\boldsymbol{\Omega}$. In our problem setting, only a subset of the frame collection,  $\boldsymbol{x}_{\mathcal{T}}$, for each video is observed. This implies that the posteriors of  $\boldsymbol{\lambda}_{\mathcal{T}}$ and $\boldsymbol{z}_{\mathcal{T}}^{\text{view}}$ that correspond to $\boldsymbol{x}_{\mathcal{T}}$ can be inferred directly with the inference networks, while the posteriors of $\boldsymbol{\lambda}_{\mathcal{Q}}$ and $\boldsymbol{z}_{\mathcal{Q}}^{\text{view}}$ that correspond to $\boldsymbol{x}_{\mathcal{Q}}$ are hard to compute. We use the least square method to approximate the posterior of $\boldsymbol{\lambda}_{\mathcal{Q}}$ and then explicitly compute the posterior of $\boldsymbol{z}_{\mathcal{Q}}^{\text{view}}$ based on the properties of the GP prior. For simplicity, the parameters in the inference networks are denoted by $\boldsymbol{\phi}$ and the parameters in the learnable kernels in GP are denoted by $\boldsymbol{\eta}$. The variational posterior $q_{\boldsymbol{\phi},\boldsymbol{\eta}}(\boldsymbol{\Omega} \mid \boldsymbol{x}_{\mathcal{T}}, \boldsymbol{t}_{\mathcal{S}})$ conditioned on the observed set can be written as: 
\begin{align}
  \label{eq:posterior}
      q_{\boldsymbol{\phi},\boldsymbol{\eta}}(\boldsymbol{\Omega} \mid & \boldsymbol{x}_{\mathcal{T}}, \boldsymbol{t}_{\mathcal{S}}) =  q_{\boldsymbol{\phi}}(\boldsymbol{z}^{\text{bck}}\mid \boldsymbol{x}_{\mathcal{T}}) q_{\boldsymbol{\phi}}(\boldsymbol{z}_{\mathcal{T}}^{\text{view}} \mid \boldsymbol{x}_{\mathcal{T}},\boldsymbol{t}_{\mathcal{T}})\notag \\
      & \cdot q_{\boldsymbol{\phi}}(\boldsymbol{\lambda}_{\mathcal{T}}\mid \boldsymbol{x}_{\mathcal{T}}, \boldsymbol{t}_{\mathcal{T}})q_{\boldsymbol{\phi}}(\boldsymbol{\lambda}_{\mathcal{Q}} \mid \boldsymbol{\lambda}_{\mathcal{T}}, \boldsymbol{t}_{\mathcal{S}}) \notag \\ 
      & \cdot \prod \nolimits_{k=1}^K q_{\boldsymbol{\phi}}(\boldsymbol{z}_k^{\text{obj}}\mid \boldsymbol{x}_{\mathcal{T}}) q_{\boldsymbol{\phi}}(z_k^{\text{pres}}\mid \boldsymbol{x}_{\mathcal{T}}) q_{\boldsymbol{\phi}}(\nu_k \mid \boldsymbol{x}_{\mathcal{T}}) \notag \\
      & \cdot \prod \limits_{d=1}^{D} q_{\boldsymbol{\eta}}(\boldsymbol{z}_{\mathcal{Q},d}^{\text{view}} \mid \boldsymbol{z}_{\mathcal{T},d}^{\text{view}},\boldsymbol{\lambda}_{\mathcal{S},d}) 
\end{align}
  In the following, we will introduce the inference methods for the observed view-dependent latent variables in Section \ref{subsec:observed_view_inference}, the predicted view-dependent latent variables in Section \ref{subsec:query_view_inference}, and the view-independent object-centric latent variables in Section \ref{subsec:object_inference}. The overview of the inference procedure is illustrated in Figure \ref{fig:overview}. The mathematical details of the inference procedure can be found in the Supplementary Material.

\begin{figure*}[tb]
  \centering
  \includegraphics[width=\textwidth]{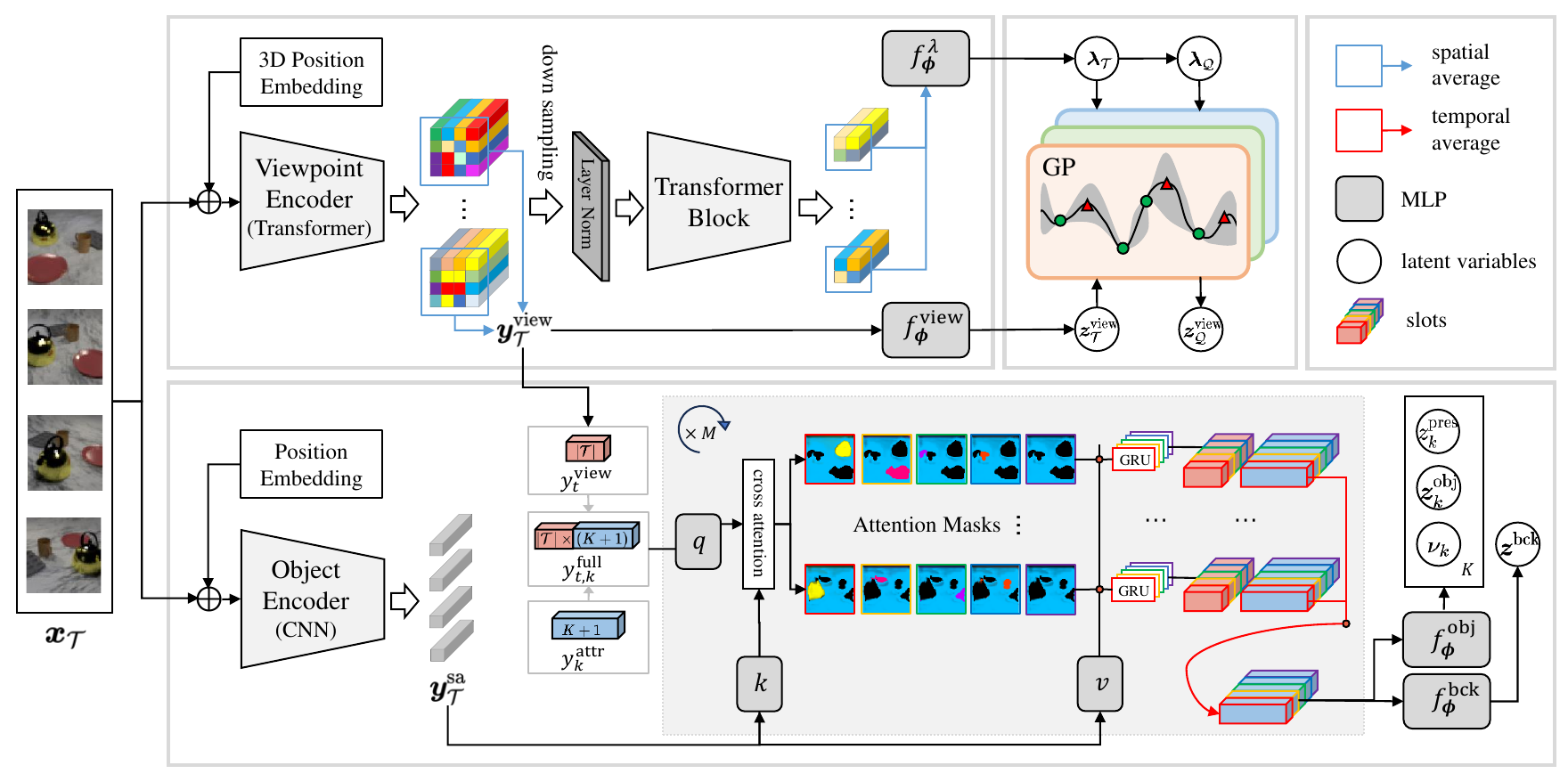}
  \caption{The inference procedure of the proposed model. The three modules correspond to the inference of observed view-dependent latent variables (top-left), the inference of predicted view-dependent latent variables (top-middle), and the inference of view-independent object-centric latent variables (bottom), respectively.}
  \label{fig:overview}
\end{figure*}

\subsubsection{Inference of Observed View-dependent Latents}
\label{subsec:observed_view_inference}
  The posteriors of the viewpoint latent variable $\boldsymbol{z}_t^{\text{view}}$ $(t\in\mathcal{T})$ and the timestamp latent variable $\boldsymbol{\lambda}_{t,d}$ $(t\in\mathcal{T}, 1 \leq d \leq D)$ are defined as:
\begin{align}
    q_{\boldsymbol{\phi}} (\boldsymbol{z}_t^{\text{view}} \mid \boldsymbol{x}_{\mathcal{T}}, \boldsymbol{t}_{\mathcal{T}}) &= \mathcal{N}(\boldsymbol{z}_t^{\text{view}} \mid \boldsymbol{\mu}_t^{\text{view}},\text{diag}(\boldsymbol{\sigma}_{t}^{\text{view}})^2) \notag \\ 
    q_{\boldsymbol{\phi}} (\boldsymbol{\lambda}_{t,d} \mid \boldsymbol{x}_{\mathcal{T}}, \boldsymbol{t}_{\mathcal{T}}) &= \mathcal{N}(\boldsymbol{\lambda}_{t,d} \mid \boldsymbol{\mu}_{t,d}^{\lambda}, \sigma_{\boldsymbol{w}}^2\boldsymbol{I}) \notag 
\end{align}
  where $[\boldsymbol{\mu}_t^{\text{view}}, \boldsymbol{\sigma}_{t}^{\text{view}}] = f_{\boldsymbol{\phi}}^\text{view}(\boldsymbol{x}_{\mathcal{T}})$ and $\boldsymbol{\mu}_{t,d}^{\lambda} = f_{\boldsymbol{\phi}}^\lambda (\boldsymbol{x}_{\mathcal{T}},\boldsymbol{w}_t)$; the variance $\boldsymbol{\sigma}_{\boldsymbol{w}}^2$ is fixed. As Figure \ref{fig:overview} shows: First, $\boldsymbol{x}_{\mathcal{T}}$ is fed into a Transformer block along with a 3D position embedding \cite[]{kabra2021simone}, where the viewpoint information with correlations between frames is learned. 
  A $|\mathcal{T}| \times L \times C$ feature map extracted by the Transformer is averaged over $L = HW$ pixels on the feature map to obtain $\boldsymbol{y}_t^{\text{view}}$ $(t \in \mathcal{T})$, and $\boldsymbol{y}_t^{\text{view}}$ is an intermediate variable to obtain $[\boldsymbol{\mu}_t^{\text{view}}, \boldsymbol{\sigma}_{t}^{\text{view}}]$ and $\boldsymbol{\mu}_{t,d}^{\lambda}$ in $f_{\boldsymbol{\phi}}^\text{view}$ and $f_{\boldsymbol{\phi}}^\lambda$, respectively.

\subsubsection{Inference of Predicted View-dependent Latents}
\label{subsec:query_view_inference}

  Inference of latent variables related to predicted viewpoints is challenging because $\boldsymbol{x}_{\mathcal{Q}}$ is not provided. Therefore, the predicted view-dependent latent variables need to be inferred through the observed viewpoints. We introduce the inference methods for $\boldsymbol{\lambda}_{\mathcal{Q}}$ and $\boldsymbol{z}_{\mathcal{Q}}^{\text{view}}$, respectively.
  
  \textbf{Inference of \textnormal{$\boldsymbol{\lambda}_{\mathcal{Q}}$}.} According to the prior distribution of $\boldsymbol{\lambda}_{t,d}$ defined in Eq.\ref{eq:lambda_prior}, $\boldsymbol{\mu}^\lambda_{t,d}$ of the posterior $q_{\boldsymbol{\phi}} (\boldsymbol{\lambda}_{t,d} \mid \boldsymbol{\lambda}_{\mathcal{T}}, \boldsymbol{t}_{\mathcal{T}})$ can be approximated to satisfy a linear function w.r.t. $\boldsymbol{w}_t$, i.e. $\boldsymbol{\mu}_{t,d}^\lambda = \boldsymbol{\hat{A}}_d \boldsymbol{w}_t, \boldsymbol{\hat{A}}_d\in \mathbb{R}^{D_{\lambda} \times |\boldsymbol{w}_t|}$. Based on the Least Square method, the optimal ${\boldsymbol{\hat{A}}_d^{*}}$ ($1\leq d \leq D$) in the linear set and the posterior of $\boldsymbol{\lambda}_{t,d}$ ($t \in \mathcal{Q}$) are:
\begin{align}
\label{eq:lambda_solve}
    q_{\boldsymbol{\phi}} (\boldsymbol{\lambda}_{t,d} \mid \boldsymbol{\lambda}_{\mathcal{T}}, \boldsymbol{t}_{\mathcal{S}}) = & \mathcal{N}({\boldsymbol{\hat{A}}_d^{*}}\boldsymbol{w}_t, \sigma_{\boldsymbol{w}}^2\boldsymbol{I})\\
    {\boldsymbol{\hat{A}}_d^{*}} = & \boldsymbol{\Phi}_d^{\top}\boldsymbol{W}_{\mathcal{T}}(\boldsymbol{W}_{\mathcal{T}}^{\top} \boldsymbol{W}_{\mathcal{T}})^{-1}
\end{align}
where $\boldsymbol{W}_{\mathcal{T}} = \big[ \boldsymbol{w}_1,...,\boldsymbol{w}_{|\mathcal{T}|}\big]^{\top}\in \mathbb{R}^{|\mathcal{T}|\times |\boldsymbol{w}_t|}$ and $\boldsymbol{\Phi}_d = \big[ \boldsymbol{\mu}_{1,d}, ..., \boldsymbol{\mu}_{|\mathcal{T}|,d}\big]^{\top}\in \mathbb{R}^{|\mathcal{T}| \times D_{\lambda}}$.

  \textbf{Inference of \textnormal{$\boldsymbol{z}_{\mathcal{Q}}^{\text{view}}$}.} $q_{\boldsymbol{\eta}}(\boldsymbol{z}_{\mathcal{Q}}^{\text{view}} \mid \boldsymbol{z}_{\mathcal{T}}^{\text{view}},\boldsymbol{\lambda}_{\mathcal{S}})$ follows the same distribution as the predictive distribution of the GPs (the details can be found in the Supplementary Material):
\begin{align} 
  q_{\boldsymbol{\eta}}(\boldsymbol{z}_{\mathcal{Q}}^{\text{view}} \mid \boldsymbol{z}_{\mathcal{T}}^{\text{view}},\boldsymbol{\lambda}_{\mathcal{S}}) = \prod_{d=1} ^D p_{\boldsymbol{\eta}}(\boldsymbol{z}_{\mathcal{Q},d}^{\text{view}} \mid \boldsymbol{z}_{\mathcal{T},d}^{\text{view}},\boldsymbol{\lambda}_{\mathcal{S},d})
\end{align}
  where $p_{\boldsymbol{\eta}}(\boldsymbol{z}_{\mathcal{Q},d}^{\text{view}} \mid \cdot )$ satisfies the multivariate Gaussian distributions $\mathcal{N}(\boldsymbol{\mu}^\text{view}_{\mathcal{Q},d},\boldsymbol{\Sigma}^\text{view}_{\mathcal{Q},d})$, and the parameters $\boldsymbol{\mu}^\text{view}_{\mathcal{Q},d}$ and $\boldsymbol{\Sigma}^\text{view}_{\mathcal{Q},d}$ are analytical functions of $\boldsymbol{\lambda}_{\mathcal{S},d}$, $\boldsymbol{z}_{\mathcal{T},d}^{\text{view}}$ and $\boldsymbol{\eta}$. 

\subsubsection{Inference of View-independent Latents}
\label{subsec:object_inference}
  The posteriors of the view-independent object-centric latent variables $\{ \boldsymbol{z}^{\text{bck}}, \boldsymbol{z}^{\text{obj}}, \boldsymbol{z}^{\text{pres}}, \boldsymbol{\nu} \}$ in Eq.\ref{eq:posterior} are defined as:
\begin{align}
    \label{eq:obj_post}
    q_{\boldsymbol{\phi}}(\boldsymbol{z}^{\text{bck}} \mid \boldsymbol{x}_\mathcal{T}) &= \mathcal{N}(\boldsymbol{z}^{\text{bck}} \mid \boldsymbol{\mu}^{\text{bck}},\text{diag}(\boldsymbol{\sigma}^{\text{bck}})^2)\\
    q_{\boldsymbol{\phi}}(\boldsymbol{z}_k^{\text{obj}} \mid \boldsymbol{x}_\mathcal{T}) &= \mathcal{N}(\boldsymbol{z}_k^{\text{obj}} \mid \boldsymbol{\mu}_k^{\text{obj}},\text{diag}(\boldsymbol{\sigma}_k^{\text{obj}})^2) \\
    q_{\boldsymbol{\phi}}(z_k^{\text{pres}} \mid \boldsymbol{x}_\mathcal{T}) &= \text{Bernoulli}(z_k^{\text{pres}} \mid \kappa_k) \\
    q_{\boldsymbol{\phi}}(\nu_k \mid \boldsymbol{x}_\mathcal{T}) &= \text{Beta}(\nu_k \mid \tau_{k,1}, \tau_{k,2})
\end{align}
  where the default range of $k$ is $1\leq k \leq K$. All the parameters of the above distributions will pass through a sequential extension of Slot Attention \cite[]{locatello2020object}, which is illustrated in Figure \ref{fig:overview}.

  The model maintains $K+1$ slots $\boldsymbol{y}^{\text{attr}} = [ \boldsymbol{y}^{\text{bck}},\boldsymbol{y}_1^{\text{obj}}, ...,\boldsymbol{y}_K^{\text{obj}}]$, $\boldsymbol{y}_k^{\text{attr}} \in \mathbb{R}^{D_s}$. Different from Slot Attention \cite[]{locatello2020object}, two types of initialization are employed for the foreground objects and the background, respectively. Then $\boldsymbol{y}_k^{\text{attr}}$ is combined with $\boldsymbol{y}_t^{\text{view}} \in \mathbb{R}^{D_v}$ ($t \in \mathcal{T}$) obtained in Section \ref{subsec:observed_view_inference} to produce $|\mathcal{T}| \times (K+1) $ slots $\boldsymbol{y}_{t,k}^{\text{full}} \in \mathbb{R}^{D_f} $ with the viewpoint information, where $D_f = D_s + D_v$. We use another encoder to extract the feature maps of $\boldsymbol{x}_{\mathcal{T}}$, denoted as $\boldsymbol{y}_{\mathcal{T}}^{\text{sa}}$. We do $M$ iterations like Slot Attention. In each iteration, Eq.\ref{eq:cross_attn} first uses the cross attention to obtain the attention masks $\boldsymbol{a}_t \in \mathbb{R}^{N\times (K+1)}$ of $K$ objects and the background. Then, the pixel-wise normalized masks of all the objects and background are multiplied with the value of $\boldsymbol{y}_t^{\text{sa}}$ to obtain the hidden state $\boldsymbol{u}_t \in \mathbb{R}^{(K+1)\times D_f}$ for GRU updating. In addition, we perform temporal mean over the updated attribute part of $\boldsymbol{\hat{y}}_{t,k}^{\text{full}}$ after GRU updating.
\begin{align}
\label{eq:cross_attn}
  \boldsymbol{a}_{t} &= \underset{K+1}{\text{Softmax}} \Big( \frac{k(\boldsymbol{y}_t^{\text{sa}}) \cdot q(\boldsymbol{y}_{t,1:K+1}^{\text{full}})^{\top}}{\sqrt{D_f}}\Big)\\
  \boldsymbol{u}_t &= \sum_{n=1}^N \Big(\underset{N}{\text{Softmax}}\big( \log \boldsymbol{a}_{t,n}\big) \cdot v(\boldsymbol{y}_{t,n}^{\text{sa}}) \Big)\\
  \hat{\boldsymbol{y}}^{\text{full}}_{t,k} &= \text{GRU}(\boldsymbol{y}^{\text{full}}_{t,k},\boldsymbol{u}_{t,k}) \quad \big[ \hat{\boldsymbol{y}}_{t,k}^{\text{attr}},\hat{\boldsymbol{y}}_{t,k}^{\text{view}} \big] \stackrel{\text{split}}{\leftarrow} \hat{\boldsymbol{y}}^{\text{full}}_{t,k}\\
  \boldsymbol{y}_{k}^{\text{attr}} &={\text{mean}}_{|\mathcal{T}|}\Big(\hat{\boldsymbol{y}}_{1:|\mathcal{T}|,k}^{\text{attr}}\Big)
\end{align}
  where $k$, $q$ and $v$ are MLPs for producing key, query and value, respectively. The procedure maintains the permutation invariance w.r.t. the input order of frames. $\boldsymbol{\mu}^{\text{bck}}$ and $\boldsymbol{\sigma}^{\text{bck}}$ are obtained through the neural network $f_{\boldsymbol{\phi}}^{\text{bck}}$ with $\boldsymbol{y}^{\text{bck}}$ as input; $\boldsymbol{\mu}_k^{\text{obj}},\boldsymbol{\sigma}_k^{\text{obj}},\kappa_k,\tau_{k,1}, \tau_{k,2}$ are obtained through the shared neural network $f_{\boldsymbol{\phi}}^{\text{obj}}$ with $\boldsymbol{y}_k^{\text{obj}}$ as input.

\subsection{Training}
\label{sec:training}
  Optimizing the evidence lower bound (ELBO) for all frames (including both observed and predicted frames) is unstable. To solve this problem, a two-stage training procedure is adopted. Let $\boldsymbol{\Omega}_{\mathcal{S}} = \{\boldsymbol{\Omega}_{\mathcal{T}},\boldsymbol{\Omega}_{\mathcal{Q}} \}$, where $\boldsymbol{\Omega}_{\mathcal{T}} = \big\{ \boldsymbol{z}^{\text{bck}},\boldsymbol{z}^{\text{obj}}, \boldsymbol{z}^{\text{pres}}, \boldsymbol{\nu}, \boldsymbol{\lambda}_{\mathcal{T}}, \boldsymbol{z}_{\mathcal{T}}^{\text{view}}\big\}$ and $\boldsymbol{\Omega}_{\mathcal{Q}} = \big\{ \boldsymbol{z}^{\text{bck}},\boldsymbol{z}^{\text{obj}}, \boldsymbol{z}^{\text{pres}}, \boldsymbol{\nu}, \boldsymbol{\lambda}_{\mathcal{Q}}, \boldsymbol{z}_{\mathcal{Q}}^{\text{view}}\big\}$, i.e. the view-independent latent variables share in both $\boldsymbol{\Omega}_{\mathcal{T}}$ and $\boldsymbol{\Omega}_{\mathcal{Q}}$. The two-stage losses are as follows:
\begin{align}
  \mathcal{L}_1 = & -\mathbb{E}_{q_{\boldsymbol{\phi},\boldsymbol{\eta}}( \boldsymbol{\Omega}_{\mathcal{T}} \mid \boldsymbol{x}_{\mathcal{T}})}\big[ \log p_{\boldsymbol{\theta},\boldsymbol{\eta}}(\boldsymbol{x}_{\mathcal{T}} \mid \boldsymbol{\Omega}_{\mathcal{T}})\big] \notag \\
  & + D_{KL}\Big( q_{\boldsymbol{\phi},\boldsymbol{\eta}}( \boldsymbol{\Omega}_{\mathcal{T}} \mid \boldsymbol{x}_\mathcal{T}) \| p_{\boldsymbol{\theta},\boldsymbol{\eta}}(\boldsymbol{\Omega}_\mathcal{T})\Big) \\
  \mathcal{L}_2 = & -\frac{1}{|\mathcal{T}|}\mathbb{E}_{q_{\boldsymbol{\phi},\boldsymbol{\eta}}( \boldsymbol{\Omega}_{\mathcal{T}} \mid \boldsymbol{x}_\mathcal{T},\boldsymbol{t}_{\mathcal{T}})}\big[ \log p_{\boldsymbol{\phi},\boldsymbol{\eta}}(\boldsymbol{x}_\mathcal{T} \mid \boldsymbol{\Omega}_{\mathcal{T}})\big]\notag \\ 
 -\frac{1}{|\mathcal{Q}|}  & \mathbb{E}_{q_{\boldsymbol{\phi}}(\boldsymbol{\Omega}_{\mathcal{T}}\mid \boldsymbol{x}_\mathcal{T},\boldsymbol{t}_{\mathcal{T}} )q_{\boldsymbol{\phi},\boldsymbol{\eta}}( \boldsymbol{\Omega}_{\mathcal{Q}} \mid \boldsymbol{\Omega}_\mathcal{T}, \boldsymbol{t}_Q)}\big[ \log p_{\boldsymbol{\theta},\boldsymbol{\eta}}(\boldsymbol{x}_\mathcal{Q} \mid \boldsymbol{\Omega}_{\mathcal{Q}}) \big] \notag \\
   +\beta & D_{KL}\Big( q_{\boldsymbol{\phi},\boldsymbol{\eta}}( \boldsymbol{\Omega}_{\mathcal{S}}\mid \boldsymbol{x}_{\mathcal{T}}, \boldsymbol{t}_{\mathcal{S}}) \| p_{\boldsymbol{\theta},\boldsymbol{\eta}}(\boldsymbol{\Omega}_{\mathcal{S}}\mid \boldsymbol{t}_{\mathcal{S}})\Big)
\end{align}
  where $\mathcal{L}_1$ is a standard ELBO of $\boldsymbol{\Omega}_{\mathcal{T}}$ on $\boldsymbol{x}_{\mathcal{T}}$ to learn object-centric representations from multiple frames and does not depend on $\boldsymbol{t}_{\mathcal{S}}$; while $\mathcal{L}_2$ adopts the curriculum learning to learn the function of viewpoint latent variables w.r.t. $\boldsymbol{t}_{\mathcal{S}}$. Let $\mathcal{S}'$ denote the subset of $\mathcal{S}$ and $|\mathcal{S}'|$ is scheduled to gradually increase during training. $\mathcal{S}'$ will be randomly divided into $\mathcal{T}$ and $\mathcal{Q}$, where $|\mathcal{Q}|\sim U(1,C)$ ($C<|S'|$ and increases during training). $\mathcal{L}_2$ averages the observed and predicted losses to balance the two losses, where $\beta \geq 1$ is a hyper-parameter follows \cite[]{burgess2018understanding}. Note that the reconstruction performance of $\mathcal{L}_2$ is worse than that of the first stage; however, it can perform well on the prediction task.

\section{Experiments}
\begin{figure*}[tb]
  \centering
  \setlength{\fboxrule}{3pt}
  \fcolorbox{red}{white}{\begin{minipage}[t]{0.4\textwidth}
    \centering
    \subfigure[MulMON]{
        \centering
        \includegraphics[width=0.45\linewidth]{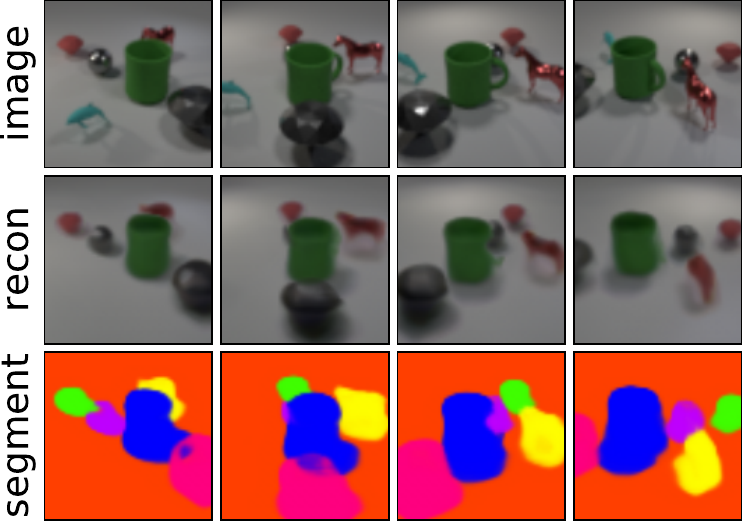}
    }
    \subfigure[SIMONe]{
        \centering
        \includegraphics[width=0.45\linewidth]{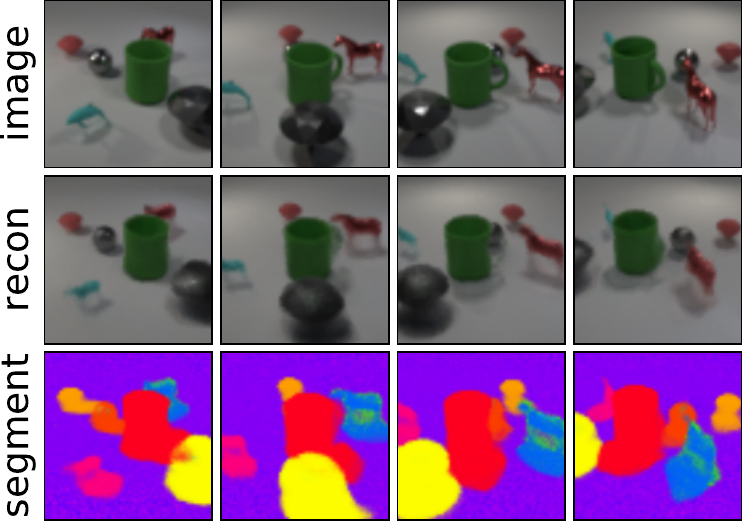}
    }

    \subfigure[OCLOC]{
        \centering
        \includegraphics[width=0.45\linewidth]{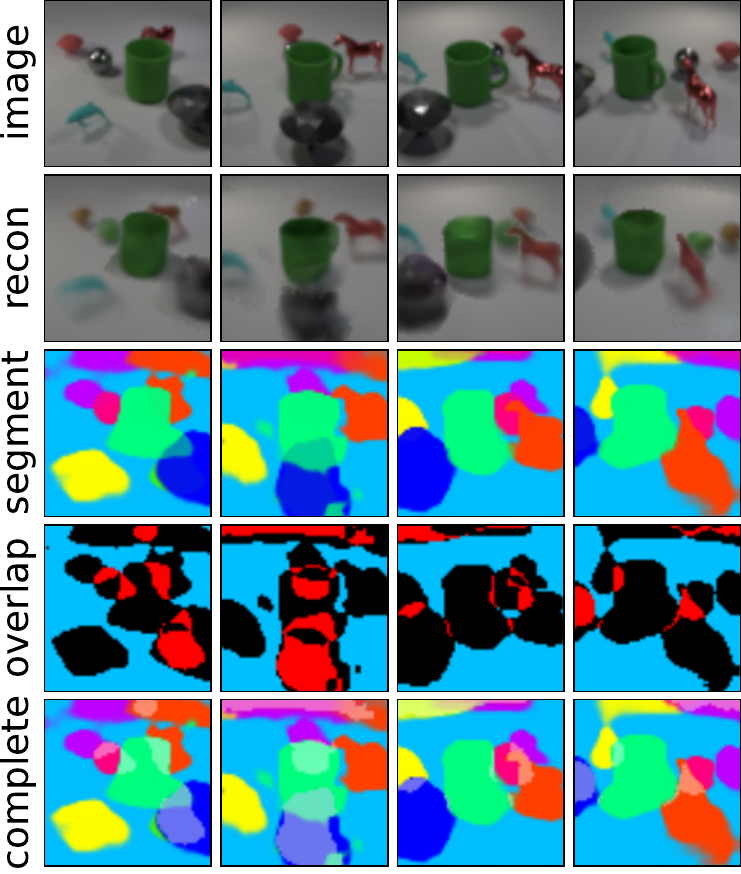}
    }
    \subfigure[Ours]{
        \centering
        \includegraphics[width=0.45\linewidth]{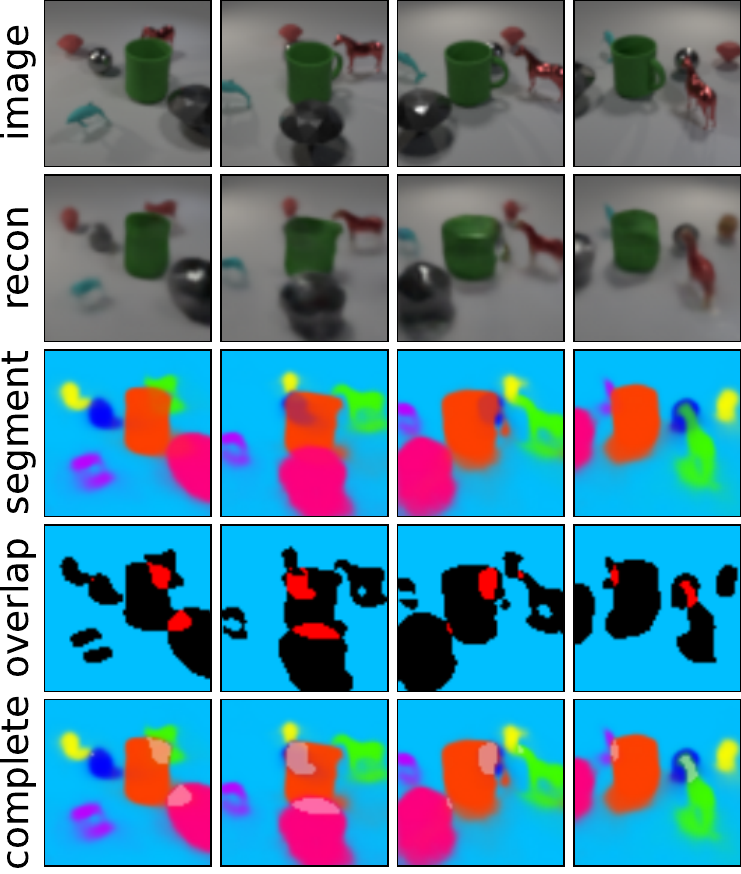}
    }
  \end{minipage}
  }
  \fcolorbox{blue}{white}{\begin{minipage}[t]{0.495\textwidth}
      \centering
      \subfigure[MulMON]{
          \centering
          \includegraphics[width=0.86\linewidth]{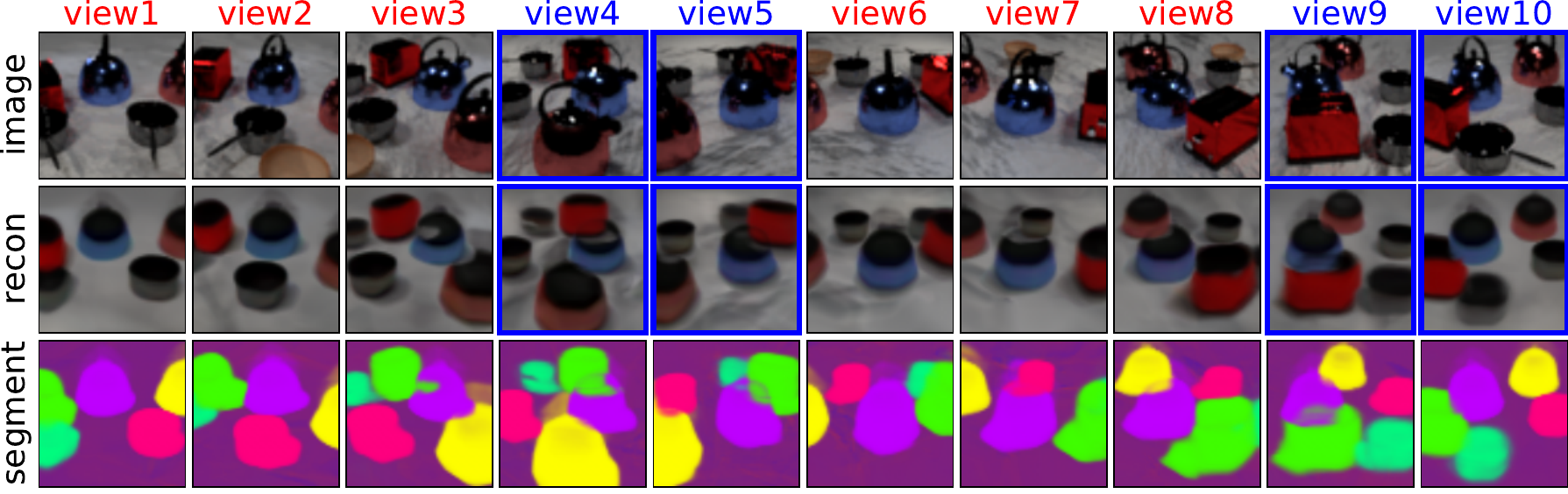}
      }

      \subfigure[Ours]{
          \centering
          \includegraphics[width=0.86\linewidth]{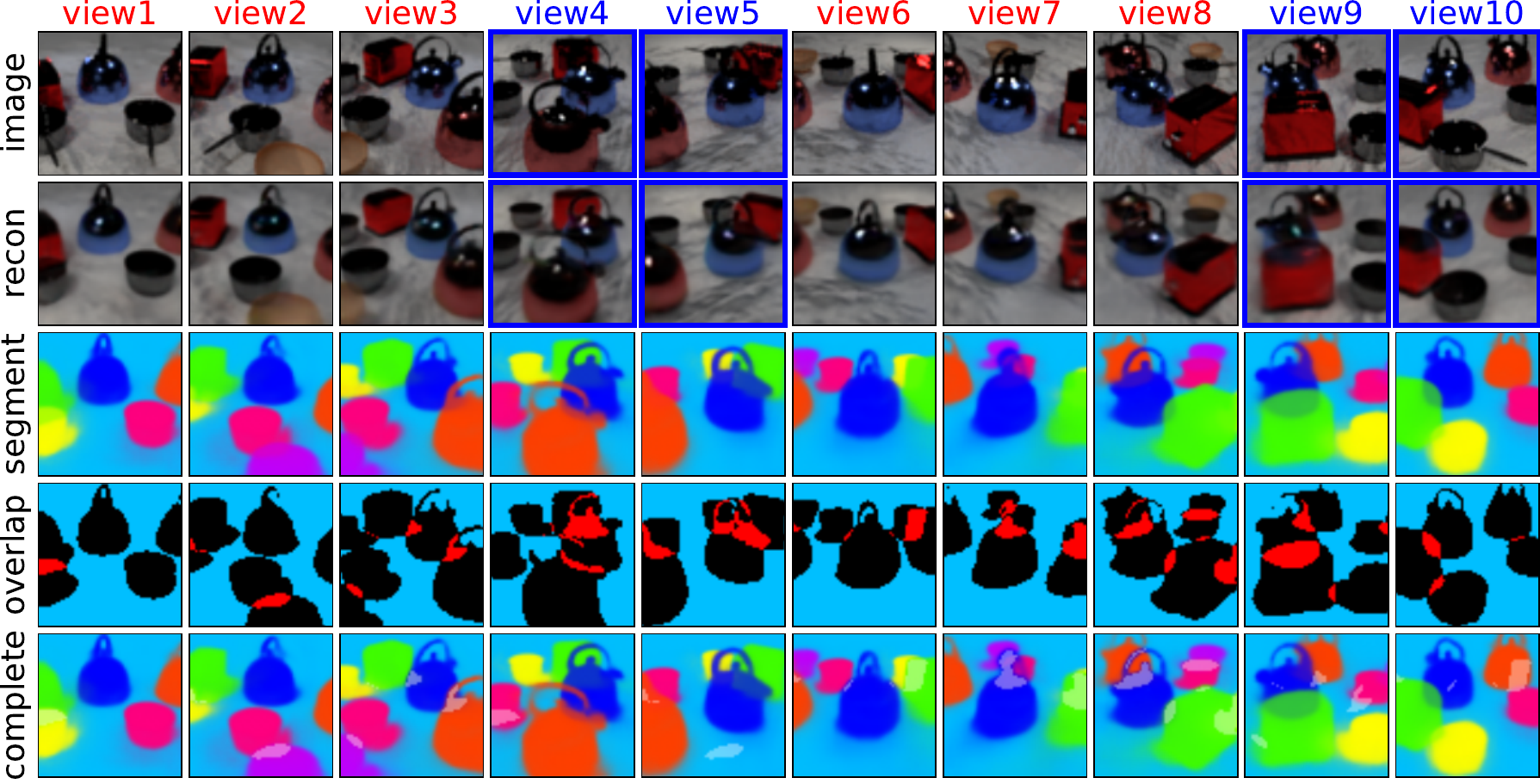}
      }
  \end{minipage}
  }
  \caption{\textbf{Left}: Visualization results of the compared methods on the \emph{observation} set of CLEVER-COMPLEX, where four consecutive frames are demonstrated. \textbf{Right}: Visualization results on the \emph{prediction} set of SHOP-SIMPLE. The `images' in blue boxes are unobserved ground truths and the `recons' in blue boxes are predicted results.}
  \label{fig:exp_overview}
\end{figure*}

\begin{figure*}[tb]
	\centering
  \setlength{\fboxrule}{3pt}
  \fcolorbox{red}{white}{
	\begin{minipage}[b]{0.605\textwidth}
		\subfigure[Video Recomposition (SHOP-COMPLEX)]{
			\includegraphics[width=\linewidth]{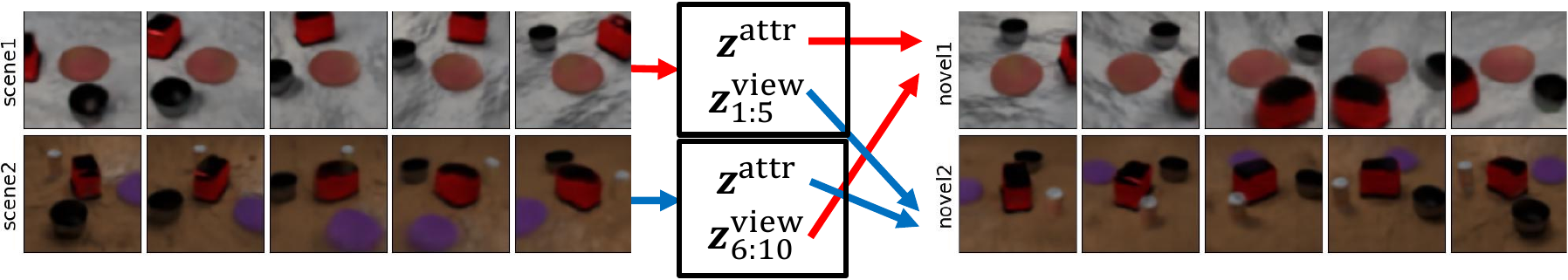} 
			\label{fig:video_recomp}
    }
	\end{minipage}
  }
  \fcolorbox{blue}{white}{
	\begin{minipage}[b]{0.3\textwidth}
		\subfigure[Video Generation (CLEVR-SIMPLE)]{
			\includegraphics[width=\linewidth]{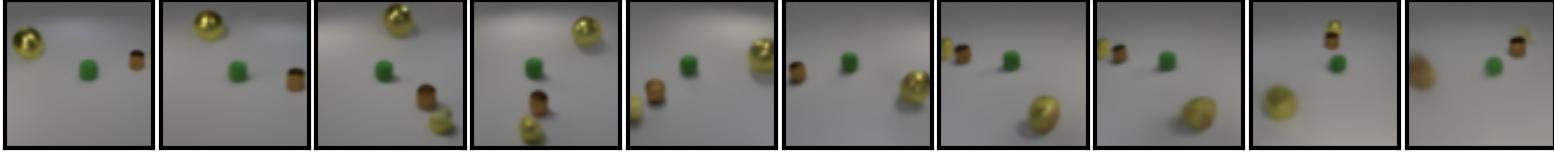}  
			\label{fig:clevr_gen}
    }

    \subfigure[Video Generation (SHOP-SIMPLE)]{
			\includegraphics[width=\linewidth]{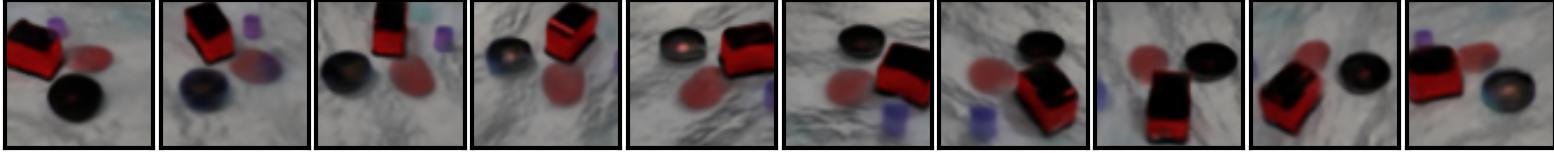}  
			\label{fig:shop_gen}
    }
	\end{minipage}
  }
	\caption{\textbf{Left}: Scene image generation from novel viewpoints through recomposing viewpoint representations and object-centric representations. \textbf{Right}: Video generation based on CLEVR-SIMPLE and SHOP-SIMPLE.}
	\label{fig:video_prop}
\end{figure*}

  We design experiments to investigate 1) how well the proposed model performs compared to state-of-the-art multi-view models in object-centric video decomposition on the observation set; 2) whether the proposed model can disentangle the 3D scene into object-centric view-invariant representations and viewpoint representations; 3) how well the proposed model handles occlusions compared to existing methods; 4) how well the proposed model makes the prediction only depending on timestamps; and 5) whether the proposed model can generate videos.

  To validate the above, we compare the proposed model\footnote{The code is available at https://github.com/FudanVI/\\compositional-scene-representation-toolbox} with three state-of-the-art models, \textbf{MulMON} \cite[]{li2020learning} with viewpoint annotations, viewpoint-free models \textbf{SIMONe} \cite[]{kabra2021simone} and \textbf{OCLOC} \cite[]{yuan2022unsupervised}. We design four synthetic video datasets, called CLEVR-SIMPLE, CLEVR-COMPLE, SHOP-SIMPLEX, and SHOP-COMPLEX, through modifying multi-view CLEVR \cite[]{johnson2017clevr} and SHOP \cite[]{nazarczuk2020shop} based on the official code. The two SHOP datasets are more challenging than the two CLEVR datasets in terms of the object texture; the two COMPLEX versions are more challenging than the two SIMPLE versions because of more types of objects and backgrounds. 
  
  We train the proposed model with the introduced two-stage strategy. Stage 1 can reconstruct the observation set without supervision while Stage 2 can predict unobserved set only with timestamp supervision. We train the proposed model on all the datasets using the Adam optimizer with a learning rate 4e-4 for 300K gradient steps. The increment of curriculum learning is 2. 

  \textbf{Video Decomposition.} Since the proposed model maintains the view-invariant object-centric representations in 3D structure, video decomposition is crucial to evaluating the completeness and accuracy of learned representations. Figure \ref{fig:exp_overview} (Left) demonstrates the visualization results on CLEVR-COMPLEX. The proposed model can accurately represent objects with complex shapes from multiple viewpoints and build crisp segregation between the foregrounds and the background. Moreover, the proposed model tends to treat shadows as parts of objects (e.g., the horse in Figure \ref{fig:exp_overview}(d)), it is reasonable for shadows to be blended with the corresponding objects due to lighting. Surprisingly, the shadow area is noticeably smaller than those of other models.

  Table \ref{table:compare}(a) reports the segmentation performance in terms of foreground objects. ARI-O measures how accurately a video is decomposed into separate objects. We find that, except for CLEVR-SIMPLE, the proposed model outperforms the other models, especially on the two SHOP datasets, probably because the 3D representations integrity of objects helps reconstruct better masks. SIMONe and OCLOC fail to capture the objects on SHOP-COMPLEX. A possible reason is that the background is indistinguishable with the objects in SHOP-COMPLEX, such that these models cannot represent the background separately during the inference. Although OCLOC models the background separately, sampling from permutation-equivalent slots may affect the extraction of the background representation.
  
\begin{table*}
  \caption{Performance comparison of MulMON, SIMONe and the proposed model (Ours). ARI-O is adopted for evaluating segmentation, IoU and OOA are adopted for evaluating segmentation with occlusions, and MSE is adopted for evaluating reconstruction. Except for MSE in (d), all results are recorded in `mean $\pm$ std' over 5 random seeds. `-S' and `-C' are short for `SIMPLE' and `COMPLEX', respectively.}\label{table:compare}	
\centering
\subtable[ARI-O (observation set)]{
    \begin{minipage}[a]{0.48\textwidth}
    \centering
    \renewcommand\arraystretch{1.4}
    \scalebox{0.7}{
  \begin{tabular}{ccccc} 
      \toprule[1.5pt]
    \multirow{2}{*}{Model}&CLEVR-S&CLEVR-C& SHOP-S &  SHOP-C \\
    \cline{2-5}
    ~ & ARI-O$\uparrow$ & ARI-O$\uparrow$ & ARI-O$\uparrow$ & ARI-O$\uparrow$ \\
  \hline
  MulMON (cond) & \large \textbf{96.4}  $\pm$ 0.1  & \large 92.9  $\pm$ 0.2 & \large 88.3  $\pm$ 0.6 & \large 87.1  $\pm$ 0.2\\
  SIMONe & \large 91.0  $\pm$ 0.0  & \large 91.4  $\pm$ 0.0 & \large 55.3  $\pm$ 0.0 & \large 33.5  $\pm$ 0.0\\
  OCLOC & \large 92.7  $\pm$ 0.8  & \large 82.7  $\pm$ 0.8 & \large 91.3  $\pm$ 0.4 & \large 29.3  $\pm$ 0.5\\
  Ours & \large 95.9  $\pm$ 0.3  & \large \textbf{94.1}  $\pm$ 0.3 & \large \textbf{95.8} $\pm$ 0.1 & \large \textbf{94.9}  $\pm$ 0.4\\
  \bottomrule[1.5pt]
  \end{tabular}
  }
\end{minipage}
}
\subtable[IoU and OOA (observation set)]{
    \begin{minipage}[a]{0.48\textwidth}
    \centering
    \renewcommand\arraystretch{1.4}
    \scalebox{0.7}{
  \begin{tabular}{c|cc|cc} 
      \toprule[1.5pt]
    \multirow{2}{*}{Model}&\multicolumn{2}{c|}{IoU$\uparrow$}& \multicolumn{2}{c}{OOA$\uparrow$} \\
    \cline{2-5}
    ~ & OCLOC & Ours & OCLOC & Ours \\
  \hline
  CLEVR-S & \large 45.6 $\pm$ 0.2  & \large \textbf{59.5}  $\pm$ 0.5 & \large 93.6  $\pm$ 1.2 & \large \textbf{95.3}  $\pm$ 1.1\\
  CLEVR-C & \large 35.1  $\pm$ 0.2  & \large \textbf{50.9}  $\pm$ 0.4 & \large 89.1  $\pm$ 1.2 & \large \textbf{93.0}  $\pm$ 0.8\\
  SHOP-S &\large 61.9  $\pm$ 0.6  & \large \textbf{65.9}  $\pm$ 0.1 & \large 72.8 $\pm$ 1.4 & \large \textbf{78.9}  $\pm$ 0.4\\
  SHOP-C & \large 21.5  $\pm$ 0.3  & \large \textbf{66.2}  $\pm$ 0.6 & \large 57.9 $\pm$ 1.9 & \large \textbf{81.8}  $\pm$ 1.3\\
  \bottomrule[1.5pt]
  \end{tabular}
  }
\end{minipage}
}
\quad 
\centering
\subtable[ARI-O (prediction set)]{
    \begin{minipage}[a]{0.48\textwidth}
    \centering
    \renewcommand\arraystretch{1.4}
    \scalebox{0.7}{
  \begin{tabular}{c|ccccc} 
      \toprule[1.5pt]
    \multicolumn{2}{c}{ \multirow{2}{*}{Model} }&CLEVR-S&CLEVR-C& SHOP-S &  SHOP-C \\
    \cline{3-6}
    \multicolumn{2}{c}{~} & ARI-O$\uparrow$ & ARI-O$\uparrow$ & ARI-O$\uparrow$ & ARI-O$\uparrow$\\
  \hline
  \multirow{2}{*}{Mode 1}&MulMON & \large \textbf{96.2}  $\pm$ 0.1  & \large 91.5  $\pm$ 0.3 & \large 88.3  $\pm$ 0.5 & \large 86.9  $\pm$ 0.7\\
  ~& Ours & \large 95.5  $\pm$ 0.5  & \large \textbf{95.5}  $\pm$ 0.9 & \large \textbf{96.0}  $\pm$ 0.3 & \large \textbf{92.9}  $\pm$ 0.4\\
  \hline
  \multirow{2}{*}{Mode 2}&MulMON & \large \textbf{96.9}  $\pm$ 0.2  & \large 94.5  $\pm$ 0.2 & \large 87.1  $\pm$ 0.6 & \large 86.0  $\pm$ 0.6\\
  ~& Ours & \large 95.1  $\pm$ 0.5  & \large \textbf{95.0}  $\pm$ 0.6 & \large \textbf{95.5}  $\pm$ 0.1 & \large \textbf{93.8}  $\pm$ 0.8\\
  \bottomrule[1.5pt]
  \end{tabular}
  }
\end{minipage}
}
\subtable[MSE (prediction set)]{
    \begin{minipage}[a]{0.48\textwidth}
    \centering
    \renewcommand\arraystretch{1.4}
    \scalebox{0.7}{
  \begin{tabular}{c|ccccc} 
      \toprule[1.5pt]
    \multicolumn{2}{c}{ \multirow{2}{*}{Model} }&CLEVR-S&CLEVR-C& SHOP-S &  SHOP-C \\
    \cline{3-6}
    \multicolumn{2}{c}{~} & MSE$\downarrow$ & MSE$\downarrow$ & MSE$\downarrow$ & MSE$\downarrow$ \\
  \hline
  \multirow{2}{*}{Mode 1}&MulMON & \large \textbf{0.0014} & \large \textbf{0.0020} & \large 0.0049 & \large 0.0038\\
  ~& Ours & \large 0.0018 & \large 0.0021 & \large \textbf{0.0034} & \large \textbf{0.0036}\\
  \hline
  \multirow{2}{*}{Mode 2}&MulMON & \large \textbf{0.0014} & \large \textbf{0.0020} & \large 0.0050 & \large 0.0038\\
  ~& Ours & \large 0.0017 & \large 0.0024 & \large \textbf{0.0035} & \large \textbf{0.0038}\\
  \bottomrule[1.5pt]
  \end{tabular}
  }
\end{minipage}
}
\end{table*}
\begin{figure*}[ht]
  \centering
  \begin{minipage}{\linewidth}
      \centering
      \includegraphics[width=0.33\linewidth]{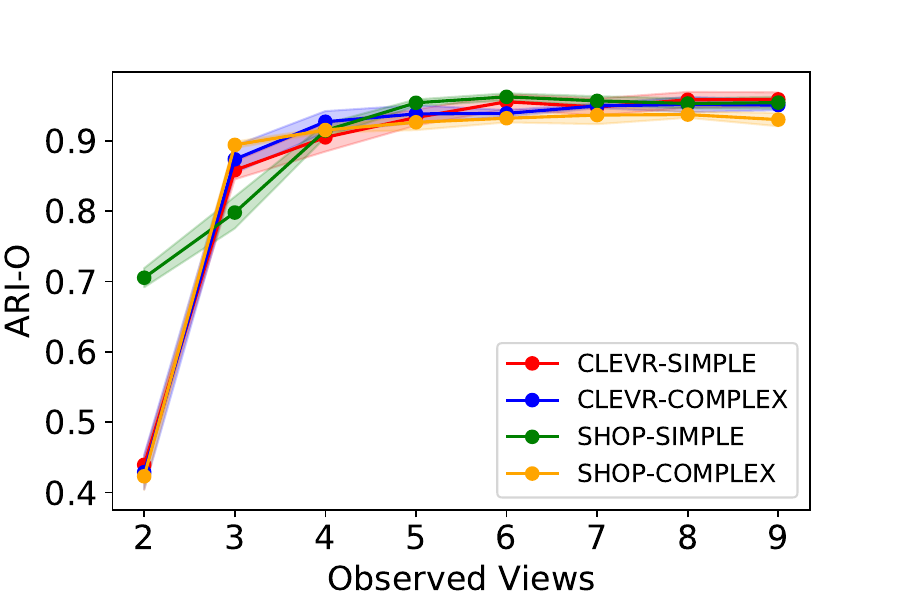}
      \centering
      \includegraphics[width=0.33\linewidth]{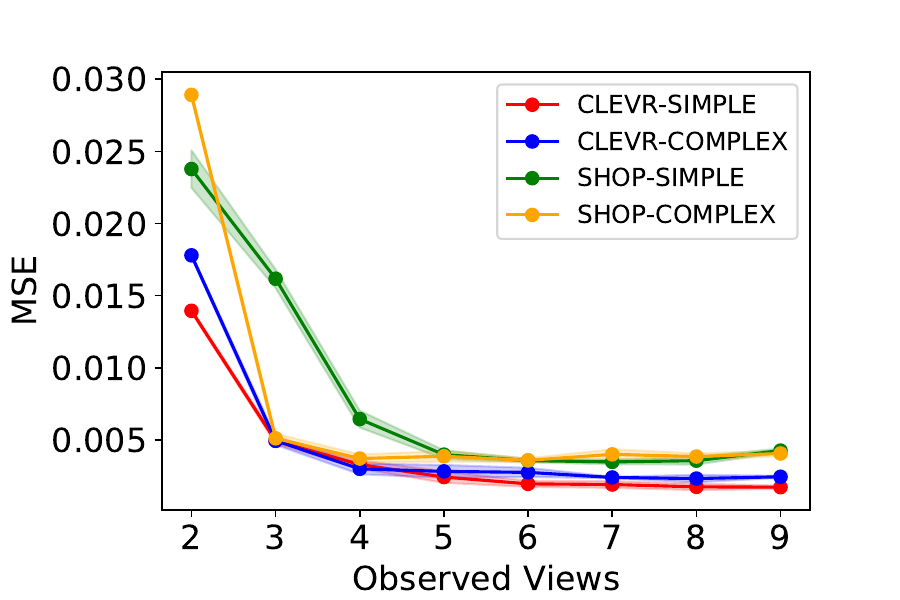}
      \centering
      \includegraphics[width=0.33\linewidth]{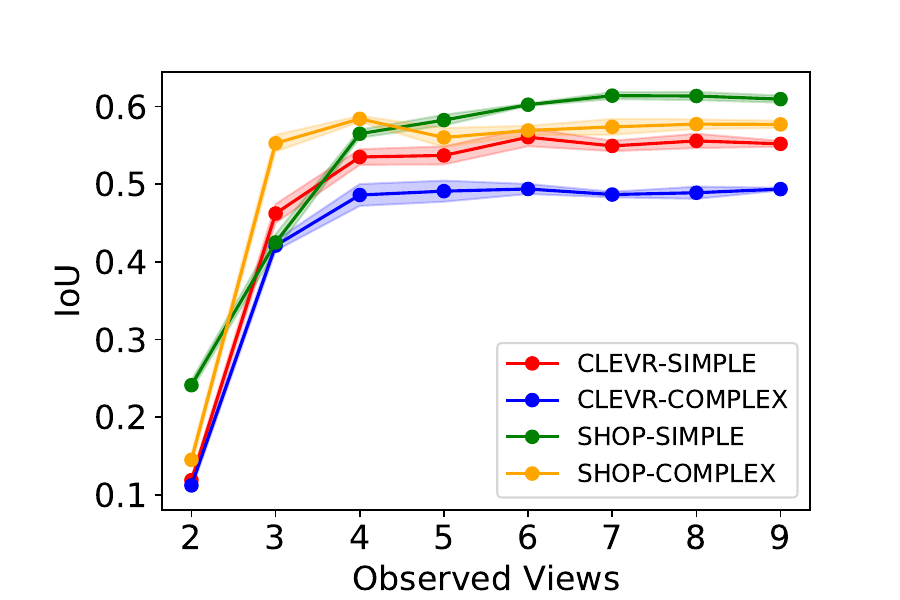}
  \caption{Single-view prediction performance in ARI-O, MSE, and IoU in terms of the number of observed views. All results are tested with 5 random seeds and each point on a curve is the mean value and the shaded band denotes $\pm$std.}
  \label{fig:curve}
  \end{minipage}
\end{figure*}

  \textbf{Video Recomposition.} An intriguing experiment is to generate scene images from novel viewpoints through cross-combining viewpoint representations and object-centric representations of objects (including $\boldsymbol{z}^{\text{bck}}$ and $\boldsymbol{z}^{\text{obj}}$). The recomposition is implemented as follows: We randomly choose two videos (each comprises 10 frames) and select the first 5 frames from one video and select the last 5 frames from the other. Then, we encode the selected frames into viewpoint representations and object-centric representations. Finally, we combine the first five object-centric representations from one video and the last five viewpoint representations from the other frame-wisely to generate the scene images from novel viewpoints. Figure \ref{fig:video_prop}(a) demonstrates that disentangled object-centric and viewpoint representations from different scenes can be effectively coupled, based on which the proposed model can generate novel views. 

  \textbf{Occlusion Evaluation.} Among the compared methods, only OCLOC is designed to handle occlusions. The comparison results on CLEVR-COMPLEX are visualized in Figure \ref{fig:exp_overview} (c) and (d). As the camera moves counterclockwise around the center, a gray ball is completely occluded behind the green mug in the second frame. The proposed model can reconstruct the complete shape of an object even it is completely occluded (e.g. the gray ball). We evaluate IoU and OOA used in \cite[]{yuan2019generative} that respectively assess the quality of reconstructed complete shapes and the accuracy of the estimated pairwise ordering of objects. The proposed model clearly outperforms OCLOC, probably because OCLOC samples the pixel-wise shape during the generation, which produces noisy pixels and large shadows.

  \textbf{GP Prediction.} Due to modeling the viewpoint latent variables with GPs, we can use the analytical posterior of $\boldsymbol{z}_{\mathcal{Q}}^{\text{view}}$ to predict the rest viewpoints given the observation set. In our experimental setting, 10 consecutive viewpoint representations in Figure \ref{fig:exp_overview} satisfy the GPs and we randomly remove four frames (i.e. the ground truths in the blue boxes are unobserved). The remaining six frames are encoded to infer $\boldsymbol{z}^{\text{obj}}$, $\boldsymbol{z}^{\text{bck}}$, $\boldsymbol{\lambda}_{\mathcal{T}}$, $\boldsymbol{\lambda}_{\mathcal{Q}}$, $\boldsymbol{z}_{\mathcal{T}}^{\text{view}}$ and $\boldsymbol{z}_{\mathcal{Q}}^{\text{view}}$. The four viewpoint representations predicted by GPs are concatenated with the object-centric representations to reconstruct the scene images. Figure \ref{fig:exp_overview}(f) shows that the proposed model can predict arbitrary-time frames given the observation. Compared with MulMON which uses viewpoint annotations, the proposed model can additionally process occlusions while reconstructing frames from novel viewpoints. To assess the segmentation performance and reconstruction quality on the prediction set, we choose four fixed frames in Mode 1 and Mode 2 to make prediction (see the Supplementary Material for details). Table \ref{table:compare}(c) and (d) show that the proposed model is comparable to MulMON on the two CLEVR datasets and clearly outperforms MulMON on the two SHOP datasets. The reconstruction loss helps improve the texture characterization of objects, which may be the reason that the proposed model achieves better performance in MSE on the two SHOP datasets.

  \textbf{Ablation Study.} GPs have a generic nature: As the number of observed variables increases, the prediction uncertainty gradually decreases. We assume the number of observed frames (hyperparameter) to be the most important factor that affects the accuracy and uncertainty of the prediction. To verify the assumption, we fix a single frame and gradually increase the number of observed frames from 2 to 9. The viewpoint representations of both the predicted frame and the observed frames are used to construct GPs together. We execute the GP prediction and plot the performance curves in ARI-O, MSE, and IoU in terms of the number of observed views in Figure \ref{fig:curve}. One can see that the proposed model gradually reduces the uncertainty and improves the performance as the number of observed views increases, and tends to be stable after the number of observed views achieves 5. 

  \textbf{Video Generation.} As we model the viewpoint latent variables with GPs, we can generate videos from the GPs along the timeline. Figure \ref{fig:video_prop}(b) and (c) plot two example videos with 10 frames generated based on CLEVR-SIMPLE and SHOP-SIMPLE. One can find that the 10 frames obviously rotate clockwise around the center, reflecting the captured correlations between viewpoints; meanwhile, the generated objects and backgrounds have no irregular shapes. 

\section{Conclusion}
  We propose a time-conditioned generative model for video decomposition and prediction. The proposed model enhances the disentanglement between viewpoint and object-centric representations, and additionally adopts GPs for viewpoint modeling, inference and generation. We design experiments to show that the proposed model can: 1) aggregate 3D object-centric information from multiple viewpoints, and as a result, outperforms the state-of-art multi-view models; 2) restore the complete shapes of objects even when completely occluded; and 3) predict the scene images from unknown viewpoints without viewpoint annotations.

\section*{Acknowledgments}
  This work was supported in part by the National Natural Science Foundation of China (No.62176060), STCSM project (No.20511100400), and the Program for Professor of Special Appointment (Eastern Scholar) at Shanghai Institutions of Higher Learning.
\bibliography{gao_704}

\title{Time-Conditioned Generative Modeling of Object-Centric Representations for Video Decomposition and Prediction\\(Supplementary Material)}
\onecolumn 
\setcounter{section}{0}
\section{Details of Training}
\subsection{Derivation of ELBO}
\begin{align}
    \log&\ p_{\boldsymbol{\theta},\boldsymbol{\eta}}(\boldsymbol{x}_{\mathcal{S}} \mid \boldsymbol{t}_{\mathcal{S}}) \geq \log p_{\boldsymbol{\theta},\boldsymbol{\eta}}(\boldsymbol{x}_{\mathcal{S}}\mid \boldsymbol{t}_{\mathcal{S}}) - D_{KL}\big( q_{\boldsymbol{\phi},\boldsymbol{\eta}}( \boldsymbol{\Omega} \mid \boldsymbol{x}_{\mathcal{T}}, \boldsymbol{t}_{\mathcal{S}}) \| p_{\boldsymbol{\theta}}(\boldsymbol{\Omega}\mid \boldsymbol{x}_{\mathcal{S}},\boldsymbol{t}_{\mathcal{S}})\big)\\
    &=\mathbb{E}_{q_{\boldsymbol{\phi},\boldsymbol{\eta}}(\boldsymbol{\Omega} \mid \boldsymbol{x}_{\mathcal{T}}, \boldsymbol{t}_{\mathcal{S}})}\big[\log p_{\boldsymbol{\theta},\boldsymbol{\eta}}(\boldsymbol{x}_{\mathcal{S}}\mid \boldsymbol{t}_{\mathcal{S}})\big] - \mathbb{E}_{q_{\boldsymbol{\phi},\boldsymbol{\eta}}(\boldsymbol{\Omega} \mid \boldsymbol{x}_{\mathcal{T}}, \boldsymbol{t}_{\mathcal{S}})}\big[\log \frac{q_{\boldsymbol{\phi},\boldsymbol{\eta}}(\boldsymbol{\Omega} \mid \boldsymbol{x}_{\mathcal{T}}, \boldsymbol{t}_{\mathcal{S}})}{p_{\boldsymbol{\theta},\boldsymbol{\eta}}(\boldsymbol{\Omega} \mid \boldsymbol{x}_{\mathcal{S}}, \boldsymbol{t}_{\mathcal{S}})} \big] \\
    &=\mathbb{E}_{q_{\boldsymbol{\phi},\boldsymbol{\eta}}(\boldsymbol{\Omega} \mid \boldsymbol{x}_{\mathcal{T}}, \boldsymbol{t}_{\mathcal{S}})}\big[\log \frac{ p_{\boldsymbol{\theta},\boldsymbol{\eta}}(\boldsymbol{x}_{\mathcal{S}},\boldsymbol{\Omega}\mid  \boldsymbol{t}_{\mathcal{S}})}{q_{\boldsymbol{\phi},\boldsymbol{\eta}}(\boldsymbol{\Omega} \mid \boldsymbol{x}_{\mathcal{T}}, \boldsymbol{t}_{\mathcal{S}})}\big] \\
    &=\mathbb{E}_{q_{\boldsymbol{\phi},\boldsymbol{\theta}}(\boldsymbol{\Omega} \mid \boldsymbol{x}_{\mathcal{T}}, \boldsymbol{t}_{\mathcal{S}})}\Big[ \sum_{m=1}^{T} \log p_{\boldsymbol{\theta},\boldsymbol{\eta}}(\boldsymbol{x}_m \mid \boldsymbol{\Omega},t_m)
    + \log \frac{\prod_{m=1}^{T}p_{\boldsymbol{\theta}}(\boldsymbol{\lambda}_m \mid t_m) \prod_{d=1}^D p_{\boldsymbol{\eta}}(\boldsymbol{z}_{1:T,d}^{\text{view}}\mid \boldsymbol{\lambda})}{q_{\boldsymbol{\phi}}(\boldsymbol{\lambda}\mid \boldsymbol{x}_{\mathcal{T}},\boldsymbol{t}_{\mathcal{S}}) \prod_{d=1}^{D} q_{\boldsymbol{\eta}}(\boldsymbol{z}_{\mathcal{Q},d}^{\text{view}}\mid \boldsymbol{z}_{\mathcal{T},d}^{\text{view}},\boldsymbol{\lambda})q_{\boldsymbol{\phi}}(\boldsymbol{z}_{\mathcal{T},d}^{\text{view}}\mid \boldsymbol{x}_{\mathcal{T}},\boldsymbol{t}_{\mathcal{S}}) }\notag\\
    &\quad \quad \quad \quad \quad \quad \quad \quad
    \log \frac{\prod_{k=1}^K p(\nu_k)p(z_k^{\text{pres}}\mid \nu_{k})p(\boldsymbol{z}_k^{\text{obj}})p(\boldsymbol{z}^{\text{bck}})}{\prod_{k=1}^K q_{\boldsymbol{\phi}}(\nu_k \mid \boldsymbol{x}_{\mathcal{T}})q_{\boldsymbol{\phi}}(z_k^{\text{pres}} \mid \boldsymbol{x}_{\mathcal{T}})q_{\boldsymbol{\phi}}(\boldsymbol{z}_k^{\text{obj}} \mid \boldsymbol{x}_{\mathcal{T}})q_{\boldsymbol{\phi}}(\boldsymbol{z}^{\text{bck}} \mid \boldsymbol{x}_{\mathcal{T}})}\Big]\\
    &=\underbrace{\sum\limits_{m\in \mathcal{T}}\mathbb{E}_{q_{\boldsymbol{\phi}}(\boldsymbol{z}_{\mathcal{T}}^{\text{view}} \mid \boldsymbol{x}_{\mathcal{T}}, \boldsymbol{t}_{\mathcal{S}}) q_{\boldsymbol{\phi}}(\boldsymbol{\Omega}^{\backslash\text{view}} \mid \boldsymbol{x}_{\mathcal{T}})} \big[ \log p_{\boldsymbol{\theta},\boldsymbol{\eta}}(\boldsymbol{x}_m \mid \boldsymbol{\Omega}_\mathcal{T},t_m)\big]}_{\text{observation reconstruction loss}}\\
    &\quad + \underbrace{\sum\limits_{m\in \mathcal{Q}}\mathbb{E}_{q_{\boldsymbol{\phi}}(\boldsymbol{\lambda}_{\mathcal{T}} \mid \boldsymbol{x}_{\mathcal{T}}, \boldsymbol{t}_{\mathcal{T}})q_{\boldsymbol{\phi}}(\boldsymbol{\lambda}_{\mathcal{Q}} \mid \boldsymbol{\lambda}_{\mathcal{T}}, \boldsymbol{t}_{\mathcal{Q}})q_{\boldsymbol{\phi}}(\boldsymbol{z}_{\mathcal{T}}^{\text{view}} \mid \boldsymbol{x}_{\mathcal{T}}, \boldsymbol{t}_{\mathcal{S}}) q_{\boldsymbol{\eta}}(\boldsymbol{z}_{\mathcal{Q}}^{\text{view}}\mid \boldsymbol{z}_{\mathcal{T}}^{\text{view}}, \boldsymbol{\lambda}_{\mathcal{S}})q_{\boldsymbol{\phi}}(\boldsymbol{\Omega}^{\backslash\text{view}} \mid \boldsymbol{x}_{\mathcal{T}})} \big[ \log p_{\boldsymbol{\theta},\boldsymbol{\eta}}(\boldsymbol{x}_m \mid \boldsymbol{\Omega},t_m)\big]}_{\text{prediction reconstruction loss}} \\
    &\quad - D_{KL}\big( q_{\boldsymbol{\phi}}(\boldsymbol{\lambda}_{\mathcal{S}} \mid \boldsymbol{x}_{\mathcal{T}}, \boldsymbol{t}_{\mathcal{S}}) \| p_{\boldsymbol{\theta}}(\boldsymbol{\lambda}_\mathcal{S} \mid \boldsymbol{t}_{\mathcal{S}})\big) \\
    &\quad - \mathbb{E}_{q_{\boldsymbol{\phi}}(\boldsymbol{\lambda}_{\mathcal{S}} \mid \boldsymbol{x}_{\mathcal{T}}, \boldsymbol{t}_{\mathcal{S}})} \big[\sum_{d=1}^D D_{KL}  \big( q_{\boldsymbol{\eta}}(\boldsymbol{z}_{\mathcal{Q},d}^{\text{view}} \mid \boldsymbol{z}_{\mathcal{T},d}^{\text{view}}, \boldsymbol{\lambda}_{\mathcal{S}})q_{\boldsymbol{\phi}}(\boldsymbol{z}_{\mathcal{T},d}^{\text{view}}\mid \boldsymbol{x}_{\mathcal{T}},\boldsymbol{t}_{\mathcal{S}}) \| p_{\boldsymbol{\eta}}(\boldsymbol{z}_{\mathcal{S},d}^{\text{view}} \mid \boldsymbol{\lambda}_{\mathcal{S}}) \big) \big] \\
    &\quad - \sum_{k=1}^K D_{KL}(q_{\boldsymbol{\phi}}(\nu_k \mid \boldsymbol{x}_{\mathcal{T}})\| p(\nu_k)) - \sum_{k=1}^K \mathbb{E}_{q_{\boldsymbol{\phi}}(\nu_k \mid \boldsymbol{x}_{\mathcal{T}})}\big[ D_{KL}\big( q_{\boldsymbol{\phi}}(z_k^{\text{pres}}\mid \boldsymbol{x}_{\mathcal{T}})\| p(z_k^{\text{pres}}\mid \nu_{k}) \big)\big] \\
    &\quad - \sum_{k=1}^K D_{KL}\big( q_{\boldsymbol{\phi}}(\boldsymbol{z}_k^{\text{obj}} \mid \boldsymbol{x}_{\mathcal{T}})\| p(\boldsymbol{z}_k^{\text{obj}})\big) - D_{KL}\big( q_{\boldsymbol{\phi}}(\boldsymbol{z}^{\text{bck}} \mid \boldsymbol{x}_{\mathcal{T}})\| p(\boldsymbol{z}^{\text{bck}})\big)
\end{align}
Here $\boldsymbol{\Omega}$ is the simplification of all latent variables, i.e., $\boldsymbol{\Omega} = \big\{ \boldsymbol{z}^{\text{bck}},\boldsymbol{z}^{\text{obj}},\boldsymbol{z}^{\text{pres}},\boldsymbol{\nu}, \boldsymbol{\lambda}_{\mathcal{S}}, \boldsymbol{z}_{\mathcal{S}}^{\text{view}}\big\}$. Let $\boldsymbol{\Omega}_{\mathcal{T}} = \big\{ \boldsymbol{z}^{\text{bck}},\boldsymbol{z}^{\text{obj}},\boldsymbol{z}^{\text{pres}},\boldsymbol{\nu}, \boldsymbol{\lambda}_{\mathcal{T}}, \boldsymbol{z}_{\mathcal{Q}}^{\text{view}}\big\}, \boldsymbol{\Omega}_{\mathcal{T}} = \big\{ \boldsymbol{z}^{\text{bck}},\boldsymbol{z}^{\text{obj}},\boldsymbol{z}^{\text{pres}},\boldsymbol{\nu}, \boldsymbol{\lambda}_{\mathcal{T}}, \boldsymbol{z}_{\mathcal{Q}}^{\text{view}}\big\}$. Now the observation reconstruction loss and prediction reconstruction loss can be respectively expressed as:
\begin{align}
    \mathcal{L}_{\mathcal{T}} &= \sum\limits_{m\in \mathcal{T}}\mathbb{E}_{q_{\boldsymbol{\phi}}(\boldsymbol{\Omega}_{\mathcal{T}} \mid \boldsymbol{x}_{\mathcal{T}}, \boldsymbol{t}_{\mathcal{T}})}\big[ \log p_{\boldsymbol{\theta},\boldsymbol{\eta}}(\boldsymbol{x}_m \mid \boldsymbol{\Omega}_\mathcal{T},t_m)\big] \\
    \mathcal{L}_{\mathcal{Q}} &= \sum\limits_{m\in \mathcal{Q}} \mathbb{E}_{q_{\boldsymbol{\phi}}(\boldsymbol{\Omega}_{\mathcal{T}} \mid \boldsymbol{x}_{\mathcal{T}}, \boldsymbol{t}_{\mathcal{T}})q_{\boldsymbol{\phi,\eta}}(\boldsymbol{\Omega}_{\mathcal{Q}} \mid \boldsymbol{\Omega}_{\mathcal{T}},\boldsymbol{t}_{\mathcal{Q}})} \big[ \log p_{\boldsymbol{\theta,\eta}}(\boldsymbol{x}_m \mid \boldsymbol{\Omega}_{\mathcal{Q}},\boldsymbol{t}_{m})\big]
\end{align}
The loss for each item is calculated as:
\begin{align}
    &\log p_{\boldsymbol{\theta,\eta}}(\boldsymbol{x}_m \mid \boldsymbol{\Omega},t_m) = \frac{1}{2\sigma_x^2} \sum_{n=1}^N \| \boldsymbol{x}_{m,n} - \sum_{k=0}^K \pi_{m,k,n}\boldsymbol{a}_{m,k,n}\|_2^2 + \frac{NC}{2}\log 2\pi \sigma_x^2 \\
    &D_{KL}\big( q_{\phi}(\boldsymbol{\lambda}_{t,d} \mid \boldsymbol{x}_{\mathcal{T}}, \boldsymbol{t}_{\mathcal{T}}) \| p_{\theta}(\boldsymbol{\lambda}_{t,d} \mid \boldsymbol{t}_{\mathcal{T}})\big) = \frac{\|\boldsymbol{\mu}_{t,d}(\boldsymbol{x}_\mathcal{T},\boldsymbol{t}_\mathcal{T}) - \boldsymbol{\mu}_{t,d}(\boldsymbol{t}_{\mathcal{T}}) \|_2^2}{\sigma_{w}^2}\\
    &D_{KL}(q_{\phi}(\nu_k \mid \boldsymbol{x}_{\mathcal{T}})\| p(\nu_k))= \log \frac{\Gamma\left(\tau_{k, 1}+\tau_{k, 2}\right)}{\Gamma\left(\tau_{k, 1}\right) \Gamma\left(\tau_{k, 2}\right)}-\log \frac{\alpha}{K} \\
    &\quad \quad \quad \quad \quad \quad \quad \quad \quad \quad +\left(\tau_{k, 1}-\frac{\alpha}{K}\right) \psi\left(\tau_{k, 1}\right)+\left(\tau_{k, 2}-1\right) \psi\left(\tau_{k, 2}\right)\\
    &\quad \quad \quad \quad \quad \quad \quad \quad \quad \quad -\left(\tau_{k, 1}+\tau_{k, 2}-\frac{\alpha}{K}-1\right) \psi\left(\tau_{k, 1}+\tau_{k, 2}\right) \\
    &\mathbb{E}_{q_{\phi}(\boldsymbol{\nu}_k \mid \boldsymbol{x}_{\mathcal{T}})}\big[ D_{KL}\big( q_{\phi}(z_k^{\text{pres}}\mid \boldsymbol{x}_{\mathcal{T}})\| p(z_k^{\text{pres}}\mid \boldsymbol{\nu}_{k}) \big)\big] = \psi\left(\tau_{k, 1}+\tau_{k, 2}\right)+\kappa_k\left(\log \left(\kappa_k\right)-\psi\left(\tau_{k, 1}\right)\right)\\
    &\quad \quad \quad \quad \quad \quad \quad \quad \quad \quad +\left(1-\kappa_k\right)\left(\log \left(1-\kappa_k\right)-\psi\left(\tau_{k, 2}\right)\right)\\
    & D_{KL}\big( q_{\phi}(\boldsymbol{\lambda}_{t,d} \mid \boldsymbol{x}_{\mathcal{T}}, \boldsymbol{t}_{\mathcal{Q}}) \| p_{\theta}(\boldsymbol{\lambda}_{t,d} \mid \boldsymbol{t}_{\mathcal{Q}})\big) = \frac{\|\boldsymbol{\mu}_{t,d}(\boldsymbol{x}_\mathcal{Q},\boldsymbol{t}_\mathcal{Q}) - \boldsymbol{\mu}_{t,d} \|_2^2}{\sigma_{w}^2}\\
    & D_{KL}\big( q_{\phi}(\boldsymbol{z}_{\text{bck}} \mid \boldsymbol{x}_{\mathcal{T}})\| p(\boldsymbol{z}^{\text{bck}})\big) = \mu^{\text{bck}^2} + \sigma^{\text{bck}^2} - \log \sigma^{\text{bck}^2} -1  \\
    & D_{KL}\big( q_{\phi}(\boldsymbol{z}_k^{\text{obj}} \mid \boldsymbol{x}_{\mathcal{T}})\| p(\boldsymbol{z}_k^{\text{obj}})\big) =  \sum_{i}\big( \mu_{k,i}^{\text{obj}^2} + \sigma_{k,i}^{\text{obj}^2} - \log \sigma_{k,i}^{\text{obj}^2} -1 \big)
\end{align}
\subsection{KL Divergence of Viewpoint Latent Variables}
\begin{align}
    &\mathbb{E}_{q_{\boldsymbol{\phi}}(\boldsymbol{\lambda} \mid \boldsymbol{x}_{\mathcal{T}}, \boldsymbol{t}_{\mathcal{S}})} \big[\sum_{d=1}^D D_{KL}  \big( q_{\boldsymbol{\eta}}(\boldsymbol{z}_{\mathcal{Q},d}^{\text{view}} \mid \boldsymbol{z}_{\mathcal{T},d}^{\text{view}}, \boldsymbol{\lambda})q_{\boldsymbol{\phi}}(\boldsymbol{z}_{\mathcal{T},d}^{\text{view}}\mid \boldsymbol{x}_{\mathcal{T}},\boldsymbol{t}_{\mathcal{S}}) \| p_{\boldsymbol{\eta}}(\boldsymbol{z}_{\mathcal{S},d}^{\text{view}} \mid \boldsymbol{\lambda}) \big) \big]\\
    =&\mathbb{E}_{q_{\boldsymbol{\phi}}(\boldsymbol{\lambda} \mid \boldsymbol{x}_{\mathcal{T}}, \boldsymbol{t}_{\mathcal{S}})}\big[\sum_{d=1}^D \mathbb{E}_{q_{\boldsymbol{\eta}}(\boldsymbol{z}_{\mathcal{Q},d}^{\text{view}} \mid \boldsymbol{z}_{\mathcal{T},d}^{\text{view}}, \boldsymbol{\lambda})q_{\boldsymbol{\phi}}(\boldsymbol{z}_{\mathcal{T},d}^{\text{view}}\mid \boldsymbol{x}_{\mathcal{T}},\boldsymbol{t}_{\mathcal{S}}) }\log \frac{q_{\boldsymbol{\eta}}(\boldsymbol{z}_{\mathcal{Q},d}^{\text{view}} \mid \boldsymbol{z}_{\mathcal{T},d}^{\text{view}}, \boldsymbol{\lambda})q_{\boldsymbol{\phi}}(\boldsymbol{z}_{\mathcal{T},d}^{\text{view}}\mid \boldsymbol{x}_{\mathcal{T}},\boldsymbol{t}_{\mathcal{S}})}{p_{\boldsymbol{\eta}}(\boldsymbol{z}_{\mathcal{S},d}^{\text{view}} \mid \boldsymbol{\lambda})}\big] \\
    =&\mathbb{E}_{q_{\boldsymbol{\phi}}(\boldsymbol{\lambda} \mid \boldsymbol{x}_{\mathcal{T}}, \boldsymbol{t}_{\mathcal{S}})}\big[\sum_{d=1}^D \mathbb{E}_{q_{\boldsymbol{\eta}}(\boldsymbol{z}_{\mathcal{Q},d}^{\text{view}} \mid \boldsymbol{z}_{\mathcal{T},d}^{\text{view}}, \boldsymbol{\lambda})q_{\boldsymbol{\phi}}(\boldsymbol{z}_{\mathcal{T},d}^{\text{view}}\mid \boldsymbol{x}_{\mathcal{T}},\boldsymbol{t}_{\mathcal{S}}) }\log \frac{q_{\boldsymbol{\eta}}(\boldsymbol{z}_{\mathcal{Q},d}^{\text{view}} \mid \boldsymbol{z}_{\mathcal{T},d}^{\text{view}}, \boldsymbol{\lambda})q_{\boldsymbol{\phi}}(\boldsymbol{z}_{\mathcal{T},d}^{\text{view}}\mid \boldsymbol{x}_{\mathcal{T}},\boldsymbol{t}_{\mathcal{S}})}{p_{\boldsymbol{\eta}}(\boldsymbol{z}_{\mathcal{Q},d}^{\text{view}}\mid \boldsymbol{z}_{\mathcal{T},d}^{\text{view}},\boldsymbol{\lambda})p_{\boldsymbol{\eta}}(\boldsymbol{z}_{\mathcal{T},d}^{\text{view}}\mid \boldsymbol{\lambda})}\big] \\
    &\textcolor{gray}{\text{// given $\boldsymbol{\lambda}\sim q_{\boldsymbol{\phi}}(\boldsymbol{\lambda} \mid \boldsymbol{x}_{\mathcal{T}}), \boldsymbol{z}_{\mathcal{T},d}^{\text{view}} \sim q_{\boldsymbol{\phi}}(\boldsymbol{z}_{\mathcal{T},d}^{\text{view}}\mid \boldsymbol{x}_{\mathcal{T}},\boldsymbol{t}_{\mathcal{S}})$, then  $q_{\boldsymbol{\eta}}(\boldsymbol{z}_{\mathcal{Q},d}^{\text{view}} \mid \boldsymbol{z}_{\mathcal{T},d}^{\text{view}}, \boldsymbol{\lambda}) = p_{\boldsymbol{\eta}}(\boldsymbol{z}_{\mathcal{Q},d}^{\text{view}} \mid \boldsymbol{z}_{\mathcal{T},d}^{\text{view}}, \boldsymbol{\lambda})$ }} \\
    =&\mathbb{E}_{q_{\boldsymbol{\phi}}(\boldsymbol{\lambda} \mid \boldsymbol{x}_{\mathcal{T}}, \boldsymbol{t}_{\mathcal{S}})}\big[\sum_{d=1}^D \mathbb{E}_{q_{\boldsymbol{\eta}}(\boldsymbol{z}_{\mathcal{Q},d}^{\text{view}} \mid \boldsymbol{z}_{\mathcal{T},d}^{\text{view}}, \boldsymbol{\lambda})q_{\boldsymbol{\phi}}(\boldsymbol{z}_{\mathcal{T},d}^{\text{view}}\mid \boldsymbol{x}_{\mathcal{T}},\boldsymbol{t}_{\mathcal{S}}) }\log \frac{q_{\boldsymbol{\phi}}(\boldsymbol{z}_{\mathcal{T},d}^{\text{view}}\mid \boldsymbol{x}_{\mathcal{T}},\boldsymbol{t}_{\mathcal{S}})}{p_{\boldsymbol{\eta}}(\boldsymbol{z}_{\mathcal{T},d}^{\text{view}}\mid \boldsymbol{\lambda})}\big] \\
    =&\mathbb{E}_{q_{\boldsymbol{\phi}}(\boldsymbol{\lambda} \mid \boldsymbol{x}_{\mathcal{T}}, \boldsymbol{t}_{\mathcal{S}})}\big[\sum_{d=1}^D \mathbb{E}_{q_{\boldsymbol{\phi}}(\boldsymbol{z}_{\mathcal{T},d}^{\text{view}}\mid \boldsymbol{x}_{\mathcal{T}},\boldsymbol{t}_{\mathcal{S}}) }\log \frac{q_{\boldsymbol{\phi}}(\boldsymbol{z}_{\mathcal{T},d}^{\text{view}}\mid \boldsymbol{x}_{\mathcal{T}},\boldsymbol{t}_{\mathcal{S}})}{p_{\boldsymbol{\eta}}(\boldsymbol{z}_{\mathcal{T},d}^{\text{view}}\mid \boldsymbol{\lambda})}\big] \\
    =&\mathbb{E}_{q_{\boldsymbol{\phi}}(\boldsymbol{\lambda} \mid \boldsymbol{x}_{\mathcal{T}}, \boldsymbol{t}_{\mathcal{S}})}\Big[\sum_{d=1}^D \mathbb{E}_{q_{\boldsymbol{\phi}}(\boldsymbol{z}_{\mathcal{T},d}^{\text{view}}\mid \boldsymbol{x}_{\mathcal{T}},\boldsymbol{t}_{\mathcal{S}}) }\big[ \log q_{\boldsymbol{\phi}}(\boldsymbol{z}_{\mathcal{T},d}^{\text{view}}\mid \boldsymbol{x}_{\mathcal{T}},\boldsymbol{t}_{\mathcal{S}}) - \log p_{\boldsymbol{\eta}}(\boldsymbol{z}_{\mathcal{T},d}^{\text{view}}\mid \boldsymbol{\lambda}) \big] \Big]
\end{align}
where $p_{\boldsymbol{\eta}}(\boldsymbol{z}_{\mathcal{T},d}^{\text{view}}\mid \boldsymbol{\lambda}) = \frac{1}{(2\pi)^{\frac{|\mathcal{T}|}{2}}} \frac{1}{|\boldsymbol{C}_{\boldsymbol{\eta}}(\mathcal{T})|^{\frac{1}{2}}} \exp \big\{ -\frac{1}{2}(\boldsymbol{z}_{\mathcal{T},d}^{\text{view}})^{\top} \boldsymbol{C}_{\boldsymbol{\eta}}^{-1}(\mathcal{T})\boldsymbol{z}_{\mathcal{T},d}^{\text{view}}\big\}$, see , in Eq \ref{eq:c_compute}.
\section{Details of Generation}
\label{sec:notation}
Suppose each \textit{static} visual scene is composed of $T$ consecutive viewpoints and each instance (video) is independently and identically distributed. Hereby, for simple description, the following will take a single video instance as an example. Let $N$ and $C$ respectively denote the number of pixels and channels in each frame $\boldsymbol{x}_t$ $(1\leq t \leq T)$ of the video, and $K$ denotes the maximum number of objects that appear in the visual scene. Each pixel $\boldsymbol{x}_{t,n}$ of each frame is a weighted summation of $K + 1$ components at
that pixel, with $K$ describing the objects and one describing the background. $K+1$ layers of the image correspond to the $K+1$ components composed of $N$ pixels. In the compositional modeling, layers of each frame $\boldsymbol{x}_t$ consist of pixel-wise weights $\boldsymbol{\pi}_t \in \mathbb{R}^{(K+1) \times N}$ and the expected pixel-wise RGB value  $\boldsymbol{a}_t \in \mathbb{R}^{(K+1) \times N  \times C}$. Both are generated by some representations (including latent variables, deterministic values, neural networks, etc.). We will express them below.

\textbf{View-independent representations.} we define a set of object-centric latent variables of $K+1$ entities from $T$ viewpoints that describes the 3D visual scene, including $\boldsymbol{z}^{\text{obj}},\boldsymbol{z}^{\text{bck}},\boldsymbol{z}^{\text{pres}},\boldsymbol{\nu}$.
\begin{itemize}[leftmargin=*]
    \item $\boldsymbol{z}^{\text{obj}} =\boldsymbol{z}_{1:K}^{\text{obj}}$ describes the 3D view-independent representations of objects. View-independent means the physical attributes of objects (such as shape, appearance, etc.) keep constant under different multiple viewpoints. $\boldsymbol{z}_k^{\text{obj}}$ $(1\leq k \leq K)$ is independently and identically distributed.
    \item $\boldsymbol{z}^{\text{bck}}$ denotes the latent representation of the background appearance. We need not represent the shape of the background because the corresponding complete shape is 1.
    \item $\boldsymbol{z}^{\text{pres}}=\boldsymbol{z}_{1:K}^{\text{pres}},\boldsymbol{\nu}=\boldsymbol{\nu}_{1:K}$ denote the latent variables that indicate the presence of objects. The advantage of using the latents is that the uncertain number of objects in different visual scenes can be added up. $z_k^{\text{pres}} (1\leq k\leq K)$ denotes whether object $k$ appears in a visual scene, following a Bernoulli distribution. The parameter of the distribution is controlled by latent variable $\nu_k$ that follows the conjugate prior, i.e. $z_k^{\text{pres}}\sim \text{Bernoulli}(\nu_k),\nu_k \sim \text{Beta}(\alpha/ K, 1)$, where $\alpha$ is the hyperparameter, $K$ denotes the object numbers.
\end{itemize}
\textbf{View-dependent representations.} Different from previous works \cite[]{li2020learning,chen2021roots}, we learn view representations through finding the relationship between frames, rather than directly leveraging viewpoint labels. Meanwhile, the view correlation based on temporal modeling can motivate the model to predict the novel scenes unseen given any time. The related view-dependent representations include $\boldsymbol{\lambda},\boldsymbol{z}^{\text{view}}$.
\begin{itemize}[leftmargin=*]
    \item $\boldsymbol{\lambda}\in \mathbb{R}^{T\times D \times D_{\lambda}}$ represents the spatial latent variable that reflects the position characteristics of the camera under different frames, where $D$ is the dimension of the view latent (i.e. $\boldsymbol{z}^{\text{view}}$) corresponding to the frame. $D_{\lambda}$ is the dimension of the spatial representation that influences the meanings of each dimension in the view latent. $\boldsymbol{\lambda}_t$ potentially affects the change of viewpoints ((e.g. the distance, height, rotation of the camera) at time $t$. $\boldsymbol{\lambda}_{t,d}$ $(1\leq t \leq T, 1 \leq d \leq D)$ is distributed in a linear subspace.
    \item $\boldsymbol{z}^{\text{view}} = \boldsymbol{z}_{1:T}^{\text{view}} \in \mathbb{R}^{T\times D}$ denotes the view latent variables. Videos perform in a way that the closer the distance of two frames, the smaller the difference between the corresponding viewpoint information, and the bigger on the contrary. To build the correlation, we define $\boldsymbol{z}^{\text{view}}$ as a Gaussian process (GP) prior parameterized by the spatial latent variable $\boldsymbol{\lambda}$. 
\end{itemize}
\textbf{Additional Notations.} In addition to the latent variables defined above, we also need some non-latent notations to generate $T$ frames, including $\boldsymbol{s}^{\text{shp}}$, $\boldsymbol{o}$, $\boldsymbol{\pi}$, $\boldsymbol{a}$.
\begin{itemize}[leftmargin=*]
    \item $\boldsymbol{s}^{\text{shp}}\in [0,1]^{T \times K \times N}$ describes the complete shape of different objects at different time $t$. $\boldsymbol{s}_{t,\cdot,k}^{\text{shp}}$ ($\cdot$ represents all indexes are selected) represents the complete shape of the $k$th object in the 2D image corresponding to the $t$th frame. The range of [0, 1] guarantees the subsequent rationality of processing the occlusion. Since the complete shape of the background is a  constant of 1, values of $\boldsymbol{s}_{t,\cdot,k}^{\text{shp}}$ can be computed by the neural network $g_{\text{shp}}$ with $\boldsymbol{z}_k^{\text{obj}}$ and $\boldsymbol{z}_t^{\text{view}}$ as inputs followed by a sigmoid activation and $\boldsymbol{z}^{\text{bck}}$ need not participate in the computation.
    \item $\boldsymbol{o} \in \mathbb{R}^{T \times K}$ describes the occlusion order of different objects in the different frame. $o_{t,k}$ denotes the order of the $k$th object under the projected 2D image at the $t$th frame. $\boldsymbol{o}$ is obtained by the neural network $g_{\text{ord}}$ with $\boldsymbol{z}_k^{\text{obj}}$ and $\boldsymbol{z}_t^{\text{view}}$ as inputs since the occlusion order of the same object varies at different viewpoints. 
    \item $\boldsymbol{\pi} \in [0,1]^{T \times (K+1) \times N}$ represents the pixel-wise weights of each layer, i.e. geometrically represents the observed shape of each object. The $\boldsymbol{\pi}$ here is different from $\boldsymbol{s}^{\text{shp}}$ in that the shape of an object may be partially observed or completely invisible due to partial or complete occlusion. $K+1$ observed shapes of the $n$th pixel at frame $t$ satisfy $\sum \nolimits_{k=0}^{K} \pi_{t,k,n} = 1$  $(1\leq t\leq T, 1 \leq n \leq N)$.
    \item $\boldsymbol{a}\in \mathbb{R}^{T\times (K+1) \times N \times C}$ describes the complete appearance of all entities (objects or backgrounds). $\boldsymbol{a}_{t,k,n}$ is numerically equivalent to the expected RGB value of component $k$ at the $n$th pixel of the $t$th frame. The background appearance $\boldsymbol{a}_{t,k}$ $(k=0)$ is achieved by the neural network $g_{\text{apc}}^{\text{bck}}$ with $\boldsymbol{z}_t^{\text{view}}$ and $\boldsymbol{z}^{\text{bck}}$ as inputs, meanwhile the $k$th object appearance $\boldsymbol{a}_{t,k}$ $(1\leq k \leq K)$ is achieved by another neural network $g_{\text{apc}}^{\text{obj}}$ with $\boldsymbol{z}_t^{\text{view}}$ and $\boldsymbol{z}_k^{\text{obj}}$ as inputs.
\end{itemize}
\textbf{Likelihood Function.} After generating the observed shapes $\boldsymbol{\pi}$ and appearance $\boldsymbol{a}$ of each layer, we can use a weighted summation of each layer to reconstruct the image. Its likelihood is expressed as:
\begin{align}
    \label{eq:likelihood function}
    \log p(\boldsymbol{x}_{1:T} \mid \boldsymbol{\pi}, \boldsymbol{a}) = &\sum_{t=1}^T \sum_{n=1}^N \log \mathcal{N}(\underbrace{\pi_{t,0,n} \cdot \boldsymbol{a}_{t,0,n}}_{\text{Background}} +\sum_{k=1}^K \underbrace{\pi_{t,k,n}\cdot \boldsymbol{a}_{t,k,n}}_{\text{Objects}}, \hat{\sigma}_{x}^2\boldsymbol{I})
\end{align}

where $\hat{\sigma}_{x}$ is the hyperparameter. The style of the likelihood function is similar to Slot Attention \cite[]{locatello2020object} in order to improve the reconstruction.



\section{Details of Inference}
\subsection{Approximation of Predicted Spatial Latent Variables}
$\boldsymbol{\lambda}_{t,d} \in \mathbb{R}^{D_{\lambda}} (1\leq t \leq |\mathcal{S}|, 1\leq d \leq D)$ denotes the spatial latent representations corresponding to $\boldsymbol{z}_{t,d}^{\text{view}}$. In the generative process, $\boldsymbol{\lambda}_{t,d}$ is distributed in a linear subspace, i.e.,
\begin{align}
    \boldsymbol{\lambda}_{t,d} \sim \mathcal{N}(\boldsymbol{A}\boldsymbol{w}_t, \sigma_{w}^2\boldsymbol{I}), \boldsymbol{A} \in \mathbb{R}^{D_\lambda \times |\boldsymbol{w}|}
\end{align}
where $\boldsymbol{A}$ and $\sigma_w^2$ are the hyperparameters. For the posterior of $\boldsymbol{\lambda}$, we can simply define the distribution on $\boldsymbol{\lambda}_{\mathcal{T}}$ that satisfies the linear distribution:
\begin{align}
    q(\boldsymbol{\lambda}_{t,d} \mid \boldsymbol{x}_t,\boldsymbol{w}_t) \sim \mathcal{N}(\boldsymbol{\mu}_d(\boldsymbol{x}_t, \boldsymbol{w}_t), \sigma_{w}^2\boldsymbol{I}), \quad t \in \mathcal{T}, 1\leq d \leq D
\end{align}
$q(\boldsymbol{\lambda}_{t,d} \mid \boldsymbol{x}_{\mathcal{T}}, \boldsymbol{w}_{\mathcal{S}})$ for $t\in \mathcal{Q}$ is difficult.  We apply the Least Square Error to find the optimal mean curve that satisfies a linear relationship w.r.t. $\boldsymbol{w}_t$:
\begin{align}
    \hat{\boldsymbol{A}}_d^{*} = \underset{\hat{\boldsymbol{A}}_d}{\arg \min } \Big\|  \boldsymbol{\Phi}_d -\boldsymbol{W}_{\mathcal{T}} \hat{\boldsymbol{A}}_d^{\top} \Big\|_2^2 \quad, \hat{\boldsymbol{A}}_d\in \mathbb{R}^{D_{\boldsymbol{\lambda}} \times |\boldsymbol{w}|}
\end{align}
where $\boldsymbol{\Phi}_d = \big[ \boldsymbol{\mu}_{1,d}, ..., \boldsymbol{\mu}_{|\mathcal{T}|,d}\big]^{\top}\in \mathbb{R}^{|\mathcal{T}| \times D_{\lambda}}$, $\boldsymbol{W}_{\mathcal{T}} = \big[ \boldsymbol{w}_1,...,\boldsymbol{w}_{|\mathcal{T}|}\big]^{\top}\in \mathbb{R}^{|\mathcal{T}|\times |\boldsymbol{w}|}$. $\hat{\boldsymbol{A}}_d^{*}$ can be analytically solved and the optimal $\hat{\boldsymbol{A}}_d^{*}$ is described as:
\begin{align}
\label{eq:lambda_solve}
\hat{\boldsymbol{A}_d^{*}} = \boldsymbol{\Phi}_d^{\top}\boldsymbol{W}_{\mathcal{T}}(\boldsymbol{W}_{\mathcal{T}}^{\top} \boldsymbol{W}_{\mathcal{T}})^{-1}
\end{align}
Then $q(\boldsymbol{\lambda}_{t,d} \mid \boldsymbol{x}_{\mathcal{T}}, \boldsymbol{w}_{\mathcal{S}})$ for $t\in \mathcal{Q}$ can be approximated as:
\begin{align}
    q(\boldsymbol{\lambda}_{t,d} \mid \boldsymbol{x}_{\mathcal{T}}, \boldsymbol{w}_{\mathcal{S}}) = \mathcal{N}(\hat{\boldsymbol{A}_d^{*}}\boldsymbol{w}_t, \sigma_{w}^2\boldsymbol{I})
\end{align}
\subsection{Gaussian Processes and Inference of predicted view latent representaions}
if the variable $\boldsymbol{z}_{\mathcal{S}} \in \mathbb{R}^{|\mathcal{S}|}$ satisfies the Gaussian Processes (GPs): 
\begin{align}
    p(\boldsymbol{z}_{S}\mid \boldsymbol{\lambda}) \sim N \Big( \boldsymbol{0} , \left[\begin{array}{ccc}
\kappa_{\eta}\left(\boldsymbol{\lambda}_{1}, \boldsymbol{\lambda}_{1}\right) & \cdots & \kappa_{\eta}\left(\boldsymbol{\lambda}_{1}, \boldsymbol{\lambda}_{|\mathcal{S}|}\right) \\
\vdots & \ddots & \vdots \\
\kappa_{\eta}\left(\boldsymbol{\lambda}_{|\mathcal{S}|}, \boldsymbol{\lambda}_{1}\right) & \cdots & \kappa_{\eta}\left(\boldsymbol{\lambda}_{|\mathcal{S}|}, \boldsymbol{\lambda}_{|\mathcal{S}|}\right)
\end{array}\right]
 \Big)
\end{align}
To simplify the analysis, we randomly divides the covariance matrix to the $\boldsymbol{z}_\mathcal{Q}$-dependent sub-matrix and  $\boldsymbol{z}_\mathcal{Q}$-independent sub-matrix (aggregate different subsets together by translation). 
\begin{align}
    \label{eq:block}
        \left[\begin{array}{ccc}
    \kappa_{\eta}\left(\boldsymbol{\lambda}_{1}, \boldsymbol{\lambda}_{1}\right) & \cdots & \kappa_{\eta}\left(\boldsymbol{\lambda}_{1}, \boldsymbol{\lambda}_{|\mathcal{S}|}\right) \\
    \vdots & \ddots & \vdots \\
    \kappa_{\eta}\left(\boldsymbol{\lambda}_{|\mathcal{S}|}, \boldsymbol{\lambda}_{1}\right) & \cdots & \kappa_{\eta}\left(\boldsymbol{\lambda}_{|\mathcal{S}|}, \boldsymbol{\lambda}_{|\mathcal{S}|}\right)
    \end{array}\right] =  \left[\begin{array}{cc}
    \boldsymbol{C}_{\eta}(\mathcal{T}) & \boldsymbol{R}_{\eta}(\mathcal{T}, \mathcal{Q}) \\
    \\
    \boldsymbol{R}_{\eta}(\mathcal{Q}, \mathcal{T}) & \boldsymbol{C}_{\eta}(\mathcal{Q})
    \end{array}\right]
\end{align}
where 
\begin{align}
    \label{eq:c_compute}
    \boldsymbol{C}_{\eta}(\mathcal{H}) &= \left[\begin{array}{ccc}
    \kappa_{\eta}\left(\boldsymbol{\lambda}_{i}, \boldsymbol{\lambda}_{i}\right) & \cdots & \kappa_{\eta}\left(\boldsymbol{\lambda}_{i}, \boldsymbol{\lambda}_{i+j}\right) \\
    \vdots & \ddots & \vdots \\
    \kappa_{\eta}\left(\boldsymbol{\lambda}_{i+j}, \boldsymbol{\lambda}_{i}\right) & \cdots & \kappa_{\eta}\left(\boldsymbol{\lambda}_{i+j}, \boldsymbol{\lambda}_{i+j}\right)
    \end{array}\right], \\
    \boldsymbol{R}_{\eta}(\mathcal{X},\mathcal{Y}) &= \left[\begin{array}{ccc}
        \kappa_{\eta}\left(\boldsymbol{\lambda}_{u}, \boldsymbol{\lambda}_{v}\right) & \cdots & \kappa_{\eta}\left(\boldsymbol{\lambda}_{u}, \boldsymbol{\lambda}_{v+n}\right) \\
        \vdots & \ddots & \vdots \\
        \kappa_{\eta}\left(\boldsymbol{\lambda}_{u+m}, \boldsymbol{\lambda}_{v}\right) & \cdots & \kappa_{\eta}\left(\boldsymbol{\lambda}_{u+m}, \boldsymbol{\lambda}_{v+n}\right)
        \end{array}\right]
\end{align}
where $i\sim i+j \in \mathcal{H}, \mathcal{H}= \big\{ \mathcal{T},\mathcal{Q}\big\}; u\sim u+m \in \mathcal{X}, v\sim v+n \in \mathcal{Y}, \mathcal{X}\neq \mathcal{Y}, \mathcal{X},\mathcal{Y} \in \big\{ \mathcal{T},\mathcal{Q}\big\}$.

given the observation set $\boldsymbol{z}_{\mathcal{T}}$ and $\boldsymbol{\lambda}_{\mathcal{S}}$, $p(\boldsymbol{z}_{\mathcal{Q}}\mid \boldsymbol{z}_{\mathcal{T}},\boldsymbol{\lambda})$ can be calculated analytically using properties of the multivariate Gaussian distribution \cite[]{bishop2006pattern}: 
\begin{align}
    \label{eq:gp}
    p(\boldsymbol{z}_{\mathcal{Q}} \mid \boldsymbol{z}_{\mathcal{T}}, \boldsymbol{\lambda}_{\mathcal{}}) &= \mathcal{N}\big(\boldsymbol{\mu}_{\boldsymbol{\eta}}(\boldsymbol{z}_{\mathcal{T}}, \boldsymbol{\lambda}),\boldsymbol{\Sigma}_{\boldsymbol{\eta}}(\boldsymbol{z}_{\mathcal{T}}, \boldsymbol{\lambda})\big) \\
    \boldsymbol{\mu}_{\boldsymbol{\eta}}(\boldsymbol{z}_{\mathcal{T}},\boldsymbol{\lambda})&=\boldsymbol{R}_{\boldsymbol{\eta}}(\mathcal{Q},\mathcal{T})\boldsymbol{C}_{\boldsymbol{\eta}}^{-1}(\mathcal{T})\boldsymbol{z}_{\mathcal{T}} \\
    \boldsymbol{\Sigma}_{\boldsymbol{\eta}}(\boldsymbol{z}_{\mathcal{T}}, \boldsymbol{\lambda})&= \boldsymbol{C}_{\boldsymbol{\eta}}(\mathcal{Q}) - \boldsymbol{R}_{\boldsymbol{\eta}}(\mathcal{Q},\mathcal{T})\boldsymbol{C}_{\boldsymbol{\eta}}^{-1}(\mathcal{T})\boldsymbol{R}_{\boldsymbol{\eta}}(\mathcal{Q},\mathcal{T})^{\top}
\end{align}
where $\boldsymbol{C}_{\boldsymbol{\eta}}(\mathcal{T})\in \mathbb{R}^{|\mathcal{T}| \times |\mathcal{T}|},\boldsymbol{R}_{\boldsymbol{\eta}}(\mathcal{Q},\mathcal{T})\in \mathbb{R}^{|\mathcal{Q}| \times |\mathcal{T}|}, \boldsymbol{C}_{\boldsymbol{\eta}}(\mathcal{Q})\in \mathbb{R}^{|\mathcal{Q}| \times |\mathcal{Q}|}$.

According to the derivation above, $q(\boldsymbol{z}_{\mathcal{Q},d}^{\text{view}}\mid \boldsymbol{z}_{\mathcal{T},d}^{\text{view}},\boldsymbol{\lambda})$ can be analytically rsampled from Eq \ref{eq:gp}, i.e. 
\begin{align}
    \boldsymbol{z}_{\mathcal{Q},d}^{\text{view}} &= \boldsymbol{\mu}_{\boldsymbol{\eta}}(\boldsymbol{z}_{\mathcal{T},d}^{\text{view}},\boldsymbol{\lambda}) + \boldsymbol{\Sigma}_{\boldsymbol{\eta}}^{\frac{1}{2}}(\boldsymbol{z}_{\mathcal{T},d}^{\text{view}},\boldsymbol{\lambda})\boldsymbol{\epsilon},\quad  \boldsymbol{\epsilon}\in \mathcal{N}(\boldsymbol{0},\boldsymbol{I}) \\
    \boldsymbol{z}_{\mathcal{Q}}^{\text{view}} &= \text{concatenate}(\boldsymbol{z}_{\mathcal{T},1}^{\text{view}},\boldsymbol{z}_{\mathcal{T},2}^{\text{view}}, \cdots, \boldsymbol{z}_{\mathcal{T},D}^{\text{view}}, \text{axis}=\text{``view dim''}),\quad \boldsymbol{z}_{\mathcal{Q}}^{\text{view}}\in \mathbb{R}^{|\mathcal{Q}| \times D}
\end{align}
\subsection{Mathematical Form of Inference}

We detail the algorithm of inference in this section. algorithm \ref{alg:inference} describes the whole mathematical form. It‘s worth mentioning that the view-independent latent variables and view-independent latent variables are inferred based on different nerual networks. More specifically speaking, the feature for view-dependent latent variables is extracted by the neural netowork $f_{\text{view}}$ and then the feature will enter into the Transformer to obtain the $\boldsymbol{\lambda}_{\mathcal{T}}$ and $\boldsymbol{z}_{\mathcal{T}}^{\text{view}}$. The feature for view-independent latent variables is extracted by the neural network $f_{\text{sa}}$ and enters into the sequential extension of Slot Attention \cite[]{locatello2020object}. During the interation in Slot Attention, the view feature $\boldsymbol{y}_{\mathcal{T}}^{\text{view}}$ from the Transoformer will enter into the Slot Attention module, and then concatenate with $\boldsymbol{y}^{\text{attr}}$ initialized with the Gaussian distribution one by one. Note that $\boldsymbol{y}_{\mathcal{T}}^{\text{view}}$ will not be updated during the iteration. different from $\boldsymbol{y}_{\mathcal{T}}^{\text{view}}$, the module will execute the temporal mean of $\boldsymbol{y}^{\text{attr}}$ at each iteration after the cross-attention.
\section{Datasets}
The datasets (CLEVR-SIMPLE, CLEVR-COMPLEX, SHOP-SIMPLE, SHOP-COMPLEX) used in this paper are modified based on the official code of CLEVR \cite[]{johnson2017clevr} and SHOP \cite[]{nazarczuk2020shop}. 
More specifically speaking, we have made some improvements to the official code of CLEVR dataset and SHOP dataset, that is, polar coordinates are used to assign a shot position $(x_t,y_t,z_t)$ to each frame of the video. In the polar coordinates, $\rho$ $(\rho>0)$ represents the radius of the object in a 3D sphere, $\phi$ $(0\leq \phi \leq \frac{\pi}{2})$ describes the angle between the object and the $z$ positve half axis, and $\theta$ $(0\leq \theta \leq 2\pi)$ describes the angle between the object and the $xy$ axis.  The function of camera coordinates $(x_t,y_t,z_t)$ with respect to time t can be described as:
\begin{align}
    x &= \rho\sin \phi \cos \theta \notag \\
    y &= \rho\sin \phi \sin \theta \notag \\
    z &= \rho\cos \phi \notag
\end{align}
\begin{algorithm}[H]
    \begin{algorithmic}
    \caption{Inference of Latent Variables}
    \label{alg:inference}
    \STATE {\bfseries Requires:} observed images $\boldsymbol{x}_{\mathcal{T}}$, timesteps $\boldsymbol{t}_{\mathcal{S}}=(\boldsymbol{t}_{\mathcal{T}},\boldsymbol{t}_{\mathcal{Q}})$, maximum iterations $M_s$.
    \STATE \textcolor{gray}{// \text{extract the feature $\boldsymbol{y}_{\mathcal{T}}^{\text{feat}}\in \mathbb{R}^{|O|\times L \times C}$},$L$ is the product of the height and width corresponding to the feature map}
    \STATE \textcolor{gray}{// $\boldsymbol{y}_{\mathcal{T}}^{\text{sa}} \in \mathbb{R}^{|\mathcal{T}| \times N \times D^{\prime}}$ is another feature map with the neural network $f_{\text{sa}}$}
    \STATE $\boldsymbol{y}_{\mathcal{T}}^{\text{feat}} = f_{\text{feat}}(\boldsymbol{x}_{\mathcal{T}}), \boldsymbol{y}_\mathcal{T}^{\text{sa}} = f_{\text{sa}}(\boldsymbol{x}_\mathcal{T}),[\boldsymbol{y}_1^{\text{sa}},\boldsymbol{y}_2^{\text{sa}},...,\boldsymbol{y}_{|\mathcal{T}|}^{\text{sa}}]=\text{split}(\boldsymbol{y}_\mathcal{T}^{\text{sa}},\text{axis}=0)$
    \STATE $\boldsymbol{y}_{O}^{\text{feat}} = \text{MultiHeadSelfAttention}\big( \text{3DPositionEmbedding}(\boldsymbol{y}_{O}^{\text{feat}})\big)$
    \STATE $\boldsymbol{y}_\mathcal{T}^{\text{view}} = \text{mean}(\boldsymbol{y}_{\mathcal{T}}^{\text{feat}},\text{axis} = 1); \quad [\boldsymbol{y}_1^{\text{view}},\boldsymbol{y}_2^{\text{view}},...,\boldsymbol{y}_{|\mathcal{T}|}^{\text{view}}] = \text{split}(\boldsymbol{y}_\mathcal{T}^{\text{view}} ,\text{axis} = 1)$
    \STATE \textcolor{gray}{// do the spatial mean on $\boldsymbol{y}_{\mathcal{T}}^{\text{feat}}$, and then encode to the posterior of $\boldsymbol{\lambda}_{\mathcal{S}} = (\boldsymbol{\lambda}_{\mathcal{T}},\boldsymbol{\lambda}_{\mathcal{Q}})$, where $\boldsymbol{\mu}_{\mathcal{S}} \in \mathbb{R}^{|\mathcal{S}|\times D \times D_{\lambda}}$}
    \STATE $\boldsymbol{y}_{\mathcal{T}}^{\text{feat}} = \text{Downsample}(\text{MultiHeadSelfAttention}(\boldsymbol{y}_{\mathcal{T}}^{\text{feat}}))$
    \STATE $\boldsymbol{y}_{\mathcal{T}}^{\lambda} = \text{mean}(\boldsymbol{y}_{\mathcal{T}}^{\text{feat}},\text{axis} = 1)$
    \STATE $\boldsymbol{w}_{\mathcal{S}} = \text{TimestepEncoding}(\boldsymbol{t}_{\mathcal{S}}), \boldsymbol{w}_{\mathcal{S}}=(\boldsymbol{w}_{\mathcal{T}},\boldsymbol{w}_{\mathcal{Q}})$
    \STATE \textcolor{gray}{// $\hat{\boldsymbol{A}}_d^{*} (1\leq d \leq D)$ can be obtained by eq \ref{eq:lambda_solve}}
    \STATE $\boldsymbol{\mu}_{\mathcal{T}} = f_{\phi}^{\lambda}(\boldsymbol{y}_{\mathcal{T}}^{\lambda},\boldsymbol{w}_{\mathcal{T}}),\boldsymbol{\mu}_{\mathcal{Q}} = \text{concatenate}(\big[ \hat{\boldsymbol{A}}_1^{*}\boldsymbol{w}_{\mathcal{\mathcal{Q}}},...,\hat{\boldsymbol{A}}_D^{*}\boldsymbol{w}_{\mathcal{Q}}\big],\text{axis}=\text{``view dim''}),
    \boldsymbol{\mu}_{S}=[\boldsymbol{\mu}_{\mathcal{T}},\boldsymbol{\mu}_{\mathcal{Q}}]$
    \STATE $\boldsymbol{\lambda}_{S} \sim N(\boldsymbol{\mu}_{S},\sigma_{\boldsymbol{w}}^2 \boldsymbol{I})$
    \STATE $\boldsymbol{y}_k^{\text{obj}} \sim N(\hat{\boldsymbol{\mu}}^{\text{obj}} ,\hat{\sigma}^{\text{obj}}\boldsymbol{I}), \quad \forall 1\leq k \leq K$
    \STATE $\boldsymbol{y}^{\text{bck}} \sim N(\hat{\boldsymbol{\mu}}^{\text{bck}} ,\hat{\sigma}^{\text{bck}}\boldsymbol{I})$
    \STATE $\boldsymbol{y}_{1:K+1}^{\text{attr}} = [\boldsymbol{y}_{1:K}^{\text{obj}},\boldsymbol{y}^{\text{bck}}], $
    \STATE \textcolor{gray}{// do the iteration of sequential Slot Attention}
    \STATE \textbf{for } $s=1$ \textbf{to } $M$ \textbf{do } $\big\{ \forall t\in \mathcal{T}, \forall 1 \leq k \leq K+1  \big\}$ 
    \STATE $\quad \boldsymbol{y}_{t,k}^{\text{full}} = [\boldsymbol{y}_t^{\text{view}},\boldsymbol{y}_k^{\text{attr}}]$
    \STATE $\quad \boldsymbol{a}_{t} = \underset{K+1}{\text{softmax}} \Big( \frac{k(\boldsymbol{y}_t^{\text{sa}}) \cdot q(\boldsymbol{y}_{t,1:K+1}^{\text{full}})^{\top}}{\sqrt{D_f}}\Big)\in \mathbb{R}^{N\times K}$ \\
    \STATE $\quad \boldsymbol{u}_t = \sum_{N} \underset{N}{\text{softmax}}\Big( \log \boldsymbol{a}_{t,n}\Big) \cdot v(\boldsymbol{y}_{t,n}^{\text{sa}}) \in \mathbb{R}^{K \times D_f}$
    \STATE $\quad \hat{\boldsymbol{y}}^{\text{full}}_{t,k} = \text{GRU}(\boldsymbol{y}^{\text{full}}_{t,k},\boldsymbol{u}_{t,k}), \quad \big[ \hat{\boldsymbol{y}}_{t,k}^{\text{view}},\hat{\boldsymbol{y}}_{t,k}^{\text{attr}} \big] \stackrel{\text{split}}{\leftarrow} \hat{\boldsymbol{y}}^{\text{full}}_{t,k}$
    \STATE $\quad \boldsymbol{y}_{k}^{\text{attr}} = {\text{mean}}_{|\mathcal{T}|}\Big(\hat{\boldsymbol{y}}_{1:|\mathcal{T}|,k}^{\text{attr}}\Big)$
    \STATE \textbf{end for}
    \STATE $[\boldsymbol{y}_1^{\text{obj}}, ..., \boldsymbol{y}_K^{\text{obj}}, \boldsymbol{y}^{\text{bck}}]= \text{split} (\boldsymbol{y}^{\text{attr}},\text{axis} = 0)$
    \STATE \textcolor{gray}{// independently and identically sample $\boldsymbol{z}_t^{\text{view}}$ for $1\leq t \leq |\mathcal{T}|$, then infer $\boldsymbol{z}_\mathcal{Q}^{\text{view}}$ for $1\leq d \leq D$} 
    \STATE $\boldsymbol{\mu}_t^{\text{view}}, \boldsymbol{\sigma}_t^{\text{view}} = f_{\phi}^{\text{view}}(\boldsymbol{y}_t^{\text{view}})$
    \STATE $\boldsymbol{z}_{\mathcal{T}}^{\text{view}} \sim
    \mathcal{N}(\boldsymbol{\mu}_\mathcal{T}^{\text{view}},\text{diag}(\boldsymbol{\sigma}_{\mathcal{T}}^{\text{view}})^2)$
    \STATE \textcolor{gray}{// \text{$\boldsymbol{\mu}_{\eta}$ and $\boldsymbol{\Sigma}_{\eta}$ can be analytically computed, see eq \ref{eq:gp}}}
    \STATE $\boldsymbol{z}_{\mathcal{Q}, d}^{\text {view }}=\boldsymbol{\mu}_{\eta}\left(\boldsymbol{z}_{\mathcal{T}, d}^{\text {view }}, \boldsymbol{\lambda}_\mathcal{S}\right)+\boldsymbol{\Sigma}_{\eta}^{\frac{1}{2}}\left(\boldsymbol{z}_{\mathcal{T}, d}^{\text {view }}, \boldsymbol{\lambda}_\mathcal{S}\right)\boldsymbol{\epsilon}, \quad \boldsymbol{\epsilon}\sim \mathcal{N}(\boldsymbol{0},\boldsymbol{I})$
    \STATE $\boldsymbol{z}_{\mathcal{Q}}^{\text{view}}=\text{concatenate}(\boldsymbol{z}_{\mathcal{Q},1}^{\text{view}},\boldsymbol{z}_{\mathcal{Q},2}^{\text{view}},...,\boldsymbol{z}_{\mathcal{Q},D}^{\text{view}}, \text{axis}=\text{``view dim''})$
    \STATE $\boldsymbol{z}_{S}^{\text{view}} = [\boldsymbol{z}_{\mathcal{T}}^{\text{view}},\boldsymbol{z}_{\mathcal{Q}}^{\text{view}}]$
    \STATE \textcolor{gray}{\text{// infer the view-independent latent variables for $1 \leq k \leq K$}}
    \STATE $\boldsymbol{\mu}^{\text{bck}},\boldsymbol{\sigma}^{\text{bck}} = f_{\phi}^{\text{bck}}(\boldsymbol{y}^{\text{bck}})$
    \STATE $\boldsymbol{\mu}_k^{\text{obj}},\boldsymbol{\sigma}_k^{\text{obj}} ,\boldsymbol{\tau}_k, \kappa_k= f_{\phi}^{\text{obj}}(\boldsymbol{y}_k^{\text{obj}})$
    \STATE \textbf{return } $\boldsymbol{\lambda}_{\mathcal{S}},\boldsymbol{z}_{\mathcal{S}}^{\text{view}},\boldsymbol{\mu}^{\text{bck}},\boldsymbol{\sigma}^{\text{bck}},\boldsymbol{\mu}_{1:K}^{\text{obj}},\boldsymbol{\sigma}_{1:K}^{\text{obj}} ,\boldsymbol{\tau}_{1;K}, \boldsymbol{\kappa}_{1:K}$
\end{algorithmic}
\end{algorithm}
When constructing the dataset, the polar coordinate configuration corresponding to the camera position of each scene (10 frames) is $\rho \sim U(\rho_{\text{min}},\rho_{\text{max}}),\phi \sim U(\phi_{\text{min}},\phi_{\text{max}}), \theta = \frac{2\pi}{10} t (0 \leq t \leq 9)$. In the image rendering process, we remove the code to check whether an object is visible (that is, whether the number of observation pixels of an object reaches the maximum threshold), so that we hope the model can retrieve the occluded or completely occluded objects from the frame relationship. The size of the generated image of CLEVR and SHOP is $108 \times 64 $. We crop the image to $64 \times 64  $ (the upper boundary is 10, the lower boundary is 74, the left boundary is 22, and the right boundary is 86). 

\begin{figure}
    \centering
    \includegraphics[width=\textwidth]{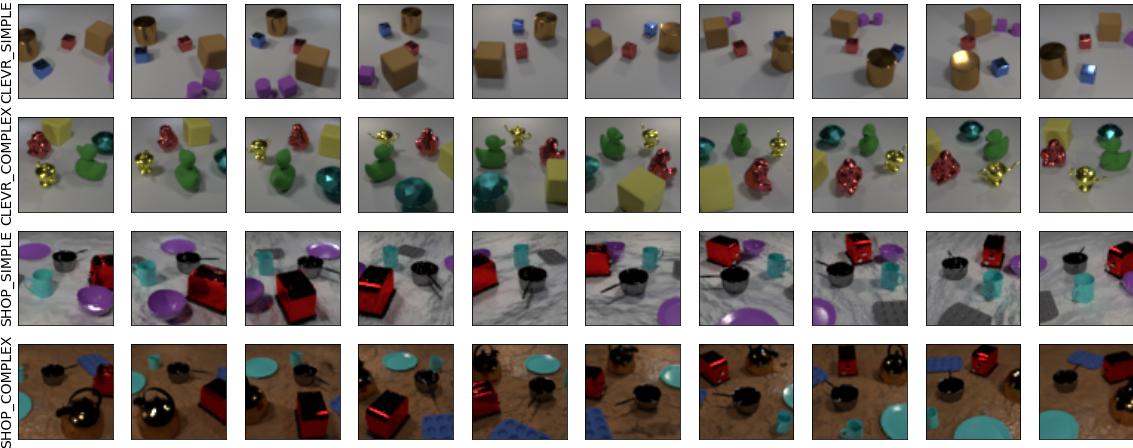}
    \caption{The demonstration of four datasets in test sets.}
    \label{fig:multi_view_dataset}
\end{figure}

\begin{figure}[H]
    \centering
    \includegraphics[width=\textwidth]{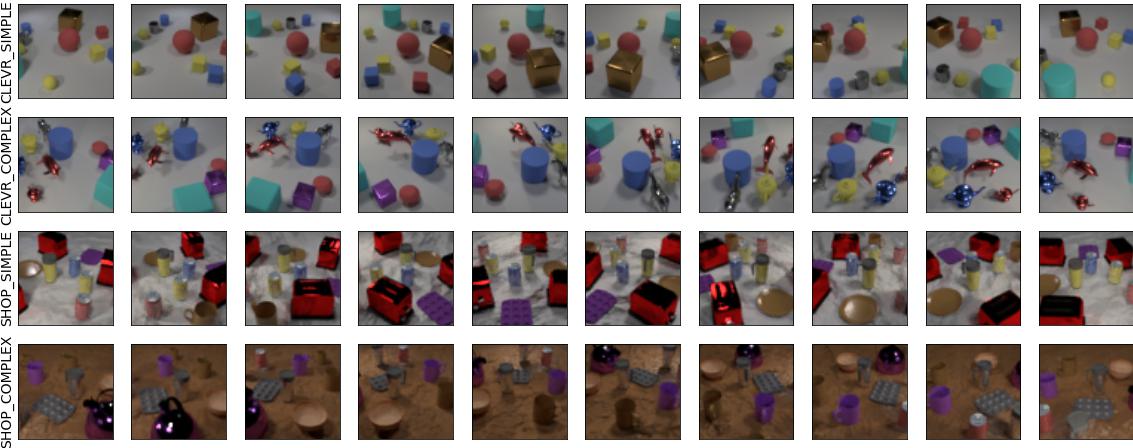}
    \caption{The demonstration of four datasets in general sets.}
    \label{fig:multi_view_dataset_general}
\end{figure}

The CLEVR is further divided into two categories: CLEVR-SIPLE and CLEVR-COMPLEX. CLEVR-SIMPLE includes 3 object categories (with intra class differences), while CLEVR-COMPLEX includes 10 object categories (with intra class differences). Compared with CLEVR-SIMPLE, CLEVR-COMPLEX has greater challenges. SHOP is further divided into two types: SHOP-SIMPlE and SHOP-COMPlEX. SHOP-SIMPlE includes 10 object categories, and the background is selected as marble background. Compared with CLEVR, its objects have greater challenges in texture and material. At the same time, the color of some objects is highly similar to the background, which makes it more difficult to identify. SHOP-COMPLEX has two background options. The second one is a brown background, whose color is highly similar to the object color, further improving the recognition difficulty. The detailed configuration can be found in Table \ref{tab:multi_view_clevr} and \ref{tab:multi_view_shop}. Figure \ref{fig:multi_view_dataset} and \ref{fig:multi_view_dataset_general} demonstrate the samples in test sets and general sets of four datasets. It can be seen that the number of objects in the general set is larger than the test sets, and correspondingly the occlusion rate is higher, leading to a more difficult inference.
\begin{table}[t]
    \centering
    \begin{tabular}{|c|c|c|c|c|c|c|c|c|} 
    \hline
    Datasets& \multicolumn{4}{c|}{CLEVR-SIMPLE}&\multicolumn{4}{c|}{CLEVR-COMPLEX}\\
    \hline
    Split&Train&Valid&Test&General& Train&Vaid&Test&General\\
    \hline
    \# of Images &5000&100&100 & 100 &5000&100&100 & 100\\
    \hline
    \# of Objects&3$\sim$6&3$\sim$6&3$\sim$6&7$\sim$10&3$\sim$6&3$\sim$6& 3$\sim$6 & 7$\sim$ 10\\
    \hline
    \# of Views & \multicolumn{4}{c|}{10} & \multicolumn{4}{c|}{10} \\
    \hline
    \# of Categories & \multicolumn{4}{c|}{3} & \multicolumn{4}{c|}{10} \\
    \hline
    \# of Backgrounds & \multicolumn{4}{c|}{1} & \multicolumn{4}{c|}{1} \\
    \hline
    Image Size &\multicolumn{4}{c|}{64$\times$64}&\multicolumn{4}{c|}{64$\times$64}\\
    \hline
    Azimuth\ $\theta $&\multicolumn{4}{c|}{[0,2$\pi$]} & \multicolumn{4}{c|}{[0,2$\pi$]}\\
    Elevation\ $\rho$ &\multicolumn{4}{c|}{[10.5,12]} & \multicolumn{4}{c|}{[10.5,12]}\\
    Distance\ $\phi$ & \multicolumn{4}{c|}{[0.15$\pi$,0.3$\pi$]} & \multicolumn{4}{c|}{[0.15$\pi$,0.3$\pi$]} \\
    \hline
    \end{tabular}
    \caption{configuration of CLEVR}
    \label{tab:multi_view_clevr}
    \centering
    \begin{tabular}{|c|c|c|c|c|c|c|c|c|} 
    \hline
    Datasets& \multicolumn{4}{c|}{SHOP-SIMPLE}&\multicolumn{4}{c|}{SHOP-COMPLEX}\\
    \hline
    Split&Train&Valid&Test&General& Train&Vaid&Test&General\\
    \hline
    \# of Images &5000&100&100 & 100 &5000&100&100 & 100\\
    \hline
    \# of Objects&3$\sim$6&3$\sim$6&3$\sim$6&7$\sim$10&3$\sim$6&3$\sim$6& 3$\sim$6 & 7$\sim$ 10\\
    \hline
    \# of Views & \multicolumn{4}{c|}{10} & \multicolumn{4}{c|}{10} \\
    \hline
    \# of Categories & \    multicolumn{4}{c|}{10} & \multicolumn{4}{c|}{10} \\
    \hline
    \# of Backgrounds & \multicolumn{4}{c|}{1} & \multicolumn{4}{c|}{2} \\
    \hline
    Image Size &\multicolumn{4}{c|}{64$\times$64}&\multicolumn{4}{c|}{64$\times$64}\\
    \hline
    Azimuth\ $\theta $&\multicolumn{4}{c|}{[0,2$\pi$]} & \multicolumn{4}{c|}{[0,2$\pi$]}\\
    Elevation\ $\rho$ &\multicolumn{4}{c|}{[10.5,12]} & \multicolumn{4}{c|}{[10.5,12]}\\
    Distance\ $\phi$ & \multicolumn{4}{c|}{[0.15$\pi$,0.3$\pi$]} & \multicolumn{4}{c|}{[0.15$\pi$,0.3$\pi$]} \\
    \hline
    \end{tabular}
    \caption{configuration of SHOP}
    \label{tab:multi_view_shop}
\end{table}

\section{Computation of Metrics}
In this section, we will introduce all the metrics used in this article, including some matrics not described in the main text. 1) Adjusted Rand Index (ARI) \cite[]{hubert1985comparing} and Adjusted Mutual Information (AMI) \cite[]{xuan2010information} assess the quality of segmentation, i.e., how accurately images are partitioned into different objects and background. Previous work usually evaluates ARI and AMI only at pixels belong to objects, and how accurately background is separated from objects is unclear. We evaluate ARI and AMI under two conditions. ARI-A and AMI-A are computed considering both objects and background, while ARI-O and AMI-O are computed considering only objects. 2) Intersection over Union (IoU) and $F_1$ score ($F_1$) assess the quality of amodal segmentation, i.e., how accurately complete shapes of objects are estimated. 3) Count assesses the accuracy of the estimated number of objects. 4) Object Ordering Accuracy (OOA) as used in \cite[]{yuan2019generative} assesses the accuracy of the estimated pairwise ordering of objects. We now desrcibe the mathematical computation in the following.
\subsection{Definition}
Suppose the test sets have $I$ visual scenes and each visual scene includes $T$ images from different viewpoints, let $\hat{K}_i$ be the be the real maximum number of objects appearing in the $i$th visual scene (the total number of objects appearing in all visual angles), and let $K_i$ be estimated maximum number of objects appearing in the $i$th visual scene. note that $\hat{K}_i$ and $K_i$ are not necessarily equal. $\hat{\boldsymbol{r}}_i \in \big\{ 0, 1\big\}^{T \times (\hat{K}_i+ 1) \times N }$ and $\boldsymbol{r}_i \in \big\{ 0, 1\big\}^{T \times (\hat{K}_i+ 1) \times N }$ respectively represent the real and estimated one-hot vector of the $T$ viewpoints in the $i$th scene  corresponding to the pixel-wise partitions (including the foreground and background). $\mathcal{D}_t^i$ denotes the index sets that belong to the object areas in the $t$th viewpoint of the $i$th scene, i.e., $\mathcal{D}_t^i = \big\{n \mid x_{t,n}^i \in \text{object areas}\big\}$. Let $\hat{U}_{t,k}^i$ be the real index sets w.r.t. object $k$ in the $t$th viewpoint of the $i$th scene, i.e., $\hat{U}_{t,k}^i =\big\{ n \mid \boldsymbol{x}_{t,n}^i \in \text{areas of object }k \big\}$ $(0\leq k \leq \hat{K}_i)$. Let $U_{t,k}^i$ be the estimated index sets w.r.t. object $k$ in the $t$th viewpoint of the $i$th scene. $\hat{U}_{t,k}^i =\big\{ n \mid \hat{\boldsymbol{x}}_{t,n}^i \in \text{areas of object }k \big\}$ $(0\leq k \leq \hat{K}_i)$, where $\hat{\boldsymbol{x}}$ is the reconstructed image. Let $\hat{\boldsymbol{s}}_{1:}$
\subsection{Adjusted Rand Index}
The computation of Adjusted Rand Index (ARI) is described as:
\begin{align}
    \label{eq:ari}
    \text{ARI} = \frac{1}{I}\sum_{i=1}^I \frac{b_{\text{all}}^i-b_{\text{row}}^i\cdot b_{\text{col}}^i / c^i}{\left(b_{\text{row}}^i+b_{\text{col}} \right)/2 - b_{\text{row}}^i\cdot b_{\text{col}}^i / c^i}
\end{align}
In order to explain the meaning of each variable above in detail, $C (x, y)$ is used here to represent the combination number, i.e., $C(x,y)=\frac{x!}{(x-y)!y!}$; $v_{\hat{k},k}^i$ denotes the dot product, i.e., $v_{\hat{k},k}^i = \sum \nolimits_{(t,n)\in \mathcal{S}}(\hat{r}_{t,k,n}\cdot r_{t,k,n})$, $b_{\text{row}}^i$, $b_{\text{col}}^i$ and $c^i$ in Eq \ref{eq:ari} are described as:
\begin{align}
    \label{eq:ari1}
    b_{\text{all}}^i &= \sum\nolimits_{\hat{k}=0}^{\hat{K}_i}\sum\nolimits_{k=0}^K C\Big(v_{\hat{k},k}^i,2\Big)  \\
    b_{\text{row}}^i &= \sum\nolimits_{\hat{k}=0}^{\hat{K}_i} C\Big(\sum\nolimits_{k=0}^K             v_{\hat{k},k}^i,2\Big) \\
    b_{\text{col}}^i &= \sum\nolimits_{k=0}^K C\Big(\sum\nolimits_{\hat{k}=0}^{\hat{K}_i} v_{\hat{k},k}^i,2\Big) \\
    c^i &= C\big(\sum \nolimits_{\hat{k}=0}^{\hat{K}_i} \sum \nolimits_{(t,n)\in \mathcal{S}} \hat{r}_{t,\hat{k},n}^i,2\big)
\end{align}
where $\mathcal{S} = \big\{1,2,...,T\big\} \times \big\{1,2,...,N\big\}$. When computing ARI-O, pixels in $\mathcal{S}$ that do not belong to objects will be removed; When ARI-A is calculated, all pixels in $\mathcal{S}$ will be used.
\subsection{Adjusted Mutual Information}
\begin{align}
    \label{eq:ami_multi_view}
    \text{AMI} = \frac{1}{I} \sum_{i=1}^I \sum_{t=1}^T \frac{\text{MI}(\hat{\boldsymbol{l}}^i,\boldsymbol{l}^i)- \mathbb{E}\big[ \text{MI}(\hat{\boldsymbol{l}}^i,\boldsymbol{l}^i)\big]}{\big(\text{H}(\hat{\boldsymbol{l}}^i)+\text{H}(\boldsymbol{l}^i)\big)/2-\mathbb{E}\big[ \text{MI}(\hat{\boldsymbol{l}}^i,\boldsymbol{l}^i)\big]}
\end{align}
where $\hat{\boldsymbol{l}}^i \in \mathbb{R}^{T\times (\hat{K}_i + 1)}$. $\hat{\boldsymbol{l}}_{t}^i$ represents the probability distribution of the $t$th viewpoint in the $i$th visual scene, i.e.,$\hat{l}_t^i = \big\{ |\hat{U}_{t,k}| / |\mathcal{D}_t^i| \mid 0 \leq k \leq \hat{K}_i\big\} $. \text{H} and \text{MI} respectively represent the entropy and mutual information of the distribution.
\begin{align}
    \label{eq:ami2}
    \text{H}(\hat{\boldsymbol{l}}^i) &= -\sum\nolimits_{k=0}^{\hat{K}_i}\sum \nolimits_{k=1}^{T} \hat{l}_{t,k}^i\log \hat{l}_{t,k}^i \\
    \text{H}(\boldsymbol{l}^i) &= -\sum\nolimits_{k=0}^{K_i} \sum \nolimits_{k=1}^{T} l_{t,k}^i\log l_{t,k}^i \\
    \text{MI}(\hat{\boldsymbol{l}}^i,\boldsymbol{l}^i) &= \sum_{m=0}^{\hat{K}_i}  \sum_{n=0}^{K_i} \sum \limits_{t=1}^{T} p_{t, m,n}^i \log\Big(\frac{p_{t, m,n}^i}{\hat{l}_{t,m}^i\cdot l_{t,n}^i}\Big)
\end{align}
where $\hat{l}_{t,k}^i$ and $l_{t,k}^i$ respectively represent the probability that the pixel in the $i$th image is partitioned to object $k$. $p_{t, m,n}^i$ indicates the probability that pixels in the $t$th frame of the $i$th image are divided into objects $m$ in the first set and objects $n$ in the second set.$ p_{t,m, n}^i $ is calculated as follows:
\begin{align}
    p_{t,m,n}^i = \frac{o_{t,m,n}^i}{|\mathcal{D}_t^i|}=\frac{|\hat{U}_{t,m}^i\cap U_{t,n}^i|}{|\mathcal{D}_t^i|}
\end{align}
The matrix $\boldsymbol{o}_t^i\in \mathbb{R}^{(\hat{K}_i+1)\times(K_i+1)}$ is called the contingency table. And the expectation of MI can be analytically computed:
\begin{align}
    \label{eq:expectation_mi}
         \mathbb{E}\big[ \text{MI}(\hat{\boldsymbol{l}}^i,\boldsymbol{l}^i)\big] = \sum_{t=1}^T \sum_{m=0}^{\hat{K}_i}&\sum_{n=0}^{K_i}\sum_{k=(a_{t,m}^i+b_{t,n}^i-N)^{+}}^{\min (a_{t,m}^i,b_{t,n}^i)}\frac{k}{N}\cdot \log\Big( \frac{N\times k}{a_{t,m}^i\times b_{t,n}^i}\Big) \notag\\
         &\frac{a_{t,m}^i!b_{t,n}^i!(N-a_{t,m}^i)!(N-b_{t,n}^i)}{N!k!(a_{t,m}^i-k)!(b_{t,n}^i-k)!(N-a_{t,m}^i-b_{t,n}^i+k)!}
\end{align}
where $(a_{t,m}^i+b_{t,n}^i-N)^{+}=\max(1,a_{t,m}^i+b_{t,n}^i-N)$, $a_{t,m}^i$ and $b_{t,n}^i$ respectively represent the sum of rows and columns w.r.t. $\boldsymbol{o}_t^i$:
\begin{align}
    a_{t,m}^i=\sum\nolimits_{n=0}^{K_i}o_{t,m,n}^i, \quad b_{t,n}^i=\sum\nolimits_{m=0}^{\hat{K}_i}o_{t,m,n}^i
\end{align}
\subsection{Intersection over Union}
In order to compute the Intersection over Union (IoU), we should define two variables: $\hat{\boldsymbol{s}}^i \in \big[0,1 \big]^{T \times \hat{K}_i\times N}$ and $\boldsymbol{s}^i \in \big[0,1\big]^{T\times K_i\times N}$, IoU from multiple viewpoints requires object index matching under multiple viewpoints, that is, 
\begin{align}
    \label{eq:multi_view_index}
    \boldsymbol{\xi}^i = \text{argmax}_{\boldsymbol{\xi}^i \in \boldsymbol{\Omega}}^i \sum \nolimits_{t=1}^T\sum\nolimits_{k=1}^{\hat{K}_i}\sum\nolimits_{n=1}^N \hat{r}_{t,k,n}^i\cdot r_{t, \xi_k^i,n}^i
\end{align}
where $\boldsymbol{\Omega}^i$ is the full arrangement of all object indexes. And the computation of IoU is desceribed as:
\begin{align}
    \text{IoU} = \frac{1}{I} \sum_{i=1}^I \frac{1}{\hat{K}_i}\sum_{k=1}^{\hat{K}_i} \frac{\sum \nolimits_{t=1}^T \sum \nolimits_{n=1}^N \min(\hat{s}_{t,k,n}^i,s_{t,k,n}^i)}{\sum \nolimits_{t=1}^T \sum \nolimits_{n=1}^N \max(\hat{s}_{t,k,n}^i,s_{t,k,n}^i)}
\end{align}

\subsection{F1 Score}
$F_1$ Score is computed as:
\begin{align}
    F_1 = \frac{1}{I} \sum_{i=1}^I \frac{1}{\hat{K}_i}\sum_{k=1}^{\hat{K}_i} \frac{2\cdot \sum \nolimits_{t=1}^T \sum \nolimits_{n=1}^N \min(\hat{s}_{t,k,n}^i,s_{t,k,n}^i)}{\sum \nolimits_{t=1}^T \sum \nolimits_{n=1}^N \big( \min(\hat{s}_{t,k,n}^i,s_{t,k,n}^i) +  \max(\hat{s}_{t,k,n}^i,s_{t,k,n}^i) \big)}
\end{align}
\subsection{Count}
$\hat{K}_i$ and $K_i$ represent the real/estimated object numbers in the $i$th visual scene. When it comes to the model with $\boldsymbol{z}^{\text{pres}}$, $K_i$ is computed through $K_i= \sum_{i=1}^{ \tilde{K}_i} \mathrm{z}_i^{\text{pres}}$. For the model without $\boldsymbol{z}^{\text{pres}}$, the method to determine the number is: if a layer has no object pixels, the number of objects will not be included. Let $\delta$ denotes the Kronecker delta function, and Count is computed as follows:
\begin{equation}
    \label{eq:ca}
    \text{Count}=\frac{1}{I}\sum\nolimits_{i=1}^I\delta_{\hat{K}_i,K_i}
\end{equation}

\subsection{Object Ordering Accuracy} 
Another set of vectors should be introduced to compute the ordering relationship of objects. Let $\hat{o}_{t,k_1,k_2}^i \in \big\{0,1 \big\},o_{t,k_1,k_2}^i \in \big\{0,1 \big\}$ respectively represent the real/estimated order of the $k_1$ object and the $k_2$ object. Here, the index order of the real object matches the estimated object index one by one. This matching relationship is obtained through the formula Eq \ref{eq:multi_view_index}. The estimated object index will be redirected to $\boldsymbol{\xi}^i$. Because it is difficult to estimate the depth ordering of two objects if they do not overlap, the following OOA calculation measures the importance of different object pairs with different weights:
\begin{align}
    \text{OOA}=\frac{1}{I} \sum_{i=1}^I \frac{1}{T} \sum_{t=1}^{T} \frac{\sum_{k_1=1}^{\hat{K}_i-1} \sum_{k_2=k_1+1}^{\hat{K}_i} w_{t, k_1, k_2}^i \delta_{\hat{o}_{t, k_1, k_2}^i, o_{t, k_1, k_2}^i}}{\sum_{k_1=1}^{\hat{K}_i-1} \sum_{k_2=k_1+1}^{\hat{K}_i} w_{t, k_1, k_2}^i}
\end{align}
The weight of object pairs $k_1$ and $k_2$ is calculated as follows:
\begin{align}
    w_{m, k_1, k_2}^i = \sum \nolimits_{n=1}^N \hat{s}_{t,k,n}^i \cdot s_{t,k,n}^i
\end{align}
The value of $w_{t, k_1, k_2}^i$ reflects the overlapping area of two different object shapes. When the overlapping area is larger, it is easier to do depth sorting, that is, its contribution to the measurement of OOA is greater; On the contrary, when there is little or no overlap ($w_{t,k_1, k_2}^i \rightarrow 0 $), it has little impact on the measurement of OOA.

\section{Hyperparameter Configuration}
We detail the hyperparameter configuration in this section, including the network design, learning rate, temperature schedule, e.t.c. . During the training, the standard deviation $\sigma_x$ of the likelihood function is chosen to be 0.2. The object slot number $K$ (i.e., the maximum number that may appear in the visual scene) is set to be 7. The dimension of $\boldsymbol{\lambda}$,$\boldsymbol{z}^{\text{view}}$, $\boldsymbol{z}^{\text{obj}}$, $\boldsymbol{z}^{\text{bck}}$ is respectively 5/3/64/16. The fixed standard deviation $\sigma_{w}$ of $\boldsymbol{\lambda}$ is 0.8. The hyperparameter $\alpha$ of $\boldsymbol{z}^{\text{view}}$ is 12.6. During the stage 1 training, since there are no labels for the model, it's very difficult to extract the feature from multiple views. We use the warm-up schedule, i.e. single-view training. We used single-view training in the first 30k steps to better initialize the network parameters (single-view learning is easier than multi-view learning), then the model gradually transits to multiple views (for example, you can directly jump to 4 viewpoints, or 2$\rightarrow$4). In the sequential extention of Slot Attention \cite[]{locatello2020object}, we have additional hyperparameters. $\boldsymbol{y}^{\text{view}}$ and $\boldsymbol{y}^{\text{attr}}$ respectively have 8 and 128 dimensions, and $D_{\text{key}}$ and $D_{\text{val}}$ are 64 and 136. The iteration step is set to be 3. In the learning, the batch size is chosen to be 32. The initial learning rate is $4e-4$, and is decayed exponentially with a factor 0.5 every 50,000 steps. In the ﬁrst 10,000 training steps, the learning rate is multiplied by a factor that is increased linearly from 0 to 1. For the temperature of $\boldsymbol{z}^{\text{pres}}$, the logarithmic temperature decreases linearly from 10 to 0.5 in the first 150k steps. Now, let us introduce the additional hyperparameters of Stage 1 training and Stage 2 training.
\paragraph{Stage 1 training} During the Stage 1 training, since the prior of view latent variables are standard Gaussian distributions, the model does not need to introduce GPs with neural networks. We set the same weight for all KLs. The weight increases linearly from 0 to 1 in the first 100k steps to stabilize the training. 
\paragraph{Stage 2 training} During the Stage 2 training, we aim to learn the view function of $t$. Since we use the pretrained model in phase 1, its feature extraction for objects and backgrounds has been stable. At this time, we need to adjust the prior of view latent variables. $\boldsymbol{A}$ in the $\boldsymbol{\lambda}$ prior is implemented by a single full-connected layer without bias. The neural networks of GPs will be detailed in the description of the neural network design. For course learning, we realize it by gradually increasing the view numbers. In the training process, we set different courses for different datasets. In short, for every tens of thousands of steps, two images corresponding to the additional viewpoints will be added as inputs. SHOP has more iterations per course than CLEVR. 
\paragraph{Neural Network Design} We list the neural networks used in the model:
\begin{itemize}
    \item $f_{\text{feat}}$ denotes the view encoder
    
    - 3 $\times$ 3, stride=(2,2), padding=(1,1), Conv(3,64), ReLU

    - 3 $\times$ 3, stride=(2,2), padding=(1,1), Conv(64,64), ReLU

    - 3 $\times$ 3, stride=(2,2), padding=(1,1), Conv(64,64)

    \item the position encoder denotes the 3D postion layer, the xy position and the time $t$ will be encoded to the feature with 192 dimensions, then make the mapping with the conv layer
    
    - 1 $\times$ 1, Conv(192,64)

    \item View Transformer encoder (before downsample) is the same configuration as SIMONe \cite[]{kabra2021simone} with 4 layers and 4 heads
    
    \item 2 $\times$ 2 Downsample
    
    \item Spatial Transformer Encoder (after downsample) is set with 4 layers and 4 heads
    
    \item $f_{\lambda}$ denotes the encoder that maps the feature to the mean and variance of $\boldsymbol{\lambda}$
    
    - Linear(66,64), ReLU

    - Linear(64,32), ReLU

    - Linear(32,15), ReLU

    - Reshape(3,5) (3 corresponds to view, 5 corresponds to the spatial attribues)

    \item Slot Attention encoder $f_{\text{sa}}$
    
    - position embedding layer: 1 $\times$ 1 Conv(4,64)

    - 5 $\times$ 5, padding=(2,2), stride=(1,1), Conv(3,64), ReLU

    - 5 $\times$ 5, padding=(2,2), stride=(1,1), Conv(64,64), ReLU

    - 5 $\times$ 5, padding=(2,2), stride=(1,1), Conv(64,64), ReLU

    - 5 $\times$ 5, padding=(2,2), stride=(1,1), Conv(64,64), ReLU

    - LayerNorm(64)

    - Linear(64,64), ReLU

    - Linear(64,64)

    \item sequential extention of Slot Attention
    
    - layerNorm(64) (input)

    - layerNorm(136) (query)
    
    - layerNorm(146) (residual)

    - query: Linear(136,64)

    - key: Linear(64,64)

    - val: Linear(64,136)

    - gru(136,136)

    - residual net: Linear(136,128), ReLU, Linear(128,136)

    \item view mapping layer that maps the feature extracted from the Transformer to the view slot in the Slot Attention 
    - Linear(64,8)

    \item $f_{\text{view}}$ denotes the view encoder that maps the view slot to the mean and variance of $\boldsymbol{z}^{\text{view}}$
    
    - Linear(8,512), ReLU

    - Linear(512,512), ReLU

    - Linear(512,6)

    \item neural networks that correspond to the learnable GP kernal: LargeFeatureExtractor $\times$ 3. The design of LargeFeatureExtractor is as follows:
    
    - Linear(5,32), ReLU

    - Linear(32,32), ReLU

    - Linear(32,64), ReLU

    - Linear(64,64), ReLU

    - Linear(64,8)

    \item $f_{\text{obj}}$ (the output is splited to [128,1,1,1]) denotes the encoder that encode the object slots to the parameters of object latent variables.
    
    - Linear(128,512), ReLU

    - Linear(512,512), ReLU

    - Linear(512,131)

    \item $f_{\text{bck}}$ denotes the encoder that encode the background slot to the parameters of the background latent variable.
    
    - Linear(128,512), ReLU

    - Linear(512,512), ReLU

    - Linear(512,32)

    \item $g_{\text{ord}}$ outputs the order value of each object from multiple viewpoints
    
    - Linear(67,512), ReLU

    - Linear(512,512), ReLU

    - Linear(512,1)

    \item $g_{\text{obj}}$ denotes the object decoder
    
    - Linear(67,4096), ReLU

    - Linear(4096,4096), ReLU

    - Linear(4096,8192),ReLU

    - Flatten()
    
    - 2 $\times$ Interpolate; 5 $\times$ 5, padding=(2,2), stride=(1,1), Conv(128,128); ReLU

    - 5 $\times$ 5, padding=(2,2), stride=(1,1), Conv(128,64); ReLU

    - 2 $\times$ Interpolate; 5 $\times$ 5, padding=(2,2), stride=(1,1), Conv(64,64); ReLU

    - 5 $\times$ 5, padding=(2,2), stride=(1,1), Conv(64,32); ReLU

    - 2 $\times$ Interpolate; 5 $\times$ 5, padding=(2,2), stride=(1,1), Conv(32,32); ReLU

    - 3 $\times$ 3, padding=(1,1), Conv(32,4)

    \item $g_{\text{bck}}$ denotes the background decoder
    
    - Linear(11,512), ReLU

    - Linear(512,512), ReLU

    - Linear(512,256), ReLU

    - Flatten()

    - 4 $\times$ Interpolate; 5 $\times$ 5, padding=(2,2), stride=(1,1), Conv(16,16); ReLU

    - 5 $\times$ 5, padding=(2,2), stride=(1,1), Conv(16,16); ReLU

    - 4 $\times$ Interpolate; 5 $\times$ 5, padding=(2,2), stride=(1,1), Conv(16,16); ReLU

    - 3 $\times$ 3, padding=(1,1), Conv(16,3)

\end{itemize}

\section{Additional Experimental Results}
In this section, we add more visualization results and comparson results of four datasets, 
including the observation evaluation and prediction evaluation. 
Since we use the Stage 1 results to evaluate the quality of the representations, 
we compare our proposed model with three models called MulMON \cite[]{li2020learning}, 
SIMONe \cite[]{kabra2021simone} and OCLOC \cite[]{yuan2022unsupervised}, 
where MulMON is trained and tested with viewpoint annotations and SIMONe and OCLOC are unsupervised generative
We use the Stage 2 results to evaluate the accuracy of novel viewpoints' predictions with the time $t$. As far as we know, there is no model that can only use time $t$ to predict novel viewpoints. For this reason, we compare it with MulMON model based on viewpoint annotations. We do not compare the proposed model with SIMONe and OCLOC in terms of the prediction since both of them cannot make predictions from novel viewpoints. Note that we compare almost all the metrics, where the computation of IoU and $F_1$ in MulMON and SIMONe is based on the mask rather than the complete shape, there will be errors to some extent. Nevetheless, we make a complete table of these data.

\subsection{Unsupervised Learning from Multiple Viewpoints} 
\begin{figure}[ht]
    \centering
    \subfigure[MulMON]{
    \begin{minipage}[a]{0.45\textwidth}
    \includegraphics[width=1\textwidth]{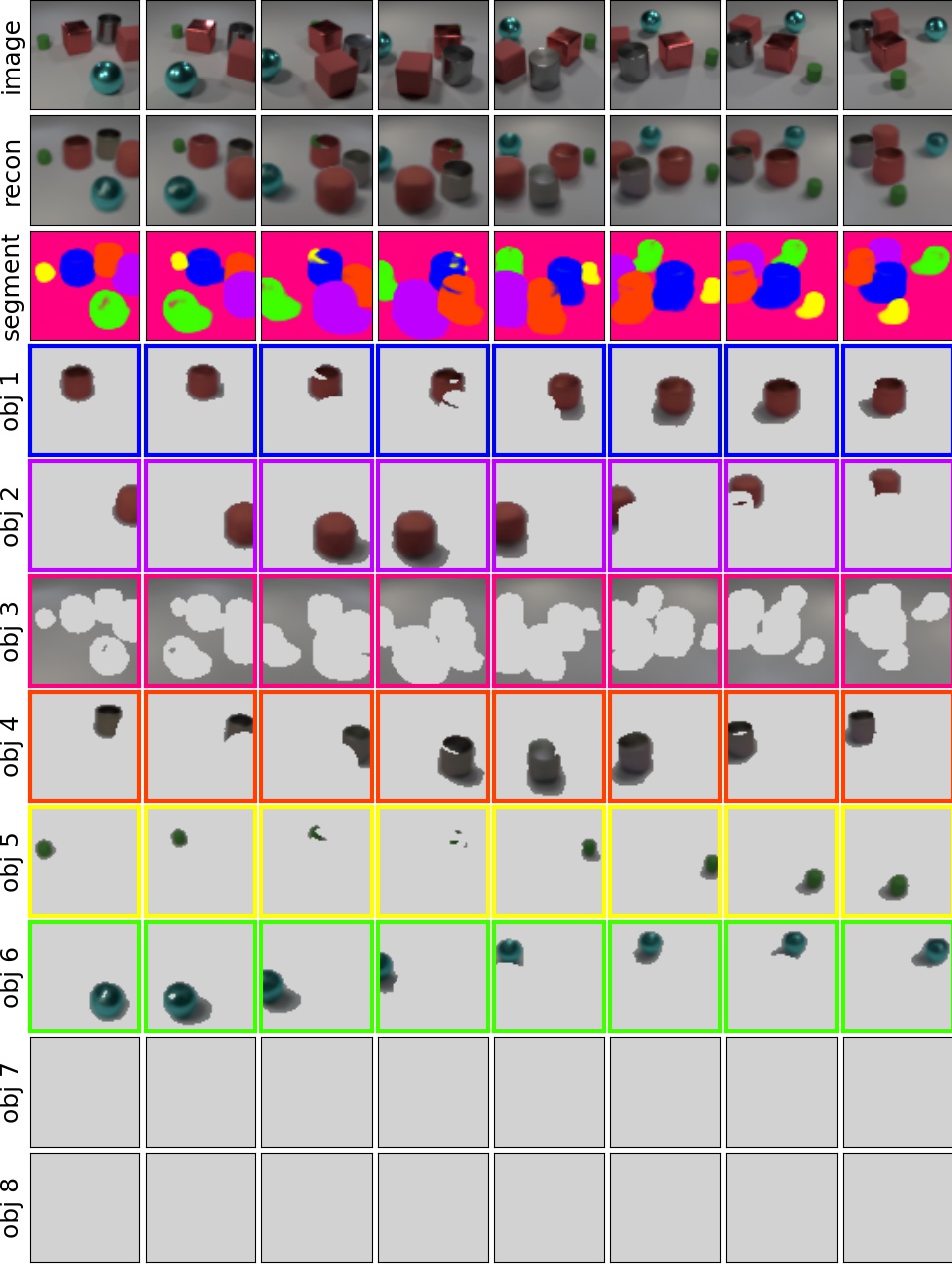}
    \end{minipage}
    }
    \subfigure[SIMONe]{
    \begin{minipage}[a]{0.45\textwidth}
    \includegraphics[width=1\textwidth]{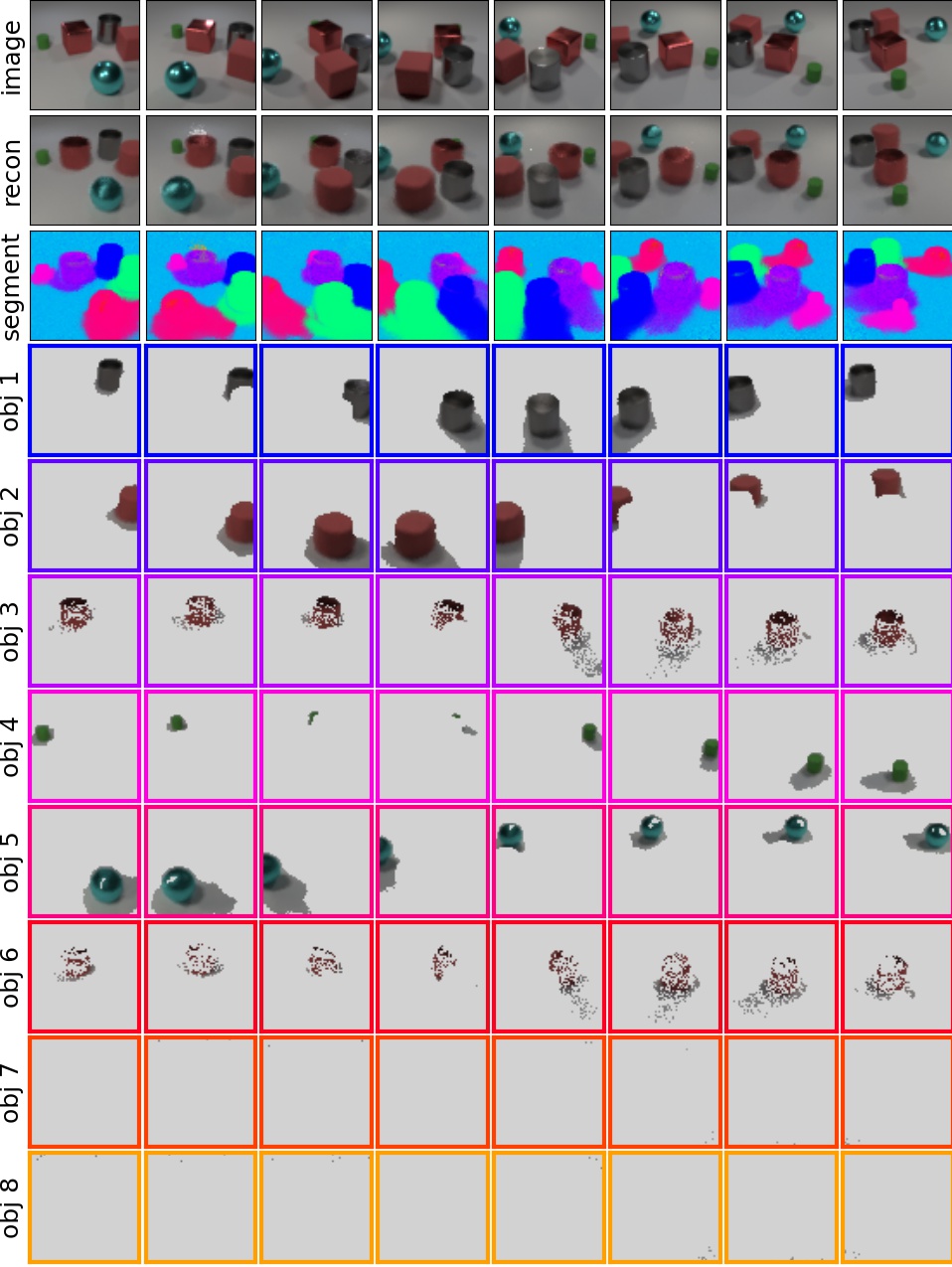}
    \end{minipage}
    }
    \\
    \centering
    \subfigure[OCLOC]{
    \begin{minipage}[a]{0.45\textwidth}
    \includegraphics[width=1\textwidth]{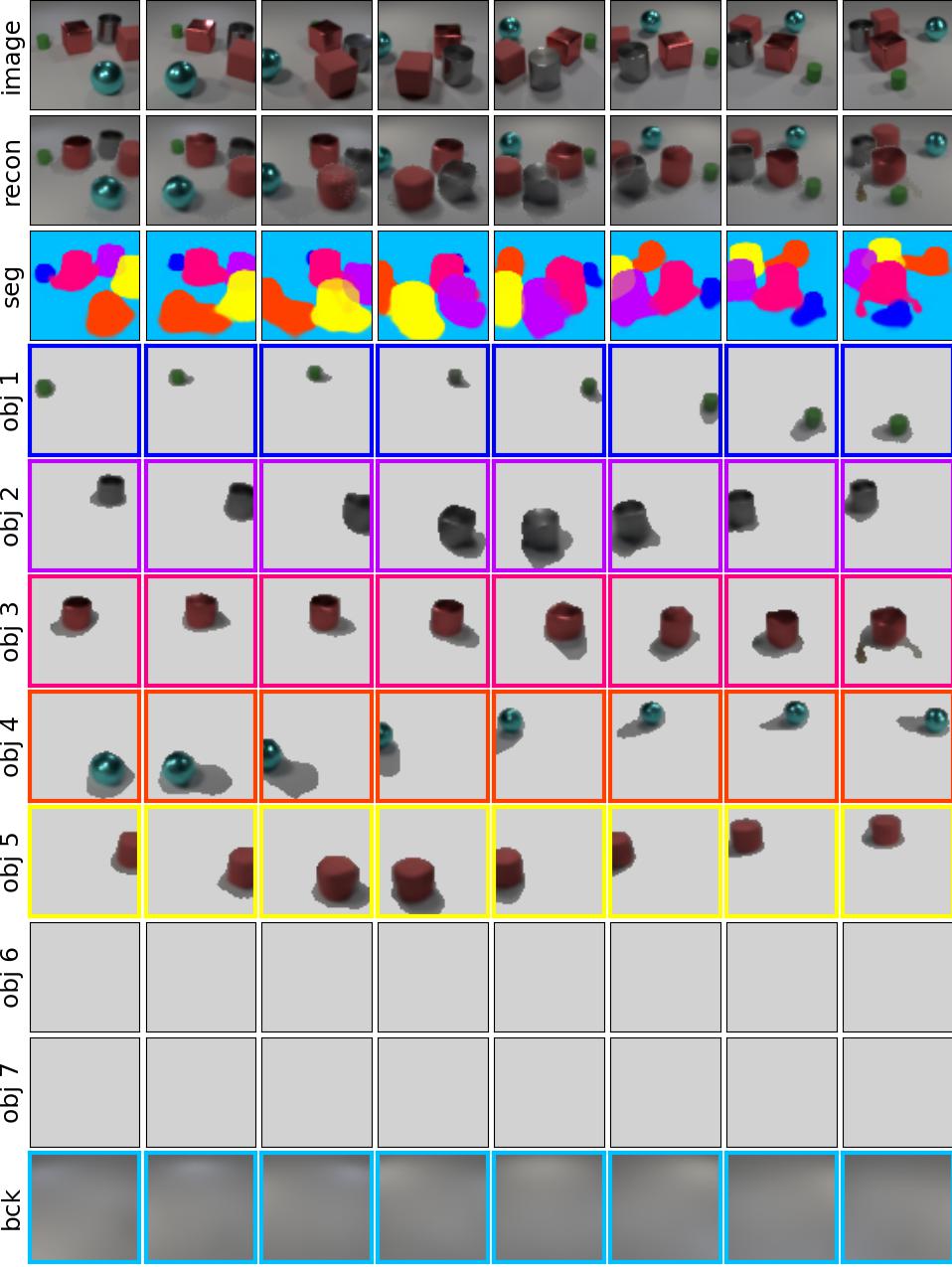}
    \end{minipage}
    }
    \subfigure[Ours]{
    \begin{minipage}[a]{0.45\textwidth}
    \includegraphics[width=1\textwidth]{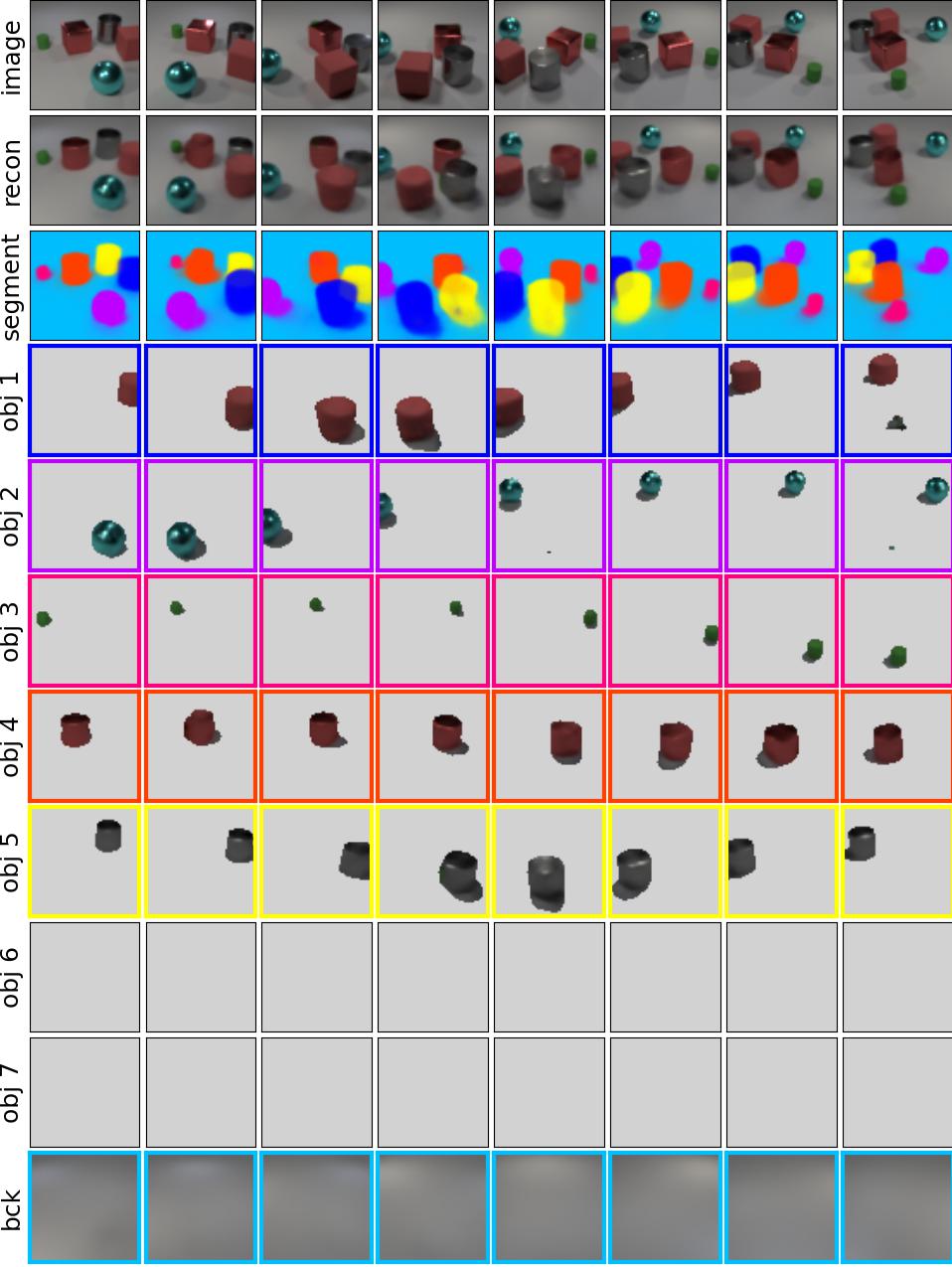}
    \end{minipage}
    }
    \caption{Qualitative comparison of observation on the CLEVR-SIMPLE dataset, observed views are 8}
    \label{fig:decompose_clevr_simple_test}
\end{figure}

\begin{figure}[ht]
    \centering
    \subfigure[MulMON]{
    \begin{minipage}[a]{0.45\textwidth}
    \includegraphics[width=1\textwidth]{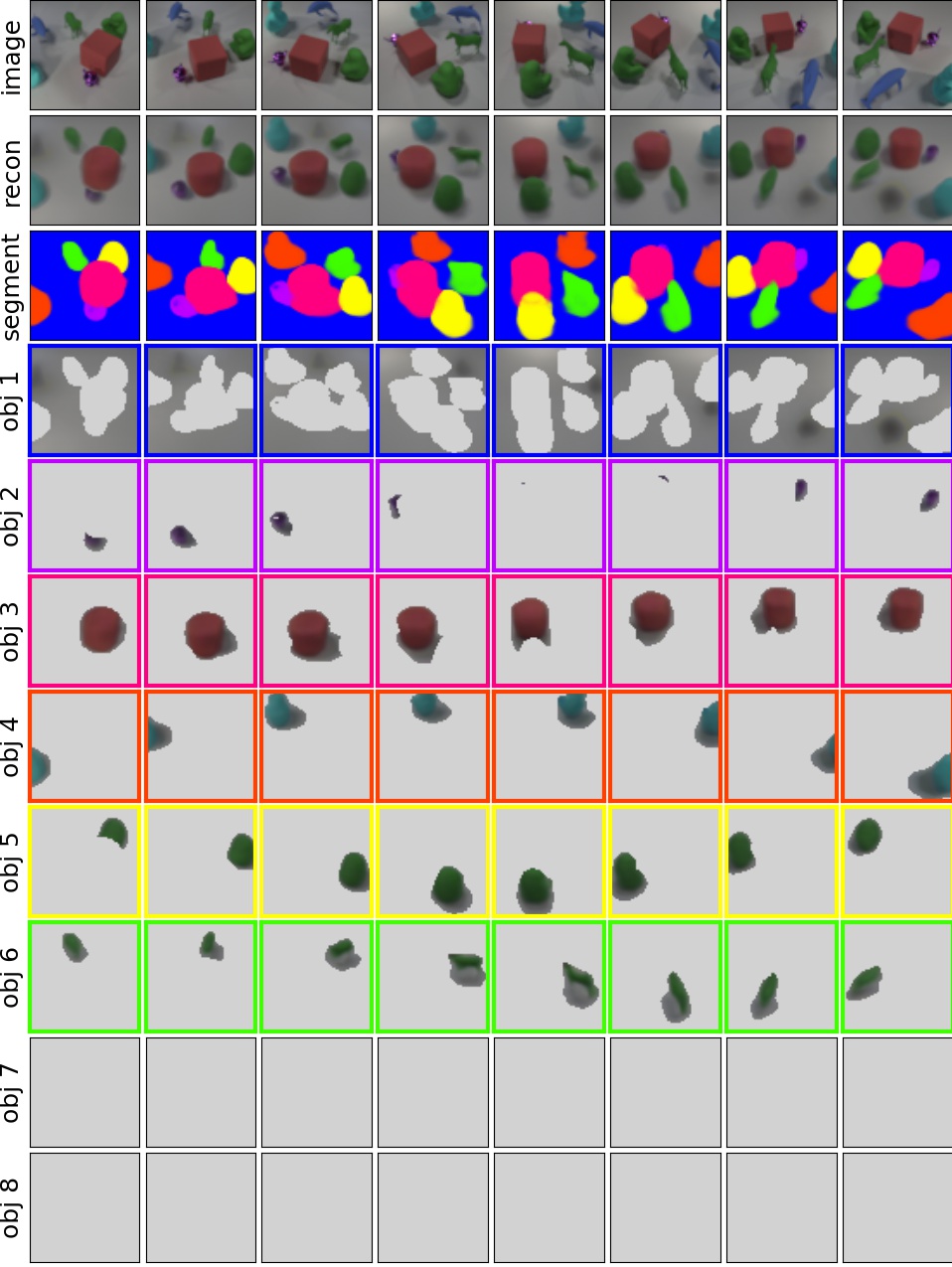}
    \end{minipage}
    }
    \subfigure[SIMONe]{
    \begin{minipage}[a]{0.45\textwidth}
    \includegraphics[width=1\textwidth]{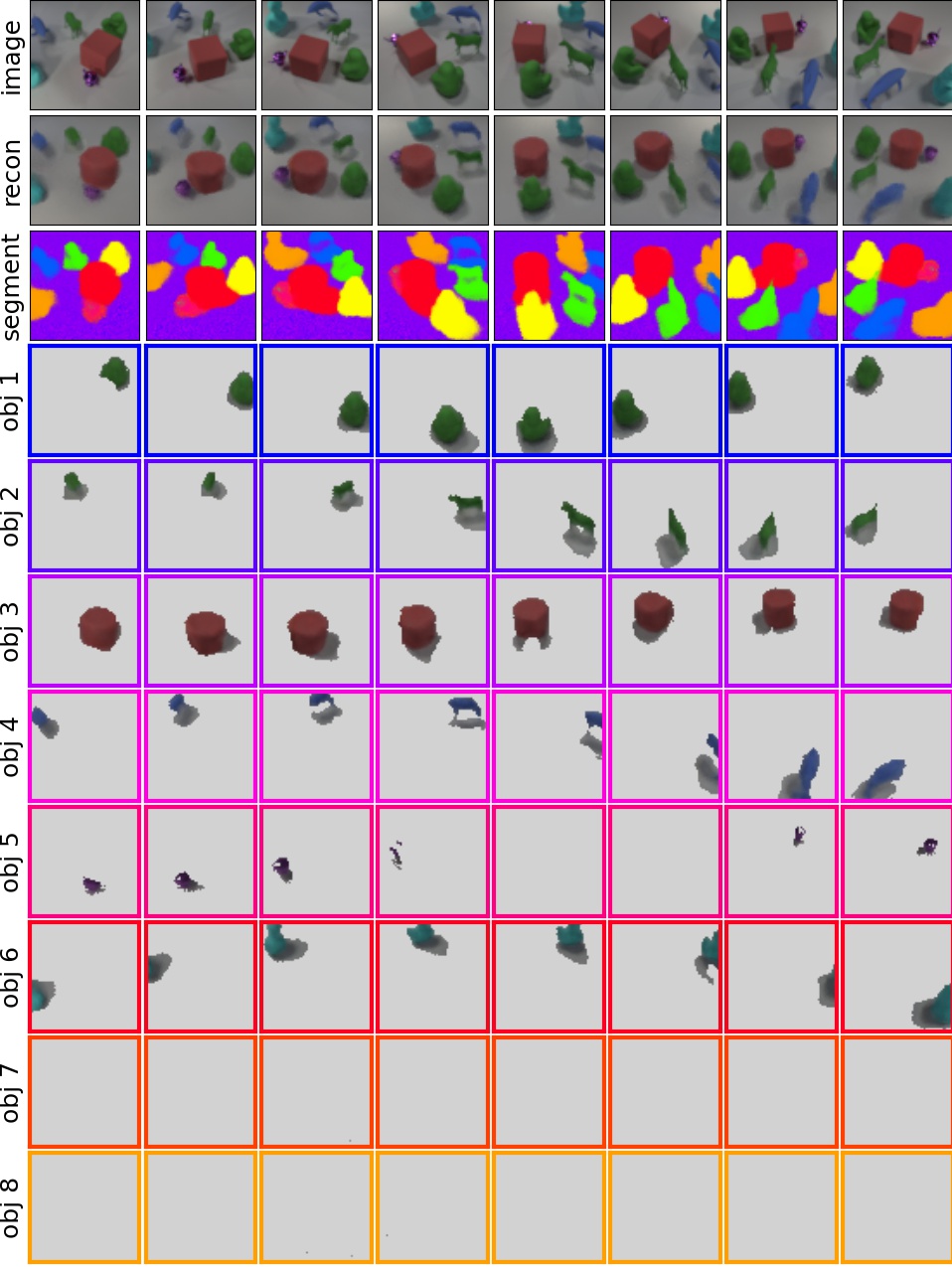}
    \end{minipage}
    }
    \\
    \centering
    \subfigure[OCLOC]{
    \begin{minipage}[a]{0.45\textwidth}
    \includegraphics[width=1\textwidth]{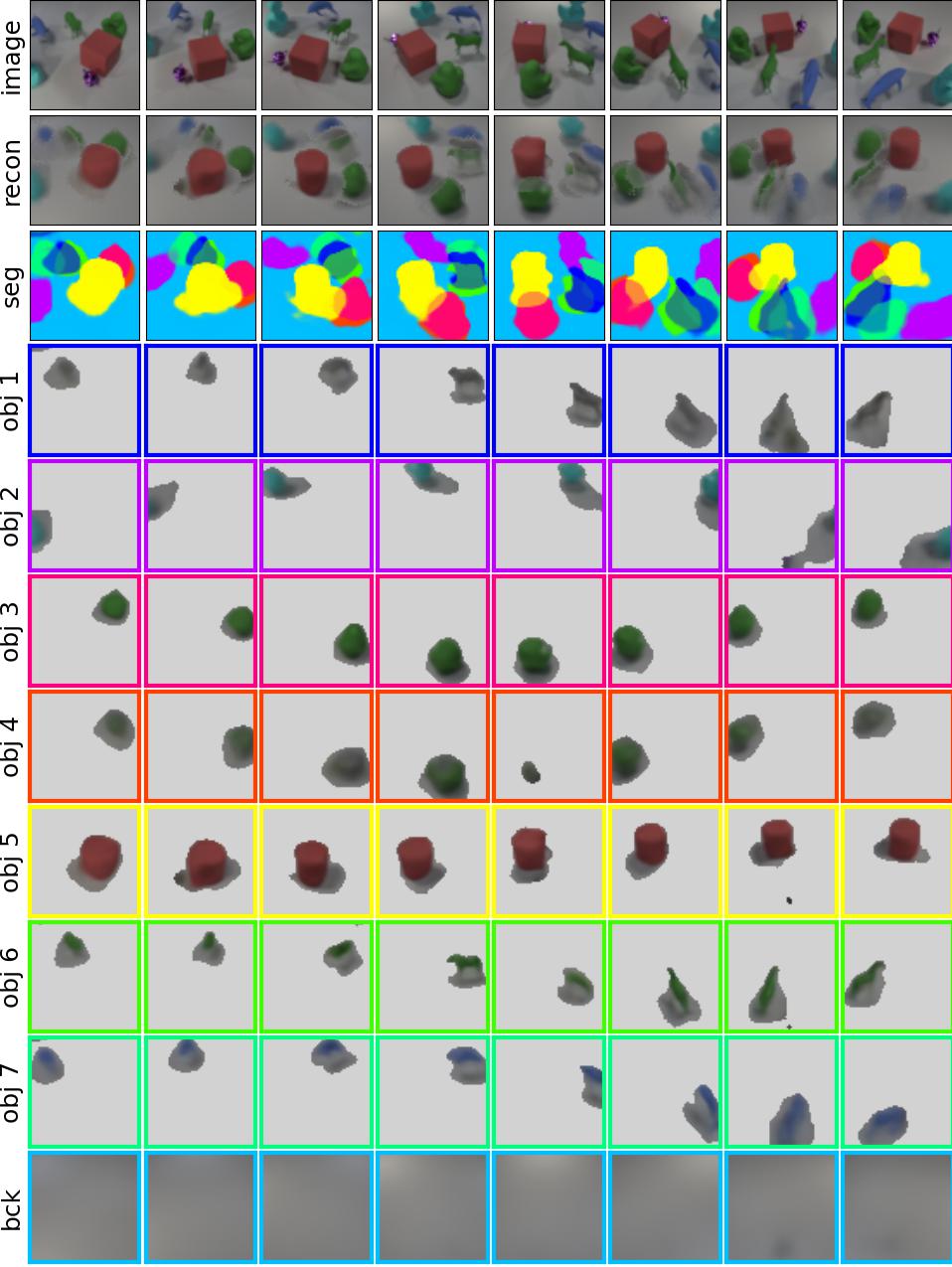}
    \end{minipage}
    }
    \subfigure[Ours]{
    \begin{minipage}[a]{0.45\textwidth}
    \includegraphics[width=1\textwidth]{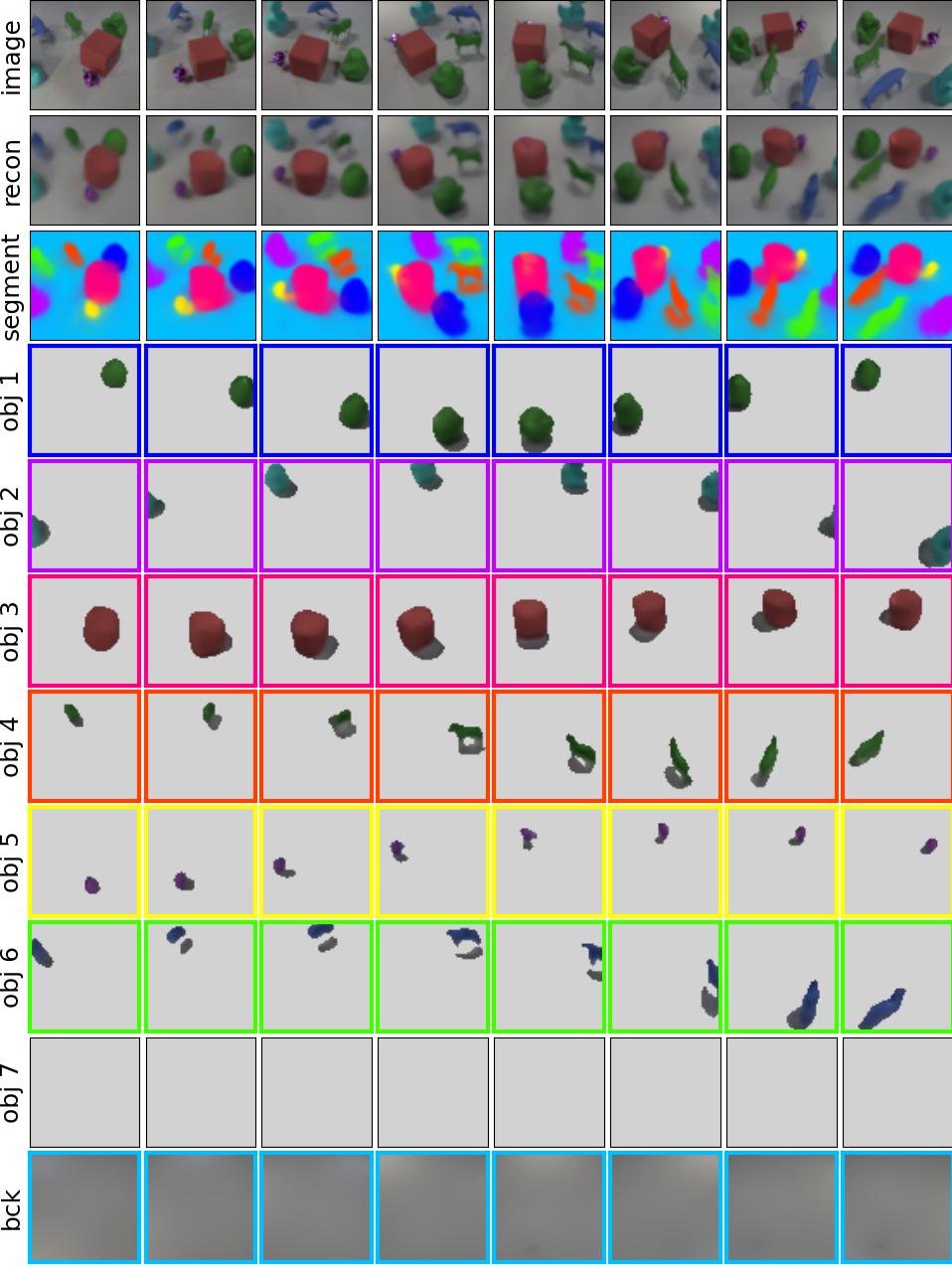}
    \end{minipage}
    }
    \caption{Qualitative comparison of observation on the CLEVR-COMPLEX dataset, observed views are 8}
    \label{fig:decompose_clevr_complex_test}
\end{figure}

\begin{figure}[ht]
    \centering
    \subfigure[MulMON]{
    \begin{minipage}[a]{0.45\textwidth}
    \includegraphics[width=1\textwidth]{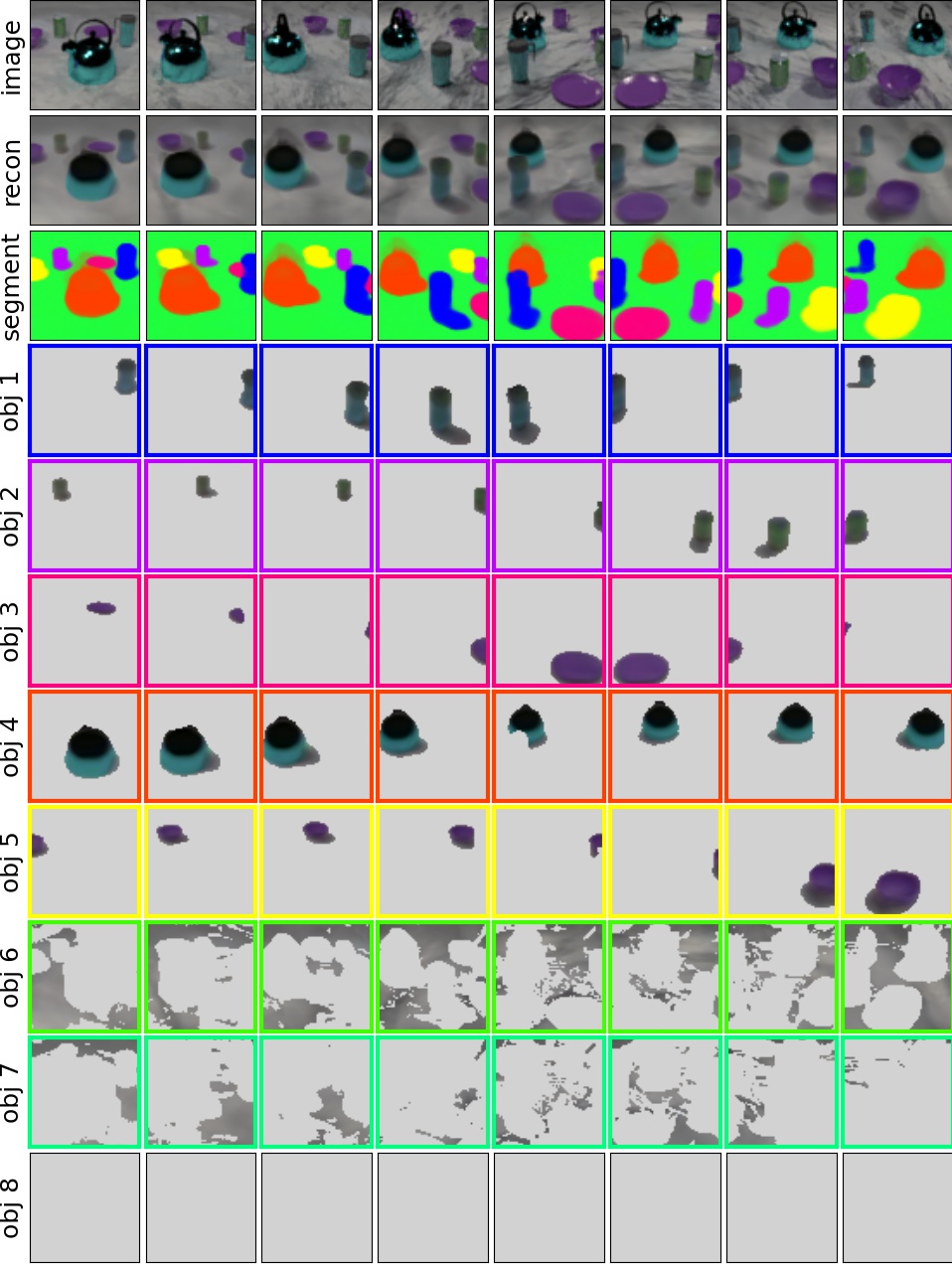}
    \end{minipage}
    }
    \subfigure[SIMONe]{
    \begin{minipage}[a]{0.45\textwidth}
    \includegraphics[width=1\textwidth]{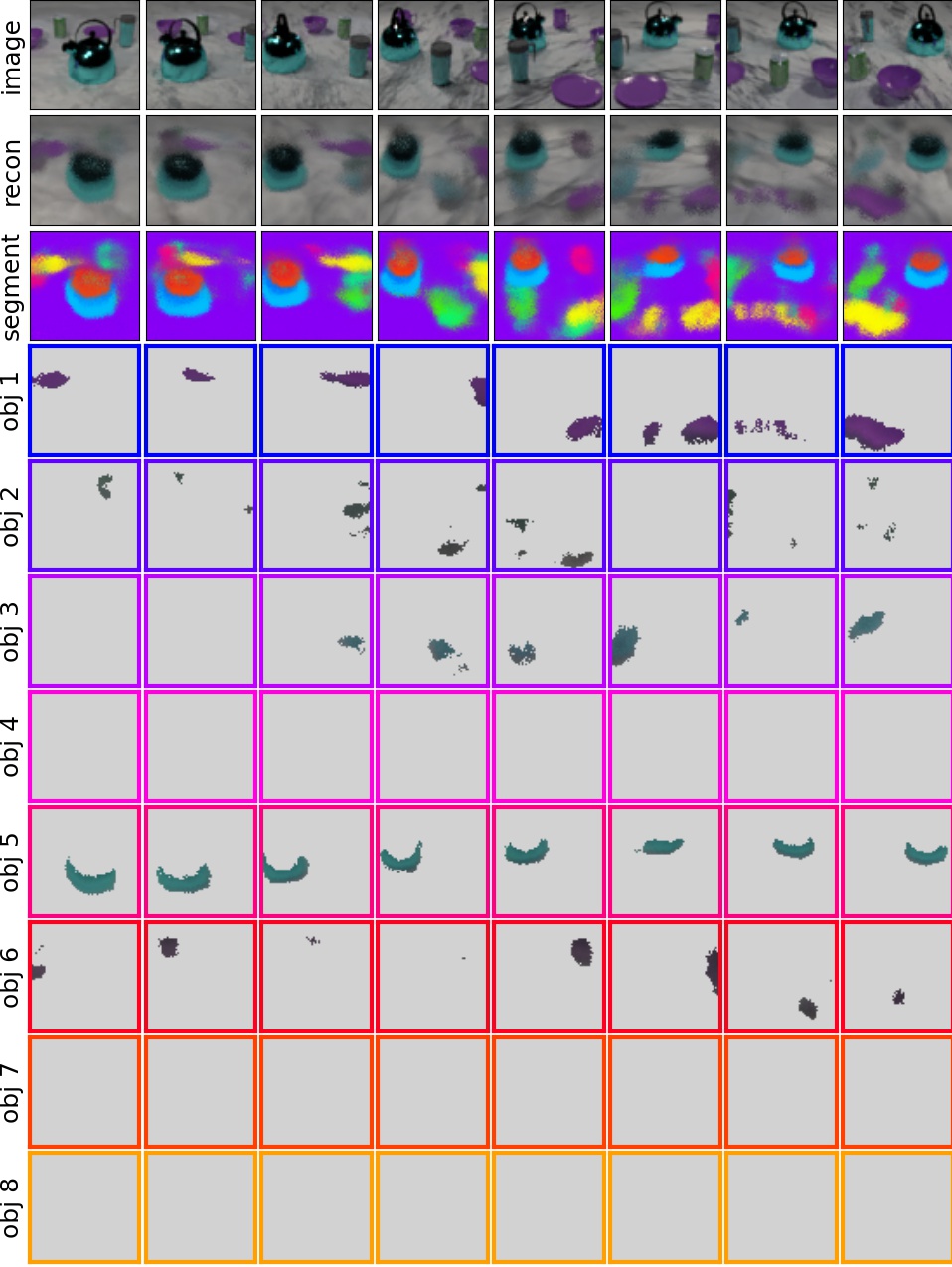}
    \end{minipage}
    }
    \\
    \centering
    \subfigure[OCLOC]{
    \begin{minipage}[a]{0.45\textwidth}
    \includegraphics[width=1\textwidth]{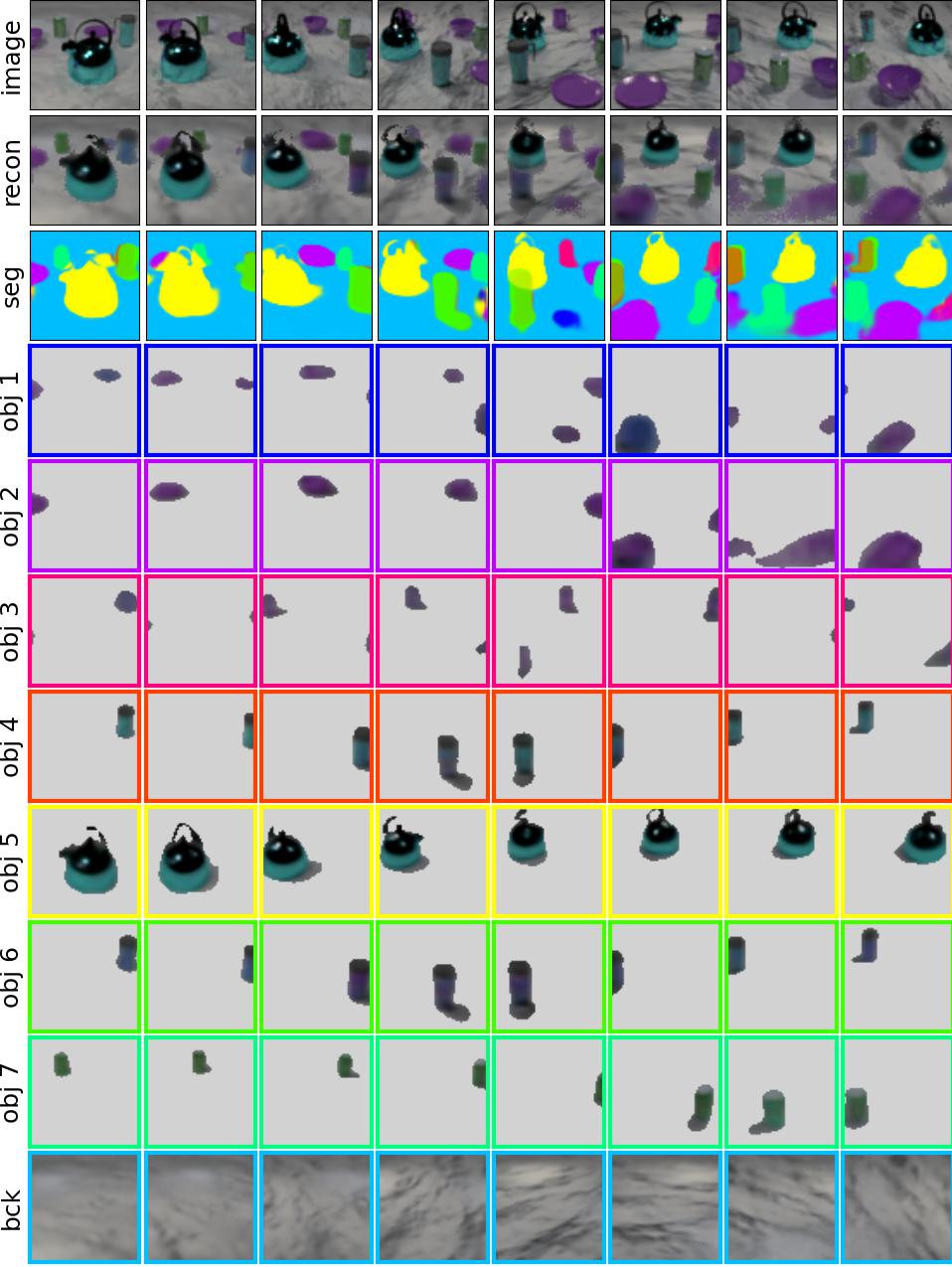}
    \end{minipage}
    }
    \subfigure[Ours]{
    \begin{minipage}[a]{0.45\textwidth}
    \includegraphics[width=1\textwidth]{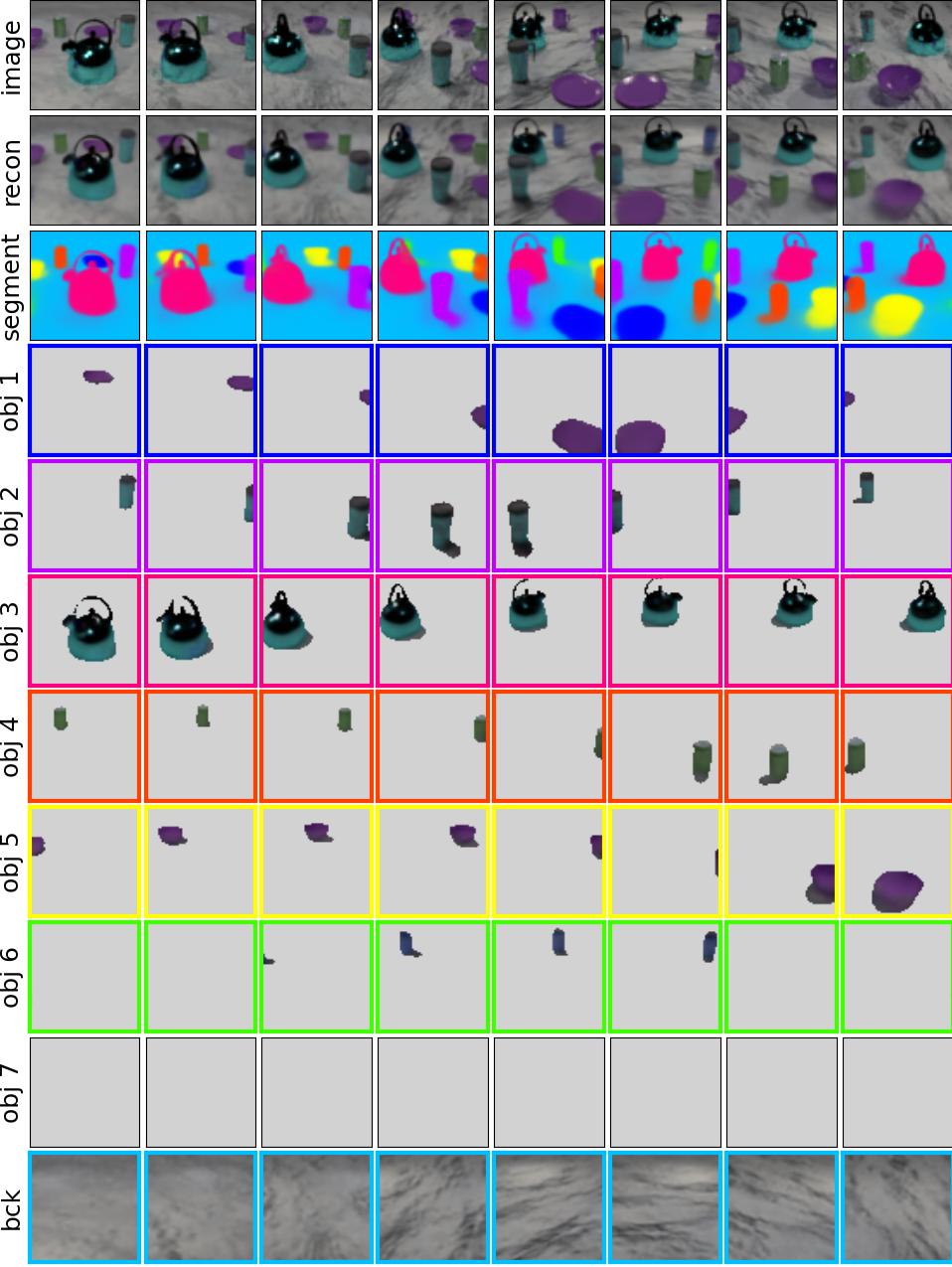}
    \end{minipage}
    }
    \caption{Qualitative comparison of observation on the SHOP-SIMPLE dataset, observed views are 8}
    \label{fig:decompose_shop_simple_test}
\end{figure}

\begin{figure}
    \centering
    \subfigure[MulMON]{
    \begin{minipage}[a]{0.45\textwidth}
    \includegraphics[width=1\textwidth]{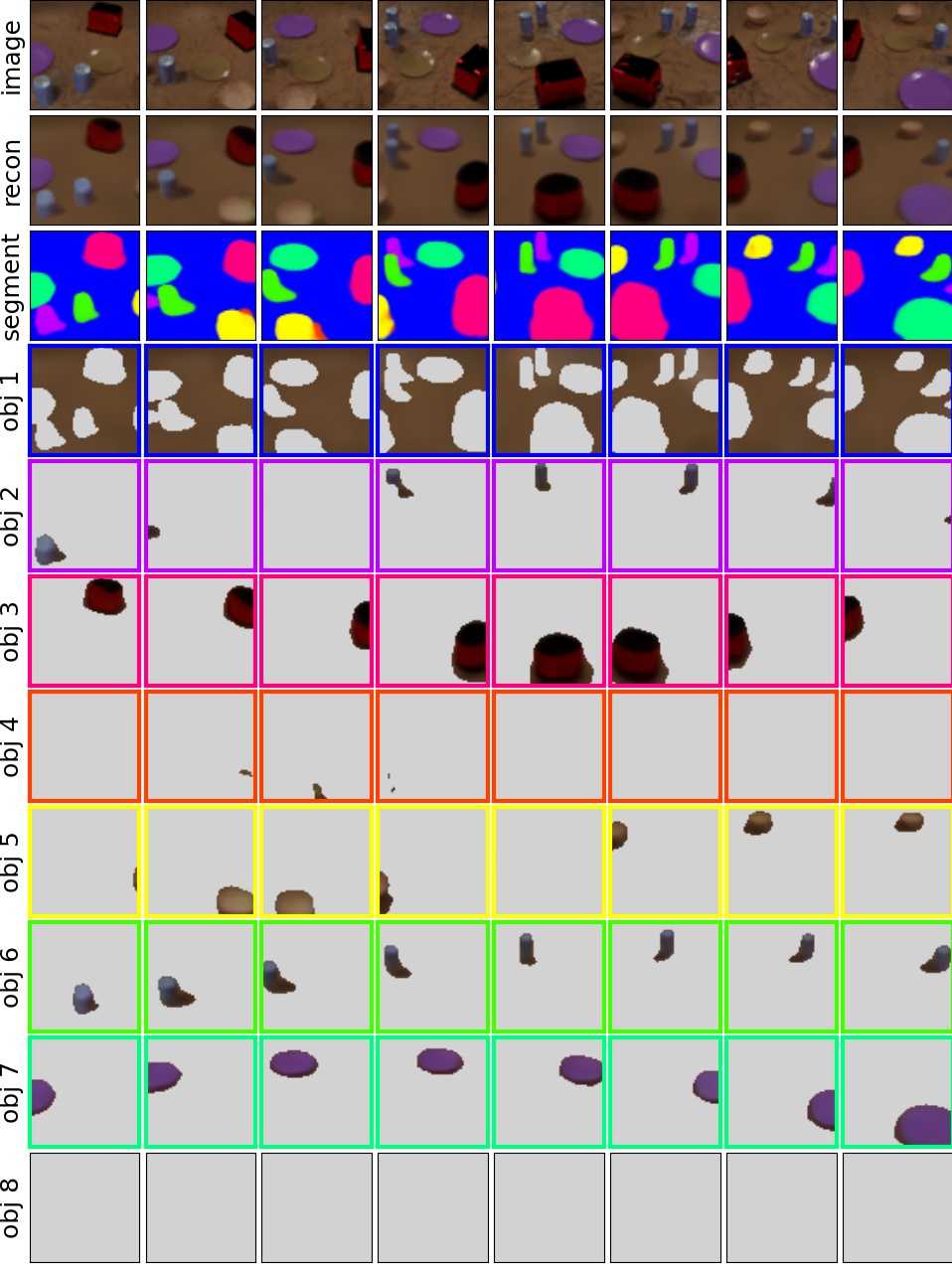}
    \end{minipage}
    }
    \subfigure[SIMONe]{
    \begin{minipage}[a]{0.45\textwidth}
    \includegraphics[width=1\textwidth]{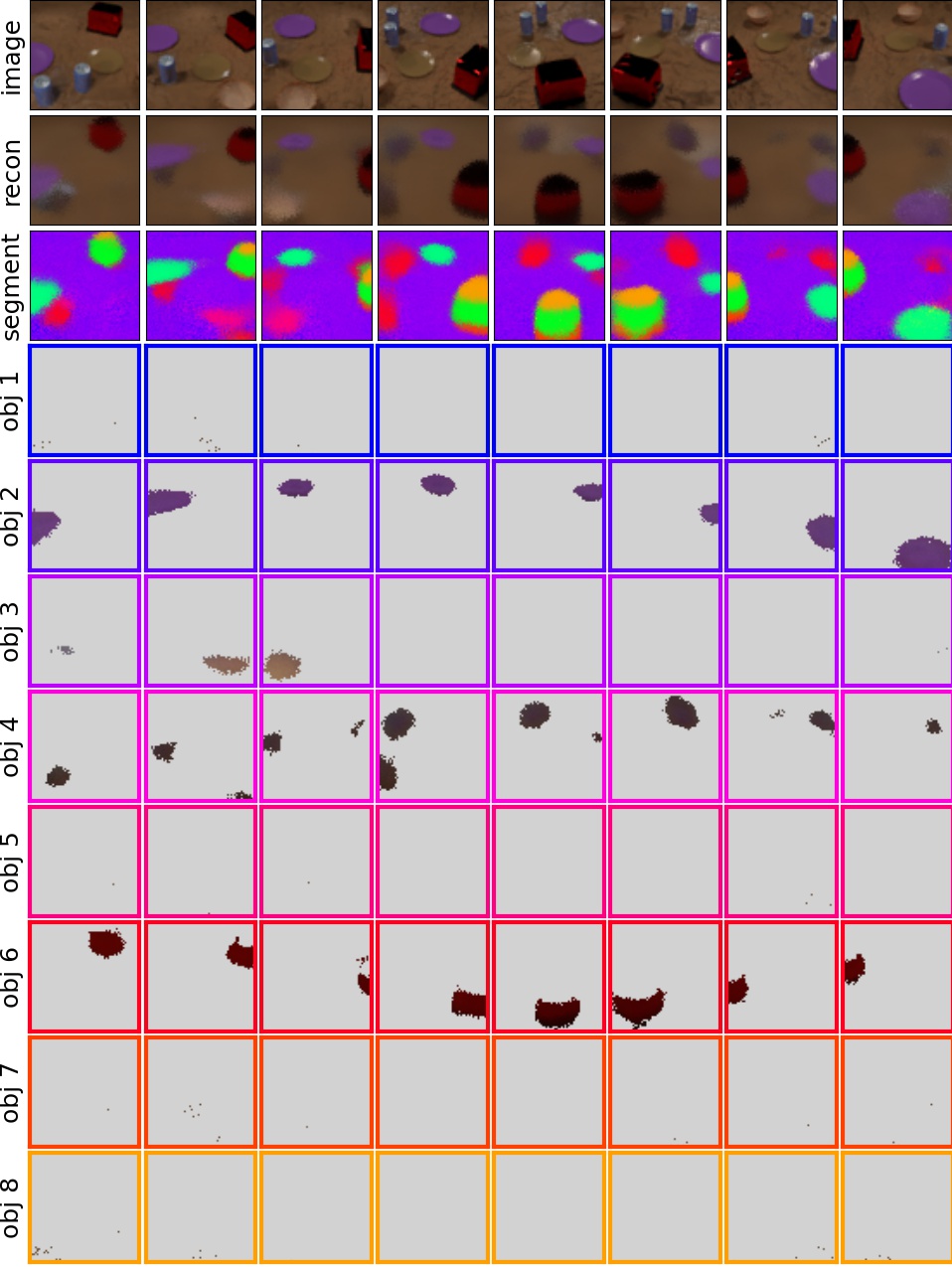}
    \end{minipage}
    }
    \centering
    \subfigure[OCLOC]{
    \begin{minipage}[a]{0.45\textwidth}
    \includegraphics[width=1\textwidth]{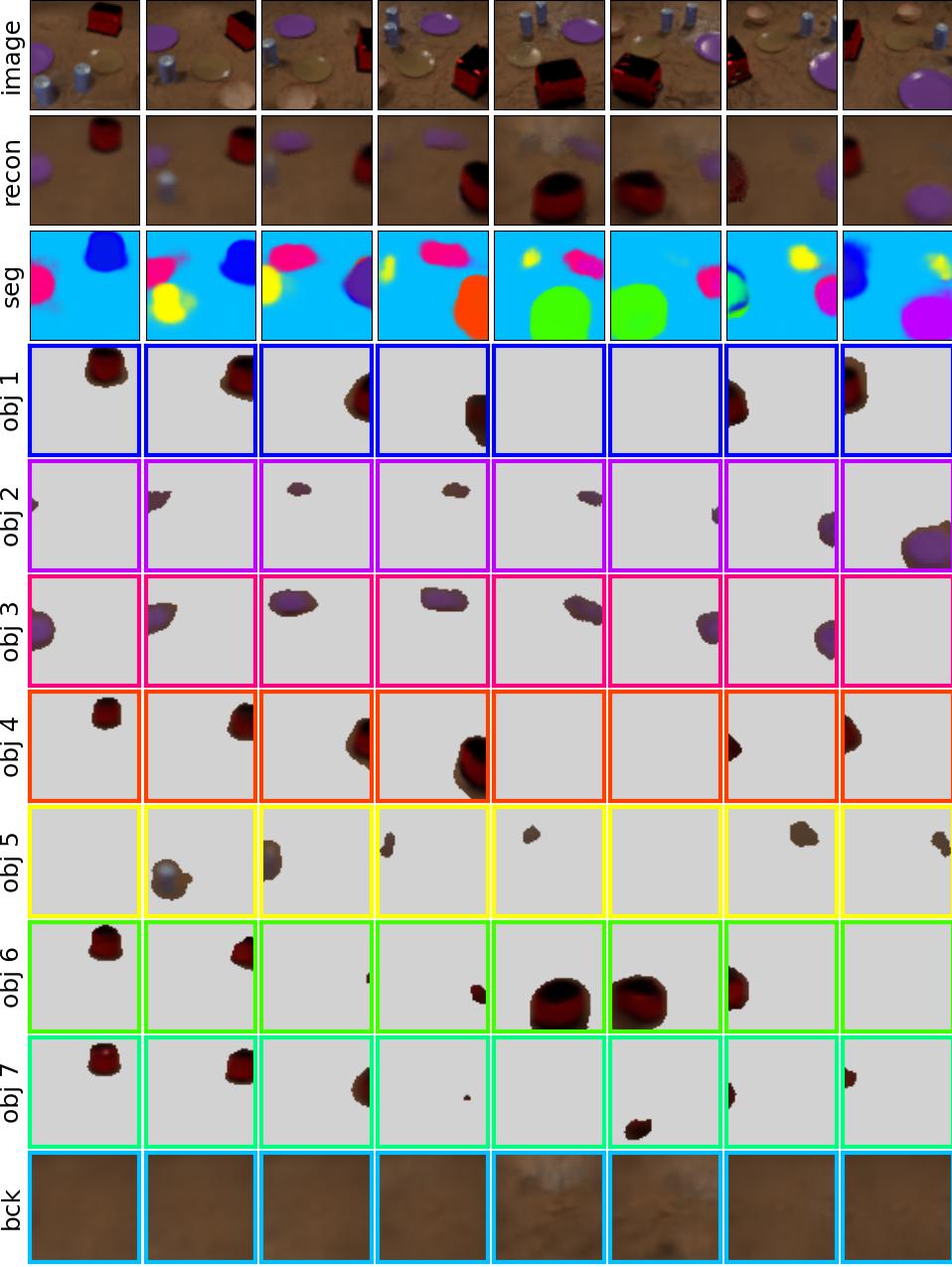}
    \end{minipage}
    }
    \subfigure[Ours]{
    \begin{minipage}[a]{0.45\textwidth}
    \includegraphics[width=1\textwidth]{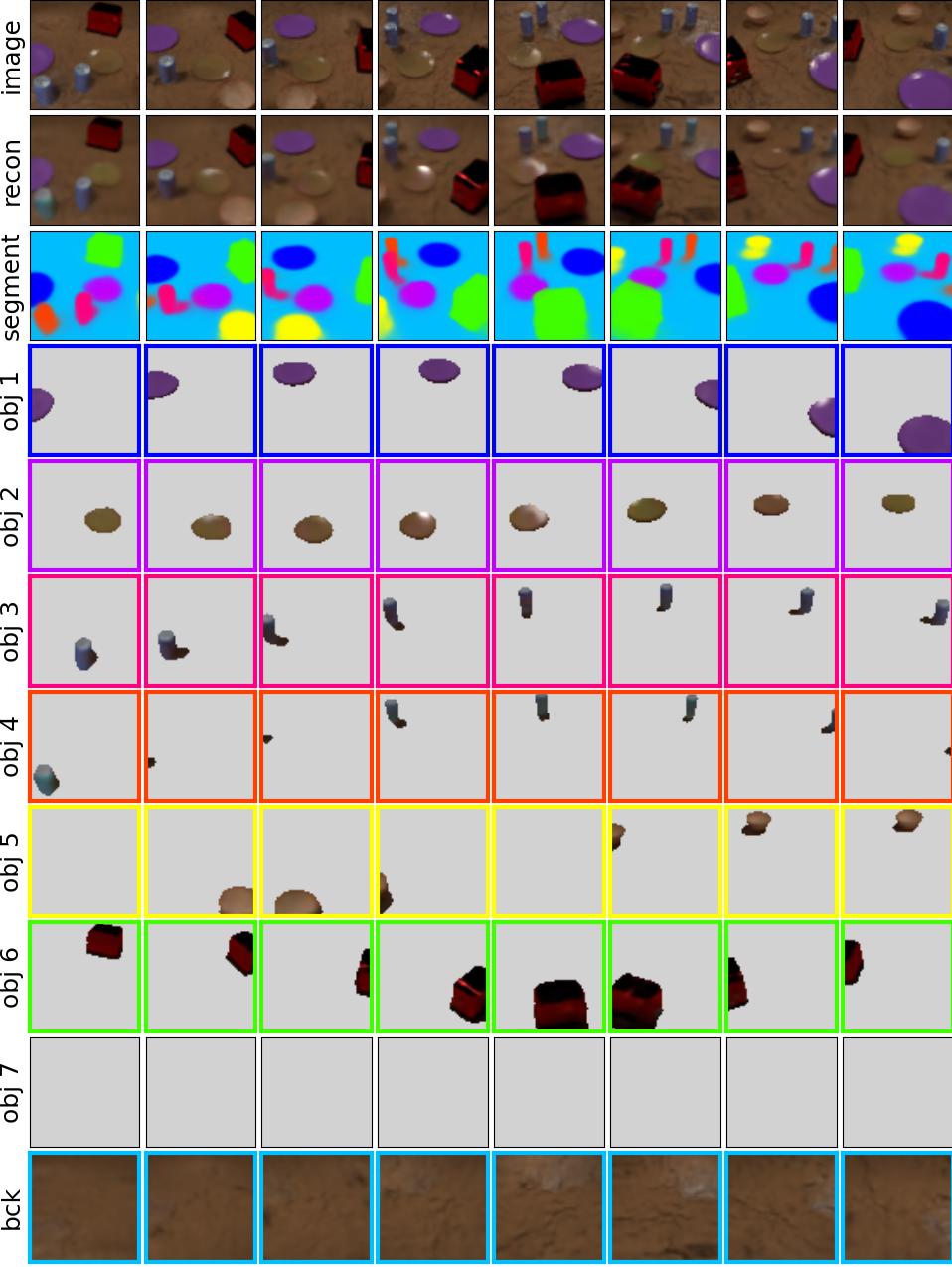}
    \end{minipage}
    }
    \caption{Qualitative comparison of observation on the SHOP-COMPLEX dataset, observed views are 8}
    \label{fig:decompose_shop_complex_test}
\end{figure}

Figures \ref{fig:decompose_clevr_simple_test}, \ref{fig:decompose_clevr_complex_test}, \ref{fig:decompose_shop_simple_test} and \ref{fig:decompose_shop_complex_test} demonstrate the compared results of four datasets. We can find that our proposed can 1) separate the background from the foreground, which is not reflected in MulMON and SIMONe 2) can completely reconstruct the occluded object from some viewpoints. 3) can effectively remove shadows.This problem is very serious in OCLOC, and we solved it effectively.

\begin{figure}
    \centering
    \subfigure[CLEVR-SIMPLE]{
    \begin{minipage}[a]{0.3\textwidth}
    \includegraphics[width=1\textwidth]{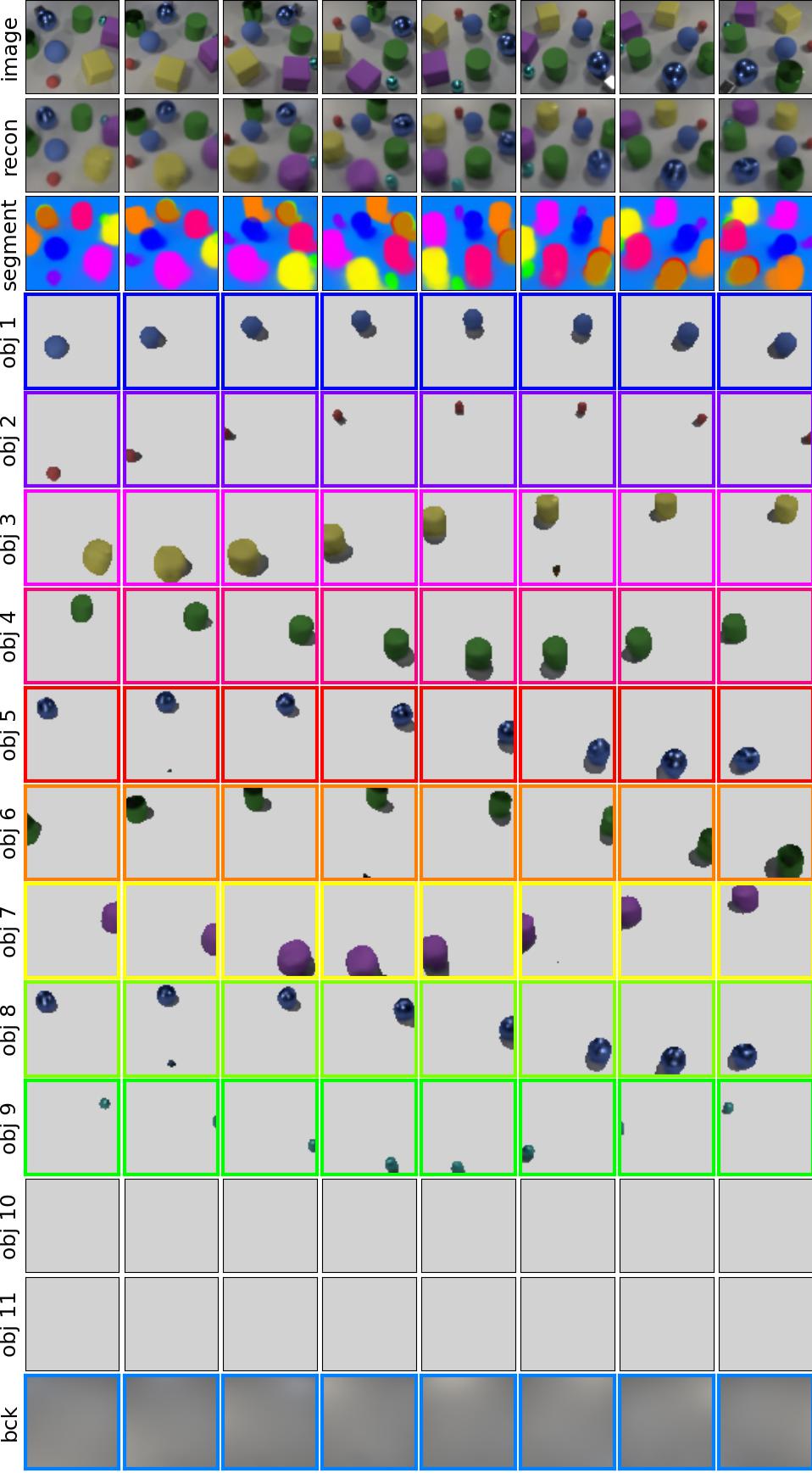}
    \end{minipage}
    }
    \subfigure[CLEVR-COMPLEX]{
    \begin{minipage}[a]{0.3\textwidth}
    \includegraphics[width=1\textwidth]{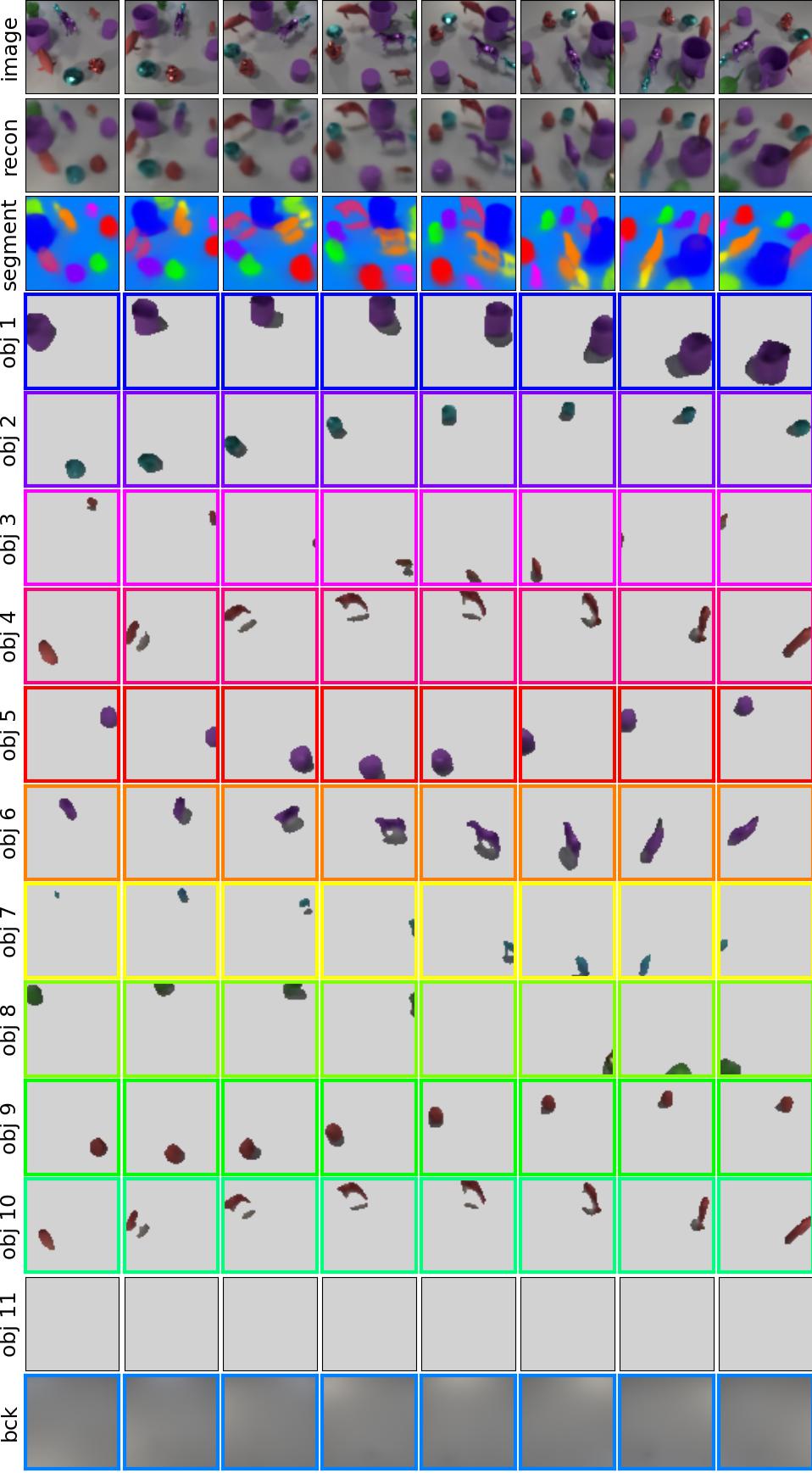}
    \end{minipage}
    }

    \centering
    \subfigure[SHOP-SIMPLE]{
    \begin{minipage}[a]{0.3\textwidth}
    \includegraphics[width=1\textwidth]{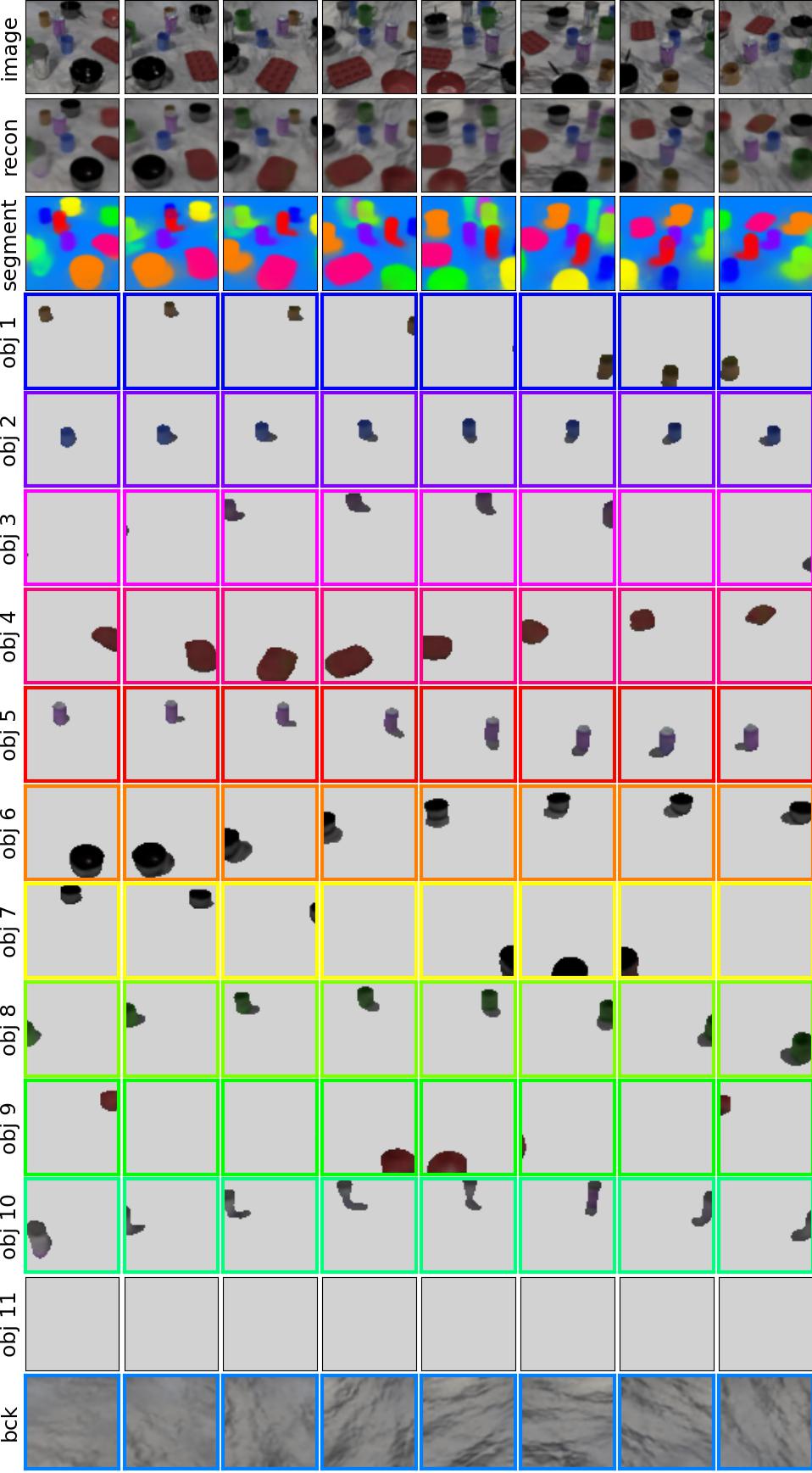}
    \end{minipage}
    }
    \subfigure[SHOP-COMPLEX]{
    \begin{minipage}[a]{0.3\textwidth}
    \includegraphics[width=1\textwidth]{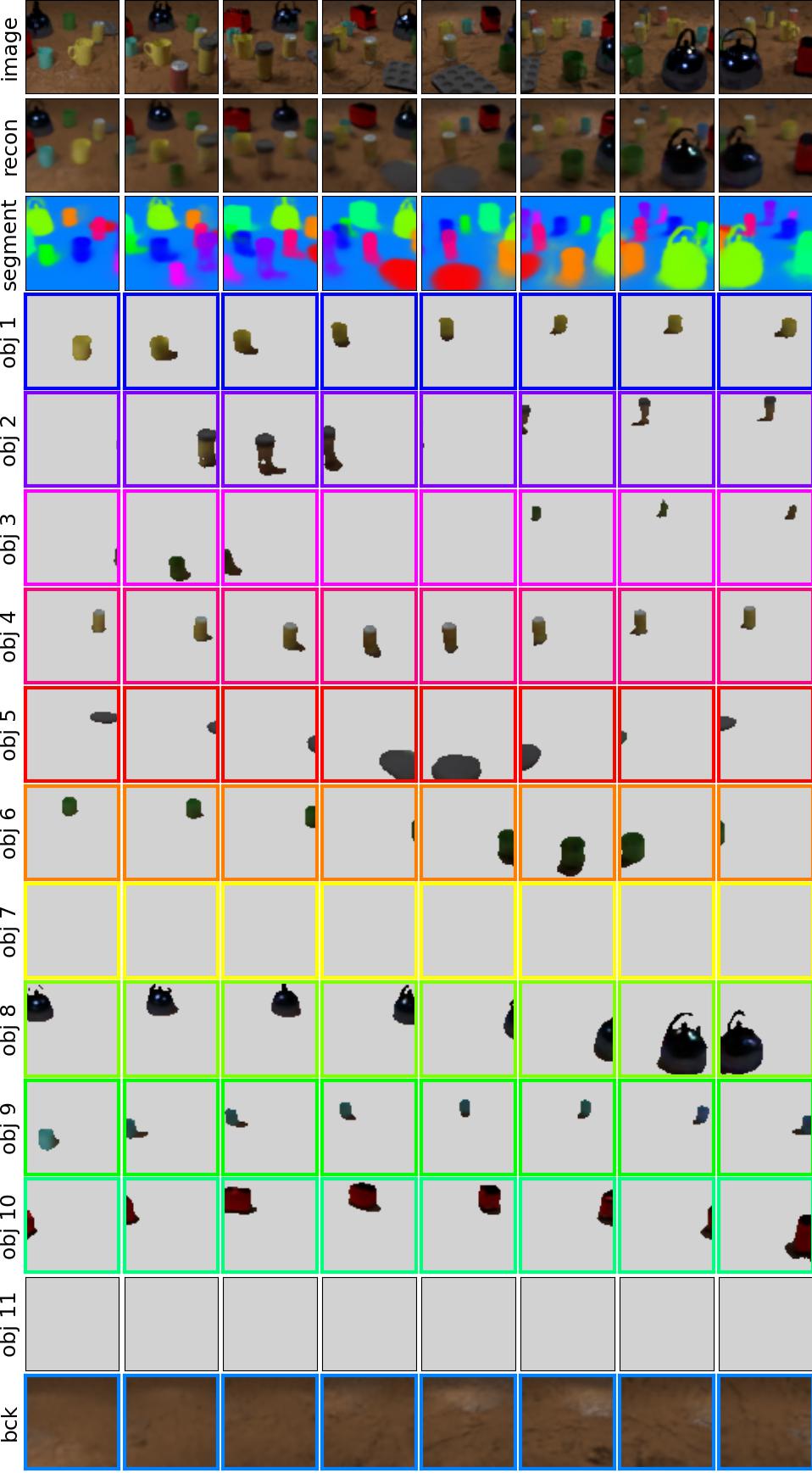}
    \end{minipage}
    }
    \caption{Visualization results on the general sets of dataset, observed views are 8}
    \label{fig:decompose_general}
\end{figure}

The performance of the model is evaluated quantitatively in terms of segmentation, complete shape, occlusion and object counting. Tables \ref{tab:segment_4views_test} and \ref{tab:segment_8views_test} demonstrate the comparison results in 4 views and 8 views, our proposed outperforms the remaining models in multiple aspects. Moreover, we compared the models in the generalization set with more objects and ;larger occlusion rate. Our model is still better than many unsupervised models, and can compete with MulMON with viewpoint annotations. Figure \ref{fig:decompose_general} describes the visualized results in general sets, we can find that the model performs well. And Tables \ref{tab:segment_4views_general} and \ref{tab:segment_8views_general} demonstrate the qualitative results of general sets.

\subsection{Prediction}
\begin{figure}
    \centering
    \subfigure[MulMON]{
    \begin{minipage}[a]{0.45\textwidth}
    \includegraphics[width=1\textwidth]{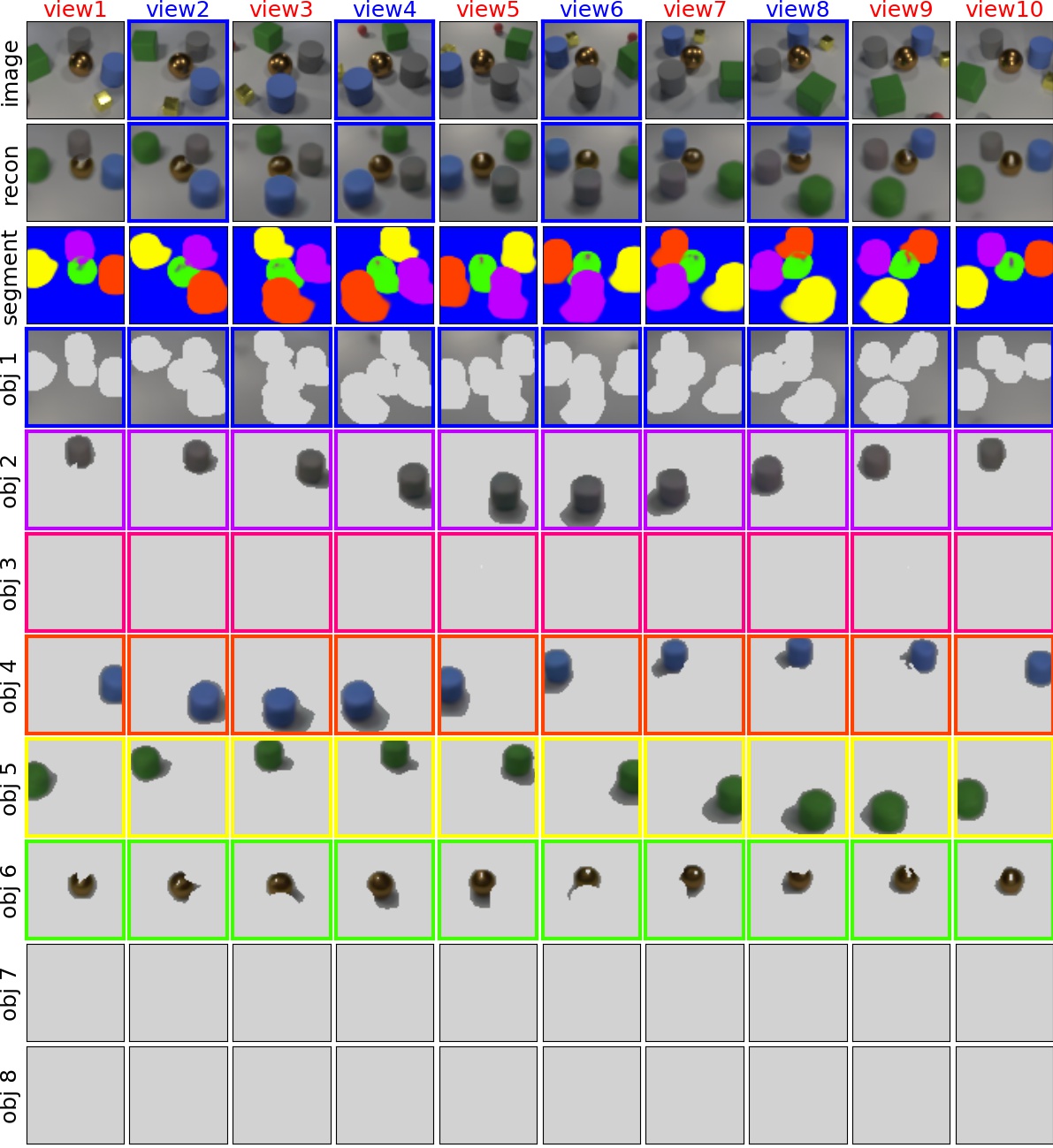}
    \end{minipage}
    }
    \subfigure[Ours]{
    \begin{minipage}[a]{0.45\textwidth}
    \includegraphics[width=1\textwidth]{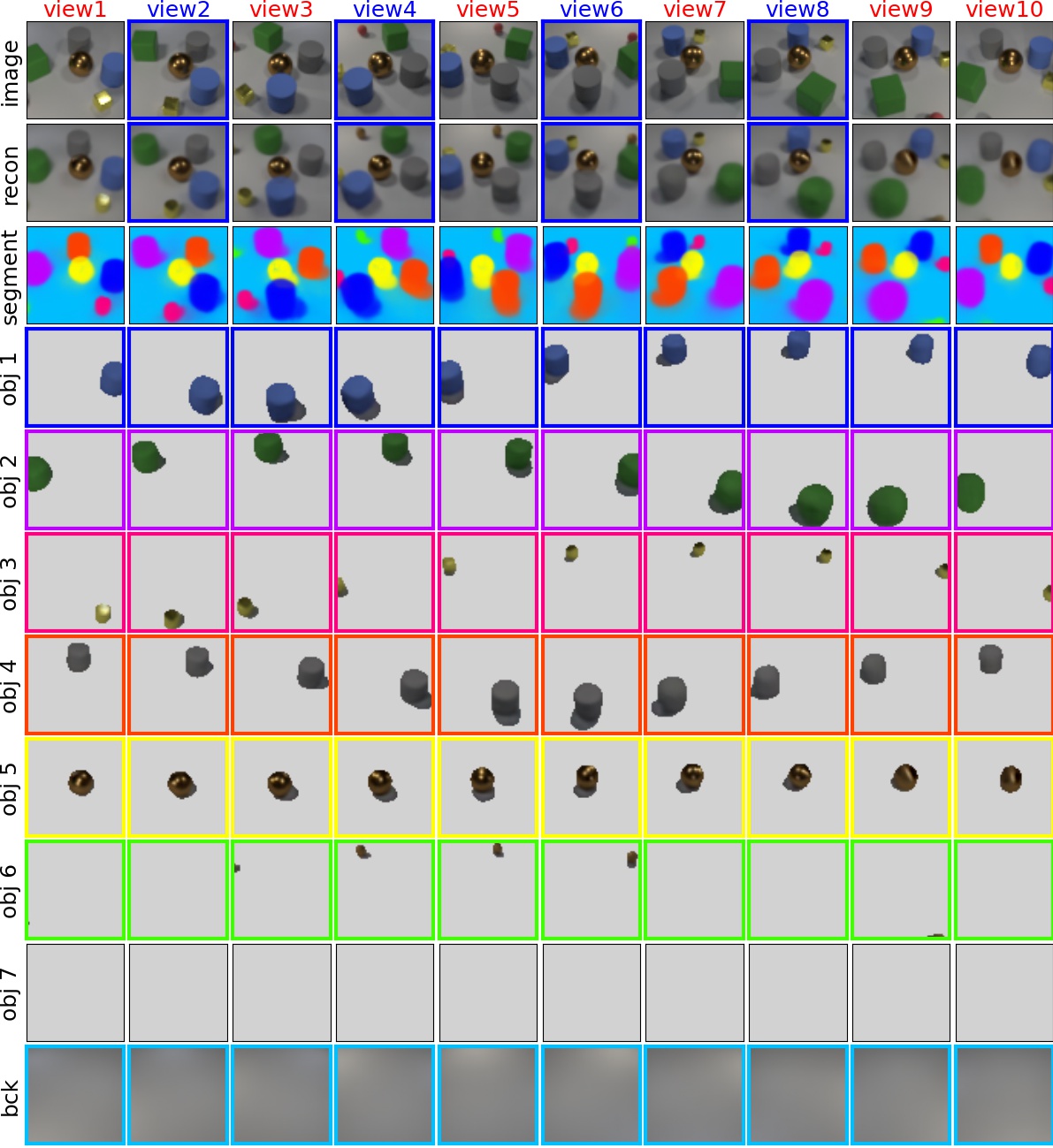}
    \end{minipage}
    }
    \caption{Qualitative comparison of prediction on the CLEVR-SIMPLE dataset. The observed views are 6, test mode is 1, query views are 4.}
    \label{fig:clevr_simple_m1_o6}
\end{figure}

\begin{figure}
    \centering
    \subfigure[MulMON]{
    \begin{minipage}[a]{0.45\textwidth}
    \includegraphics[width=1\textwidth]{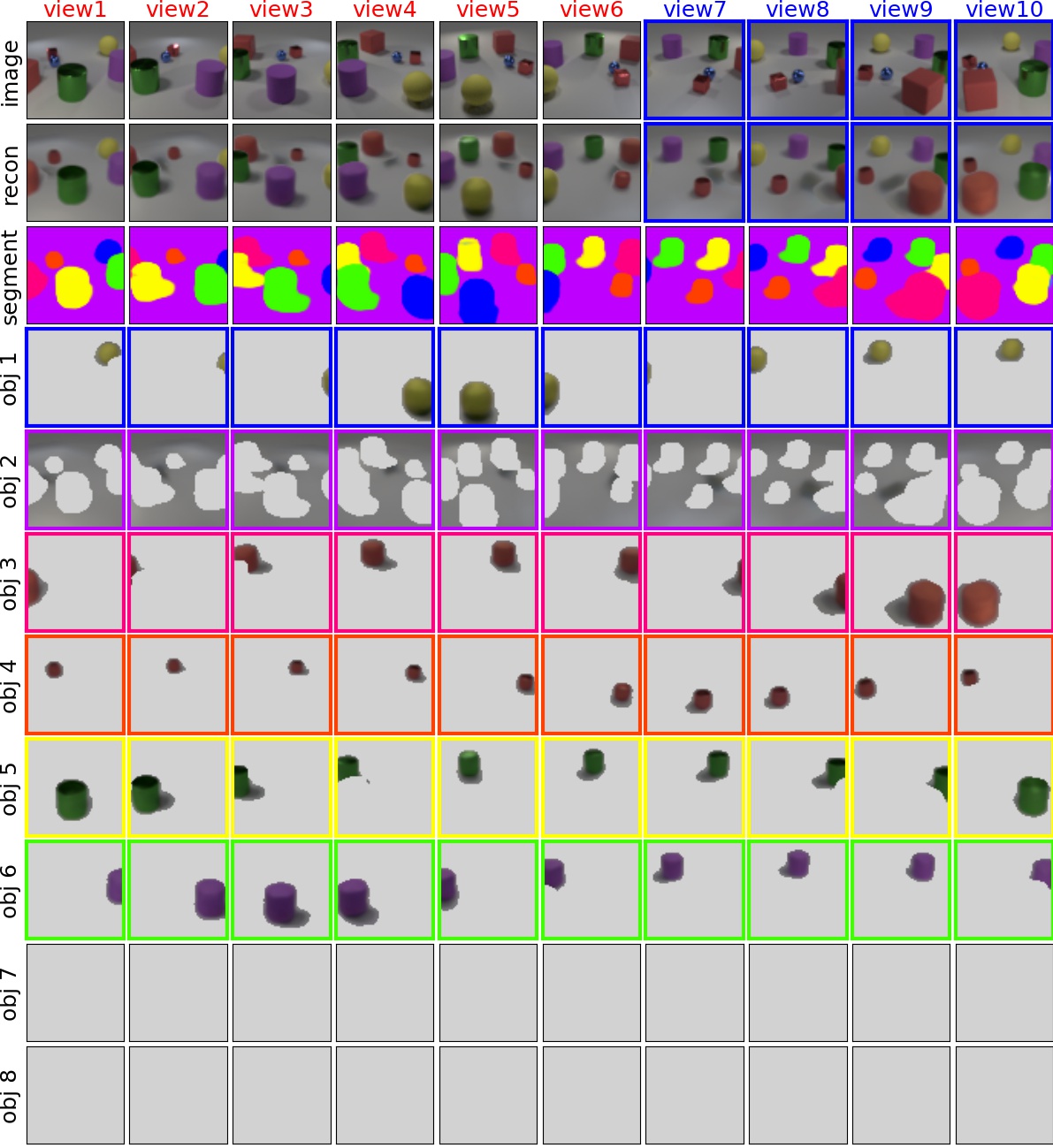}
    \end{minipage}
    }
    \subfigure[Ours]{
    \begin{minipage}[a]{0.45\textwidth}
    \includegraphics[width=1\textwidth]{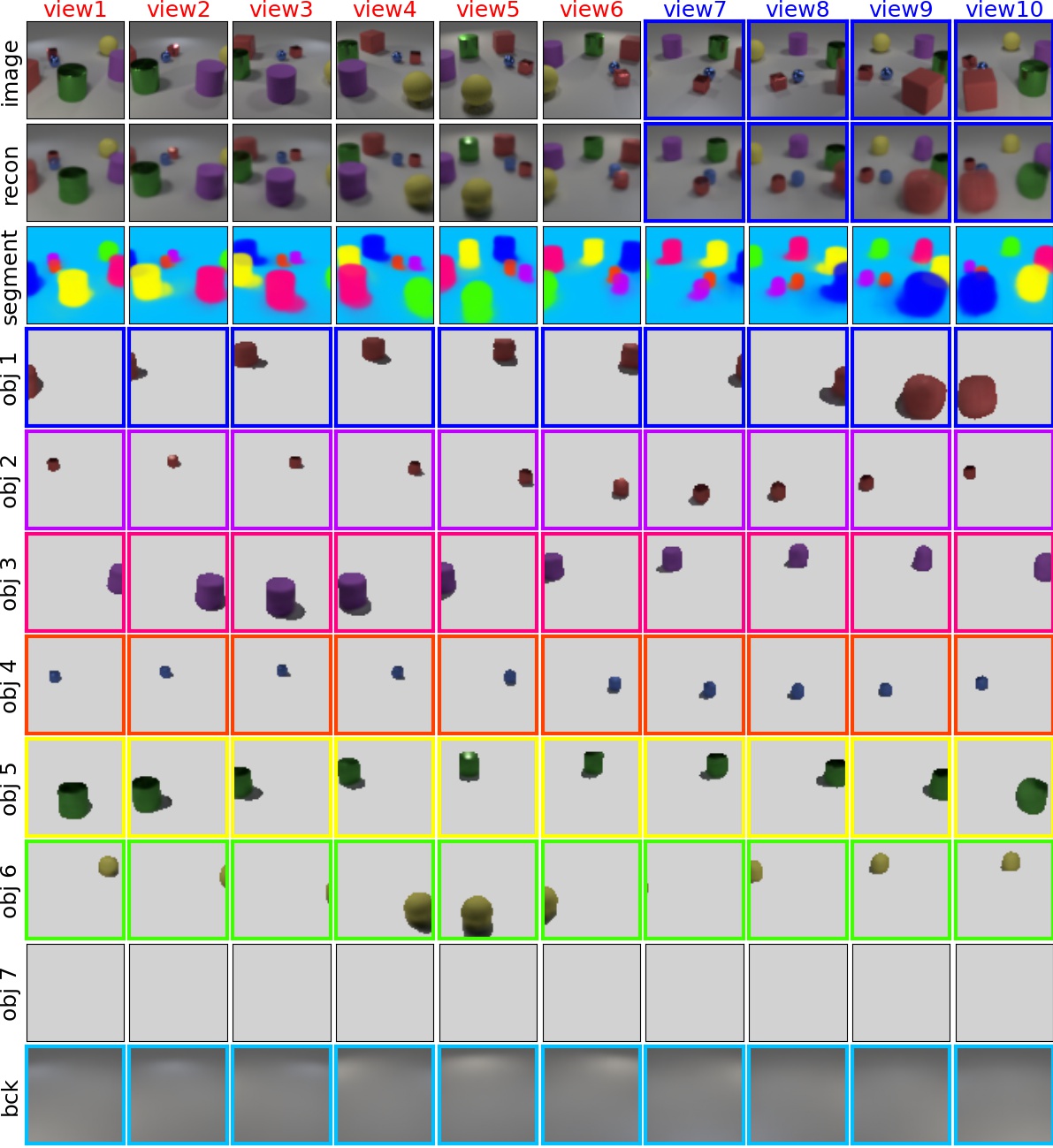}
    \end{minipage}
    }
    \caption{Qualitative comparison of prediction on the CLEVR-SIMPLE dataset. The observed views are 6, test mode is 2, query views are 4.}
    \label{fig:clevr_simple_m2_o6}
\end{figure}

\begin{figure}
    \centering
    \subfigure[MulMON]{
    \begin{minipage}[a]{0.48\textwidth}
    \includegraphics[width=1\textwidth]{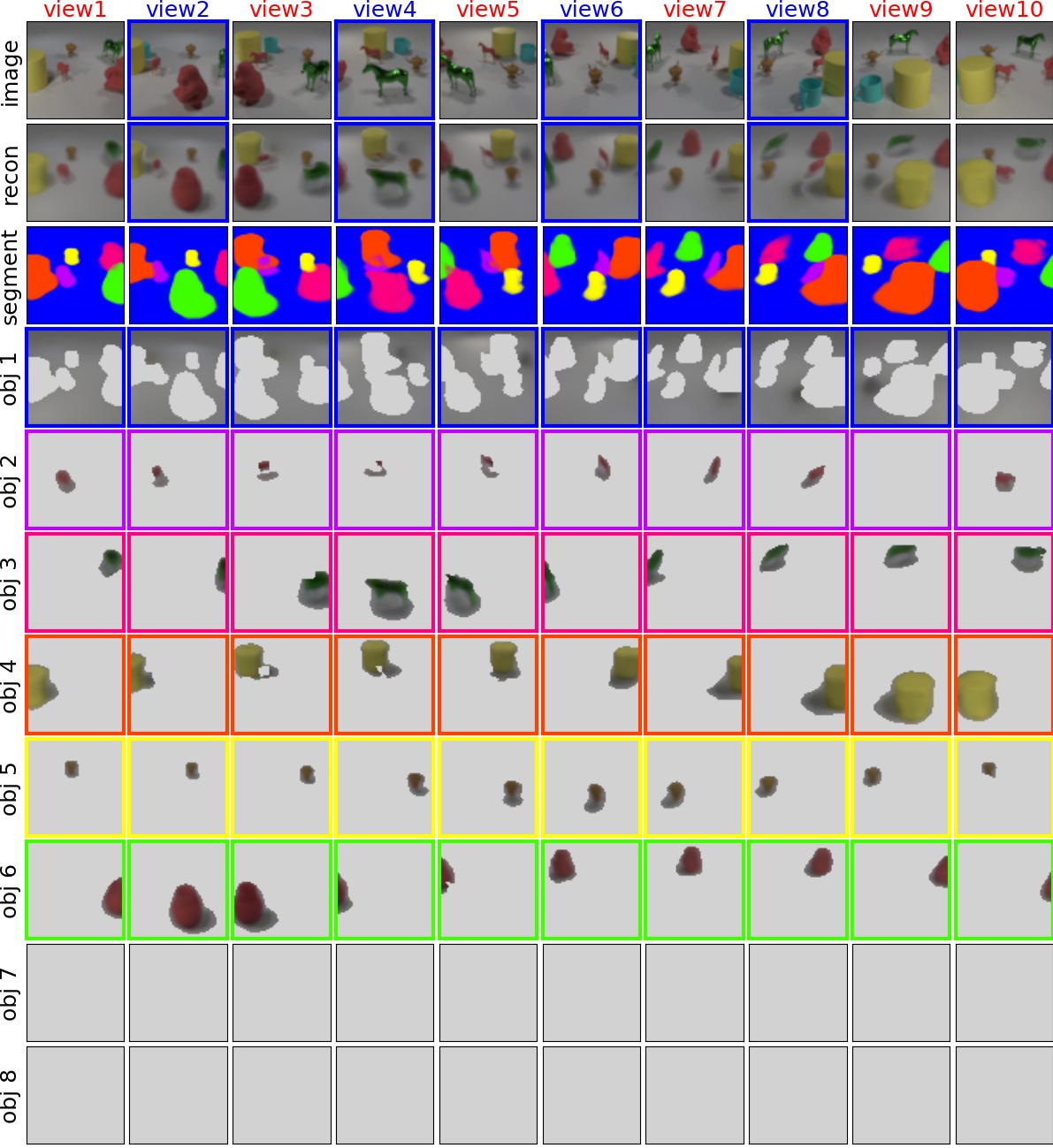}
    \end{minipage}
    }
    \subfigure[Ours]{
    \begin{minipage}[a]{0.48\textwidth}
    \includegraphics[width=1\textwidth]{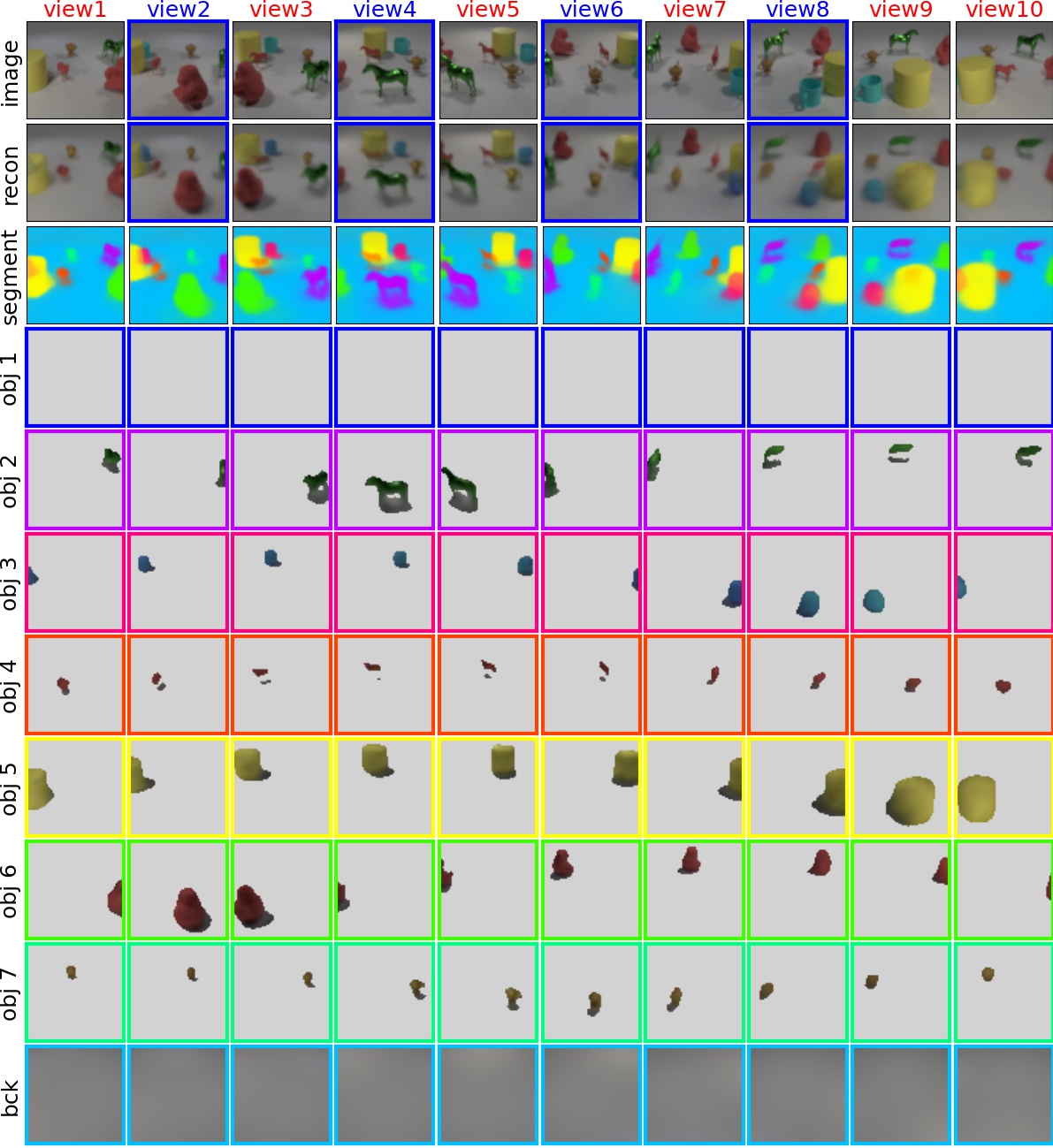}
    \end{minipage}
    }
    \caption{Qualitative comparison of prediction on the CLEVR-COMPLEX dataset. The observed views are 6, test mode is 1, query views are 4.}
    \label{fig:clevr_complex_m1_o6}
\end{figure}

\begin{figure}
    \centering
    \subfigure[MulMON]{
    \begin{minipage}[a]{0.48\textwidth}
    \includegraphics[width=1\textwidth]{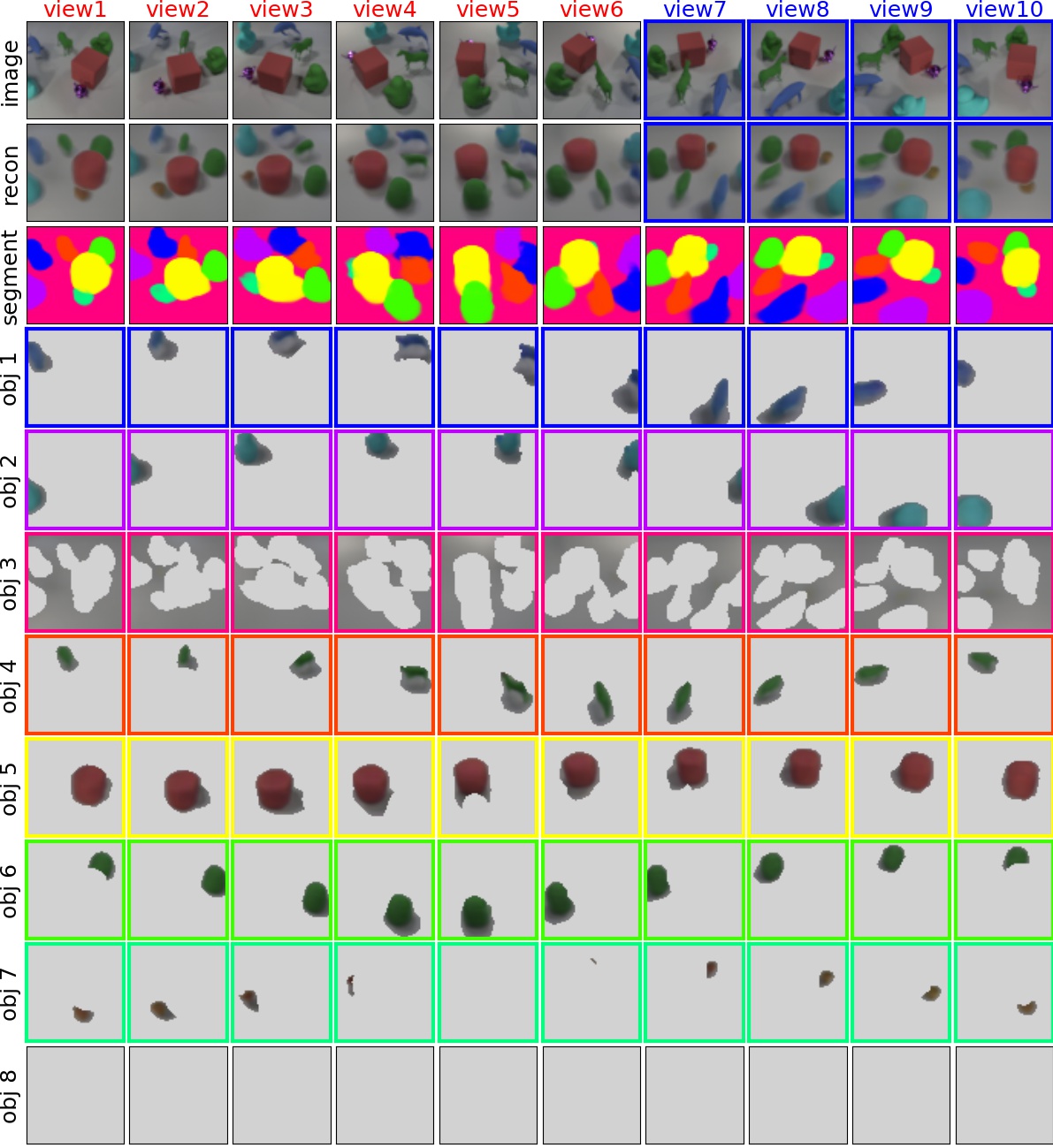}
    \end{minipage}
    }
    \subfigure[Ours]{
    \begin{minipage}[a]{0.48\textwidth}
    \includegraphics[width=1\textwidth]{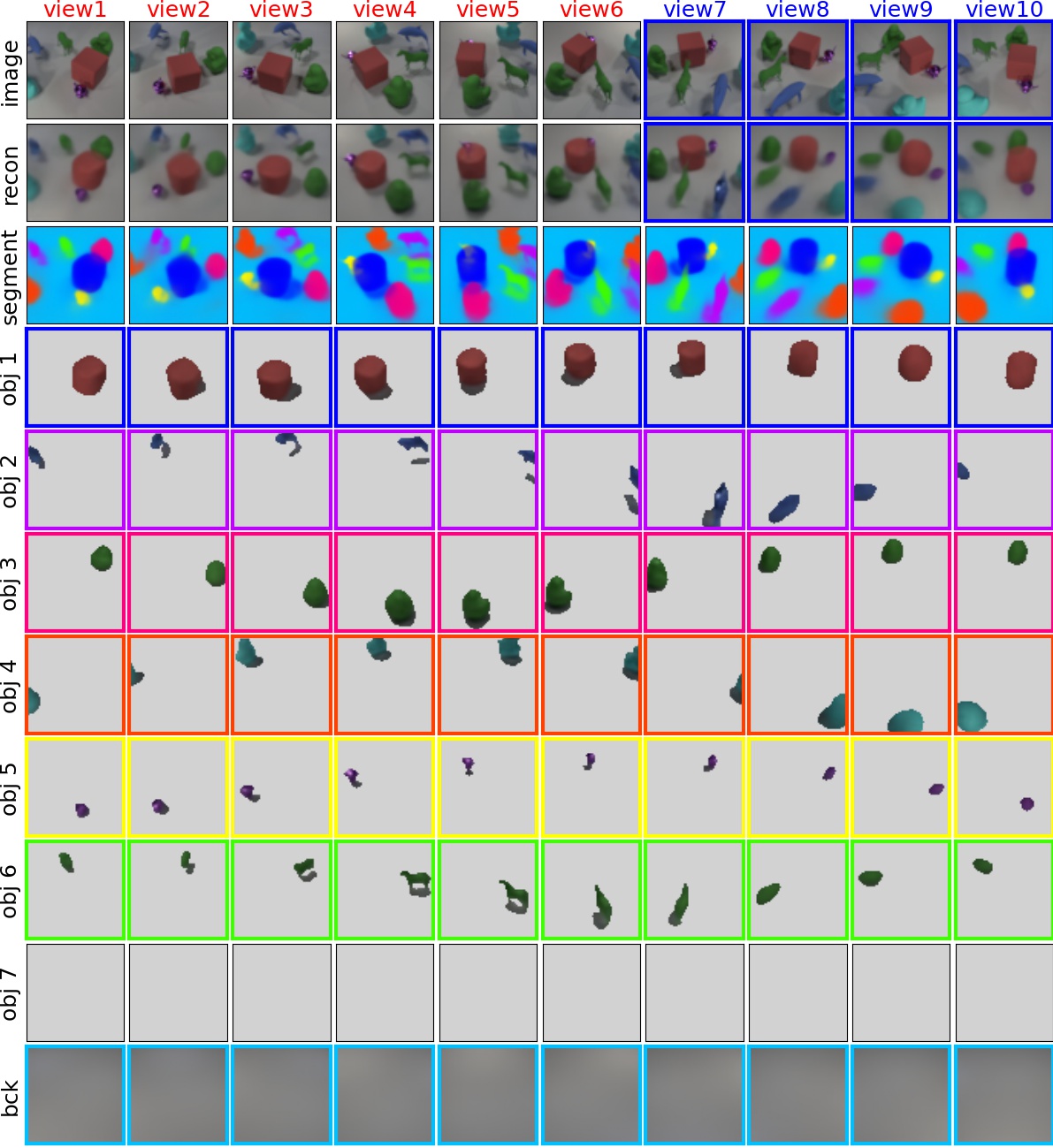}
    \end{minipage}
    }
    \caption{Qualitative comparison of prediction on the CLEVR-COMPLEX dataset. The observed views are 6, test mode is 2, query views are 4.}
    \label{fig:clevr_complex_m2_o6}
\end{figure}

\begin{figure}
    \centering
    \subfigure[MulMON]{
    \begin{minipage}[a]{0.48\textwidth}
    \includegraphics[width=1\textwidth]{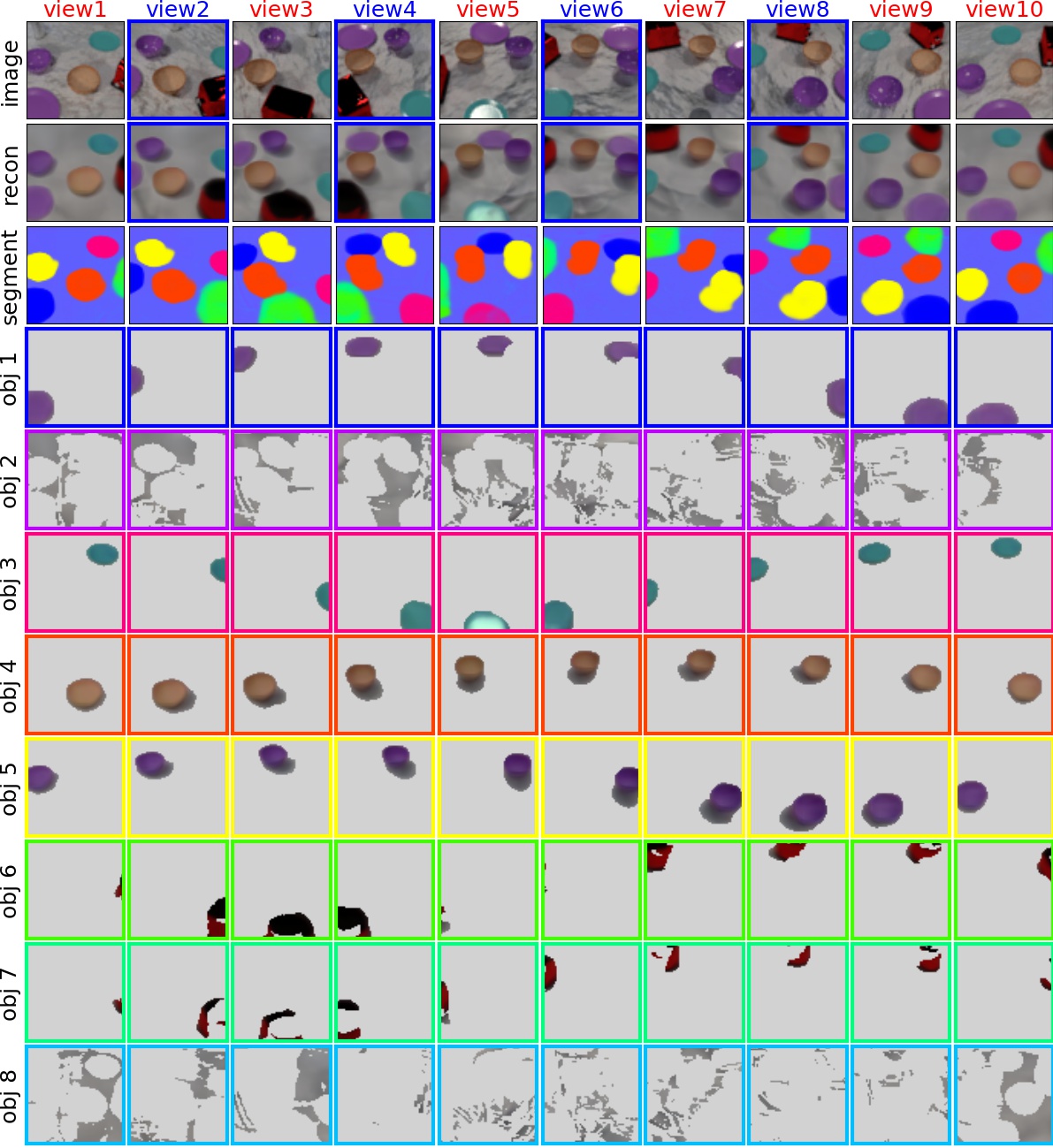}
    \end{minipage}
    }
    \subfigure[Ours]{
    \begin{minipage}[a]{0.48\textwidth}
    \includegraphics[width=1\textwidth]{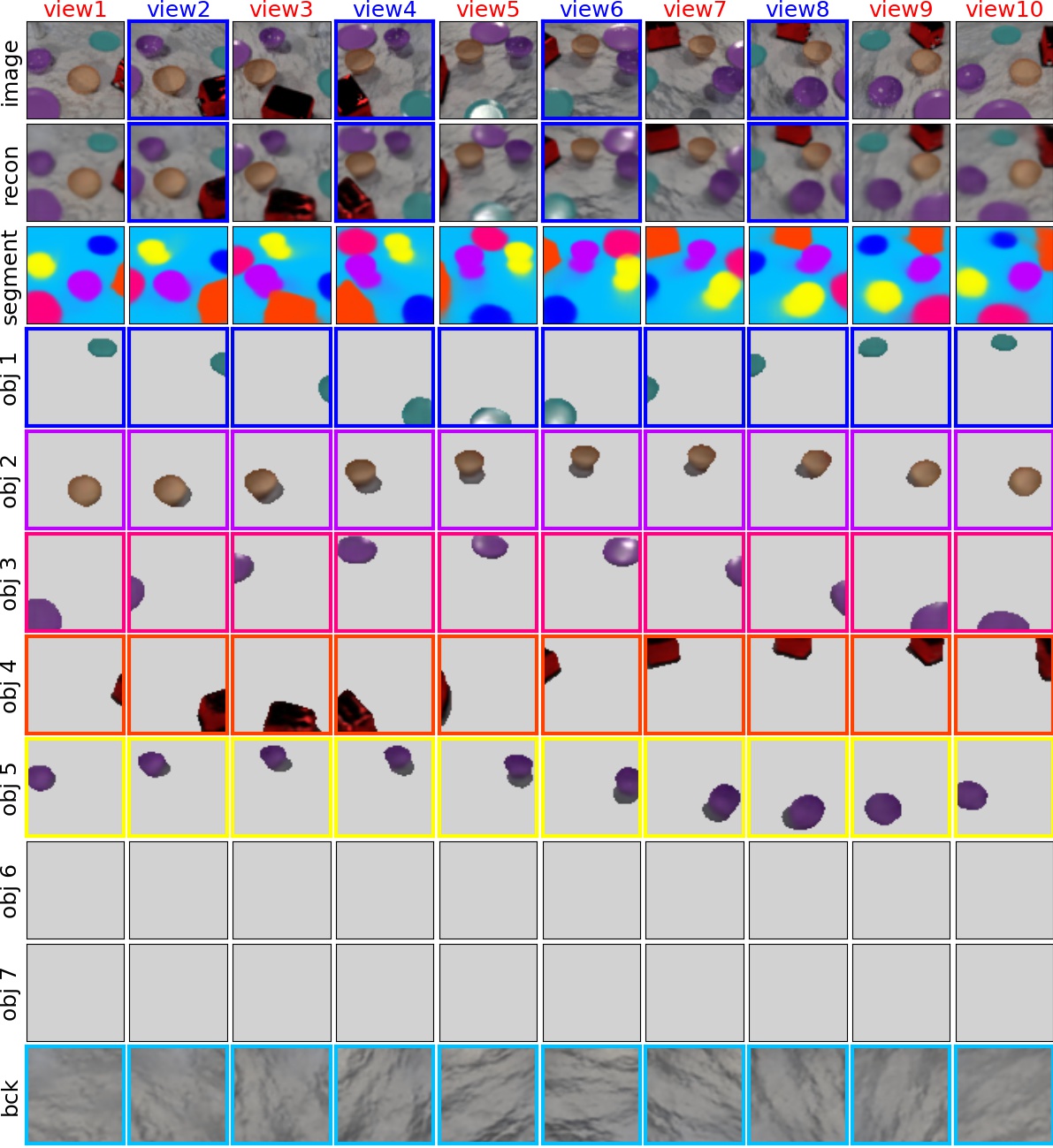}
    \end{minipage}
    }
    \caption{Qualitative comparison of prediction on the SHOP-SIMPLE dataset. The observed views are 6, test mode is 1, query views are 4.}
    \label{fig:shop_simple_m1_o6}
\end{figure}

\begin{figure}
    \centering
    \subfigure[MulMON]{
    \begin{minipage}[a]{0.48\textwidth}
    \includegraphics[width=1\textwidth]{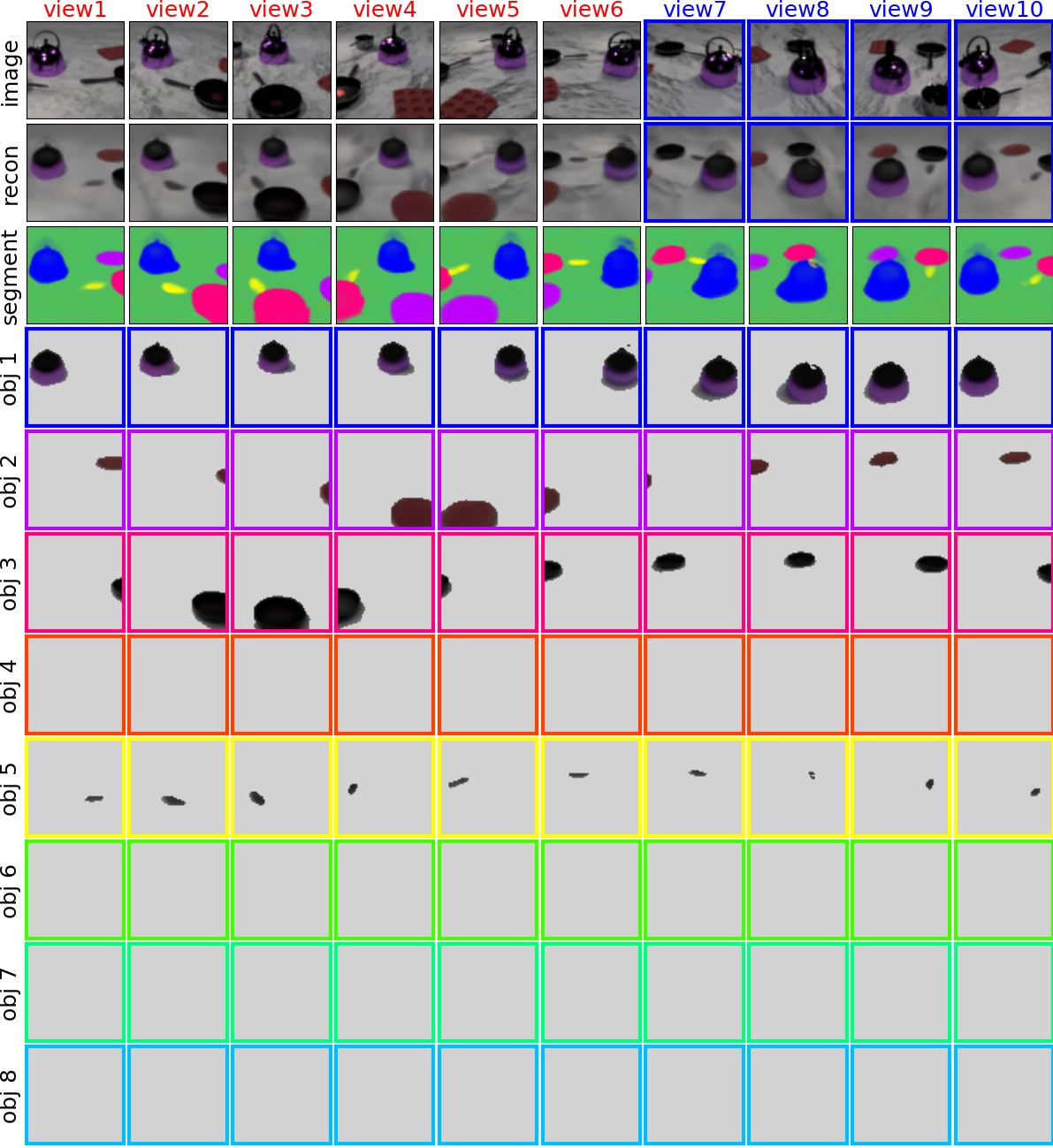}
    \end{minipage}
    }
    \subfigure[Ours]{
    \begin{minipage}[a]{0.48\textwidth}
    \includegraphics[width=1\textwidth]{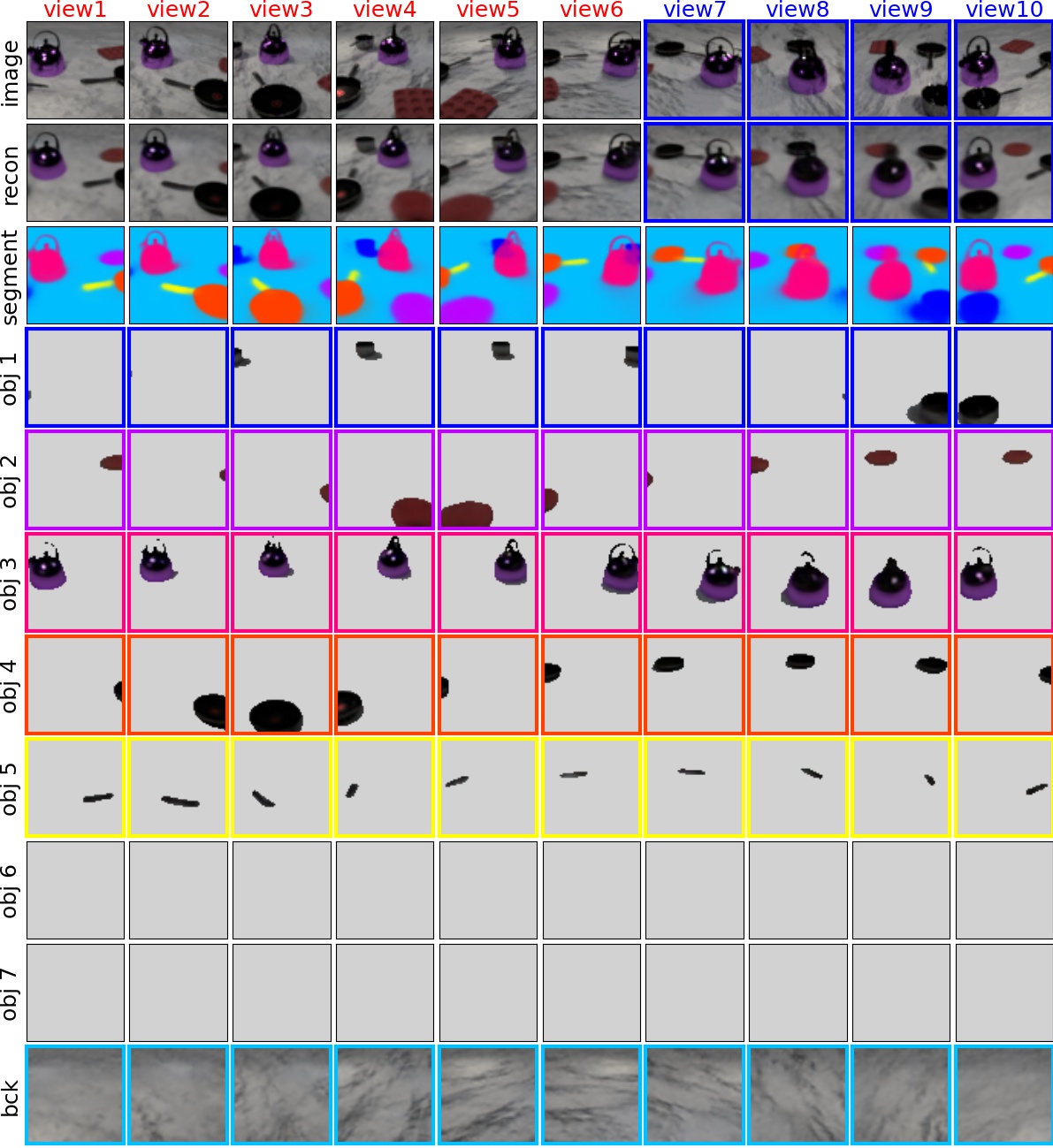}
    \end{minipage}
    }
    \caption{Qualitative comparison of prediction on the SHOP-SIMPLE dataset. The observed views are 6, test mode is 2, query views are 4.}
    \label{fig:shop_simple_m2_o6}
\end{figure}

\begin{figure}
    \centering
    \subfigure[MulMON]{
    \begin{minipage}[a]{0.48\textwidth}
    \includegraphics[width=1\textwidth]{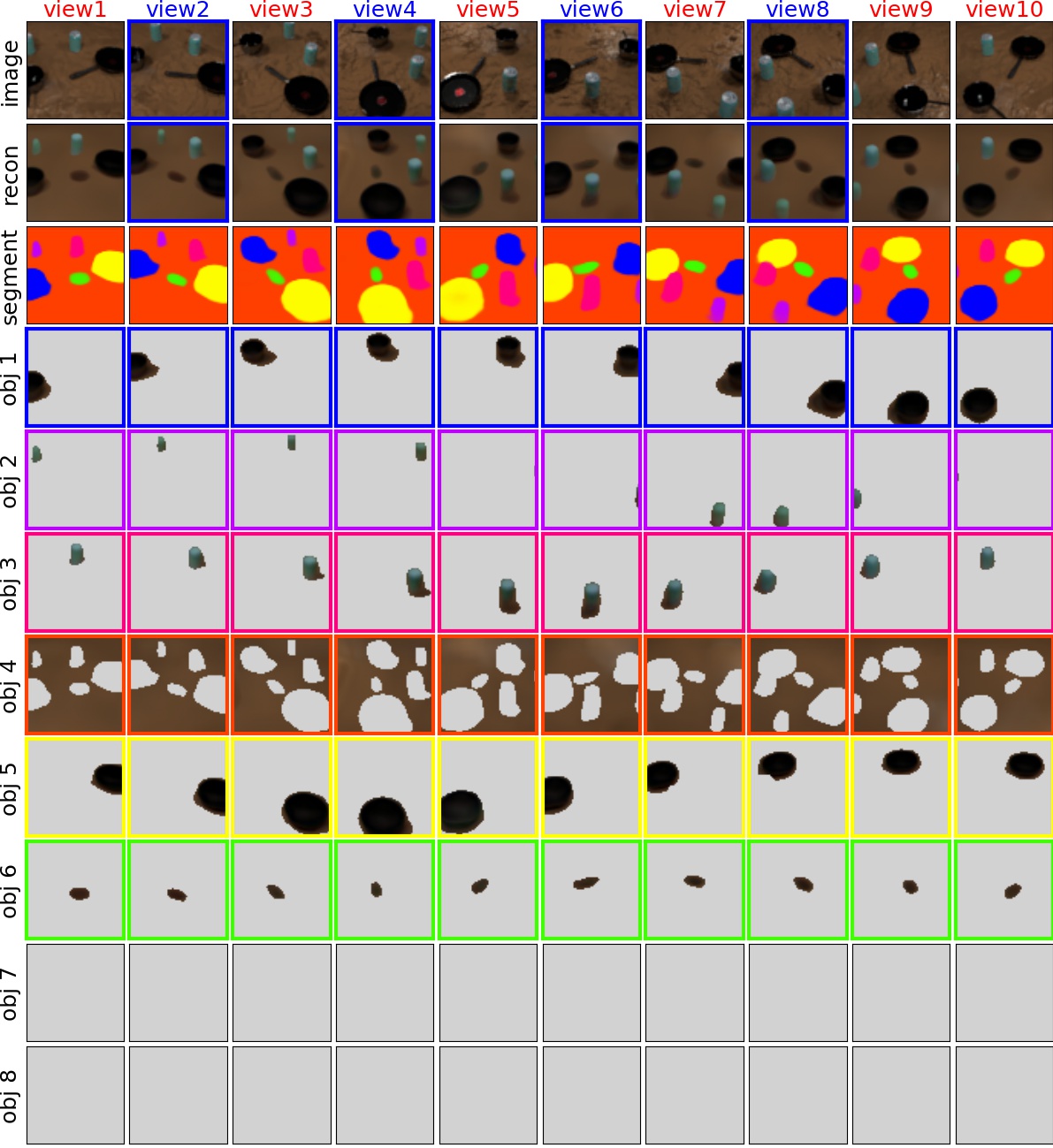}
    \end{minipage}
    }
    \subfigure[Ours]{
    \begin{minipage}[a]{0.48\textwidth}
    \includegraphics[width=1\textwidth]{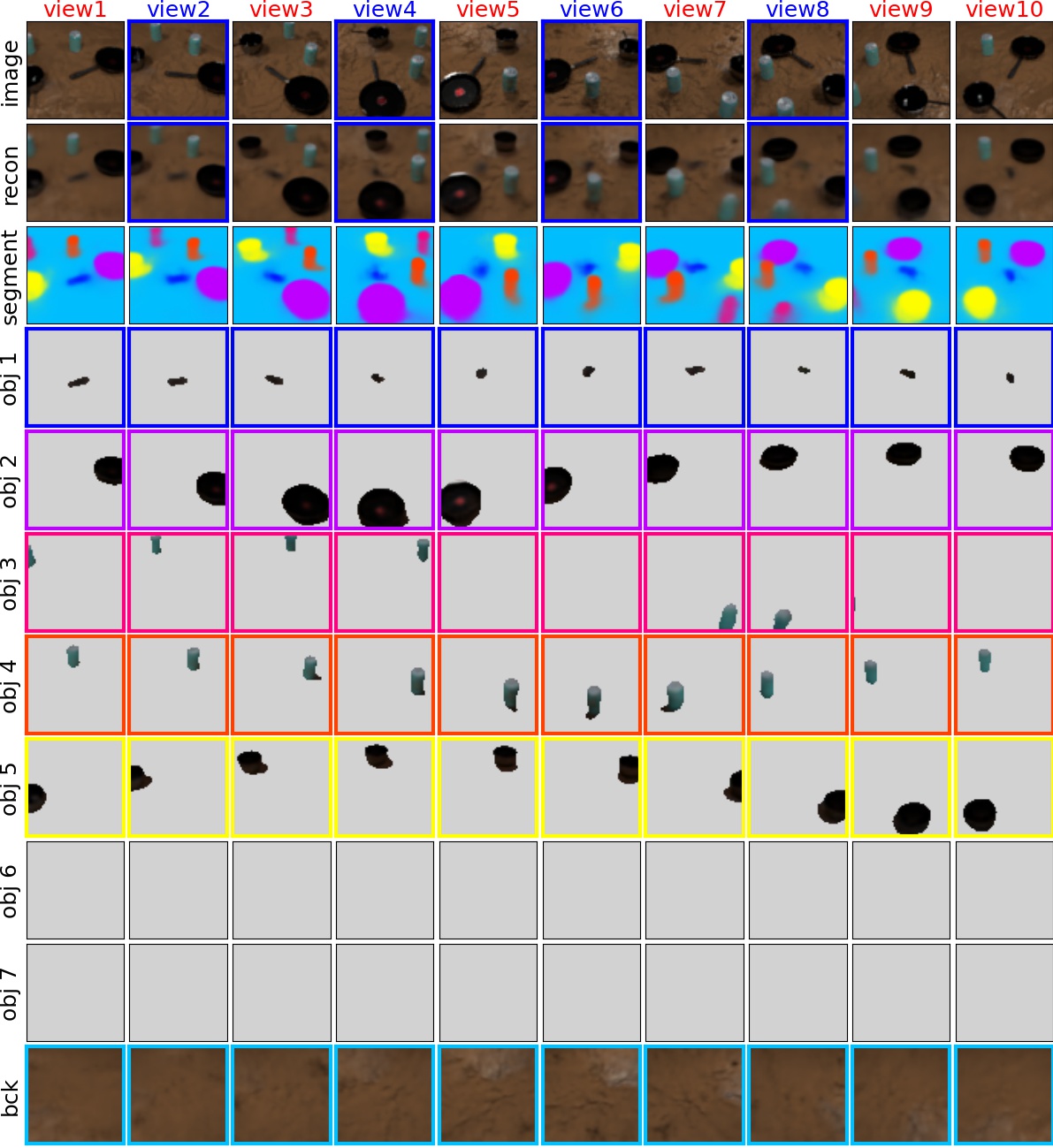}
    \end{minipage}
    }
    \caption{Qualitative comparison of prediction on the SHOP-COMPLEX dataset. The observed views are 6, test mode is 1, query views are 4.}
    \label{fig:shop_complex_m1_o6}
\end{figure}

\begin{figure}
    \centering
    \subfigure[MulMON]{
    \begin{minipage}[a]{0.48\textwidth}
    \includegraphics[width=1\textwidth]{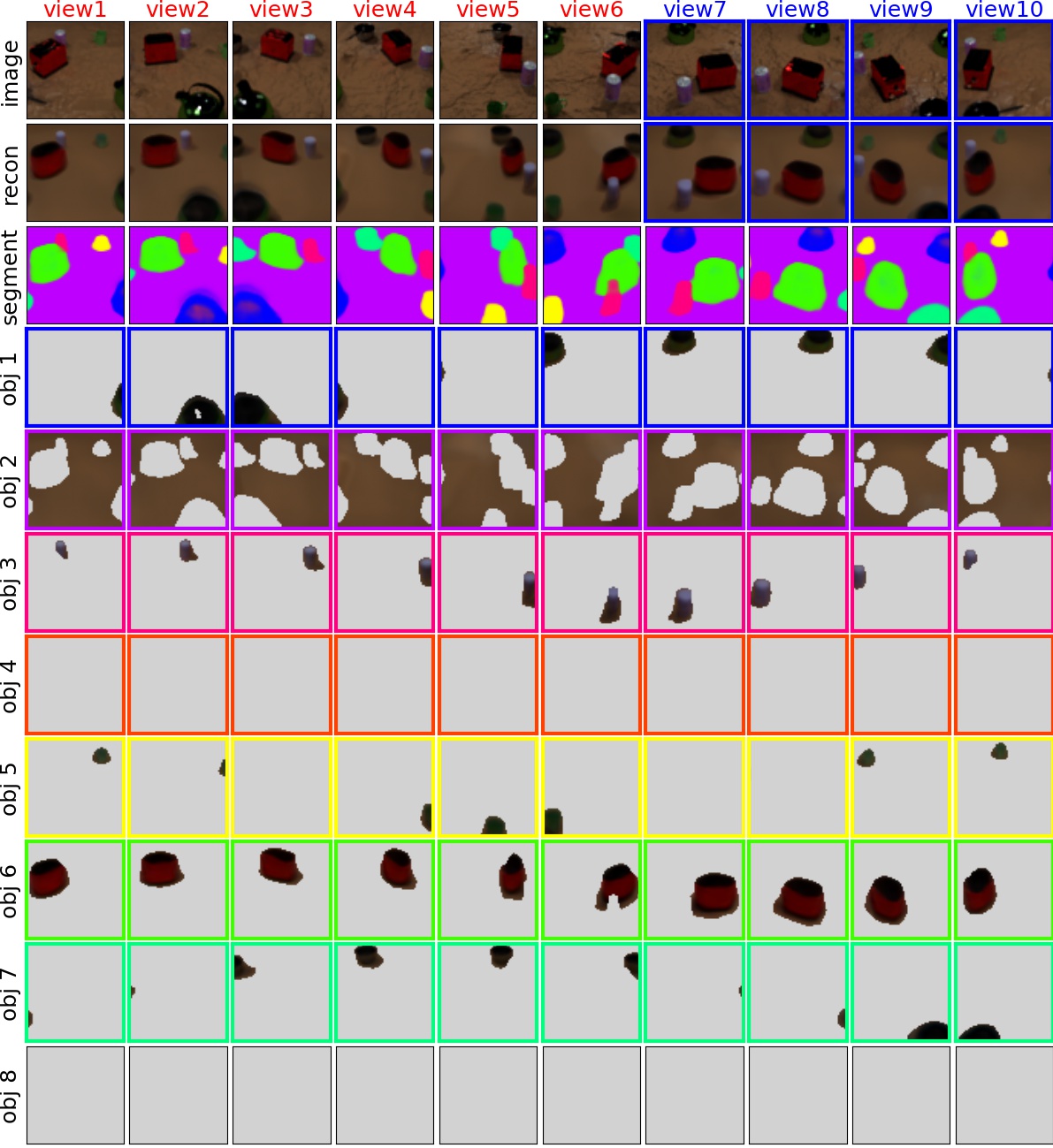}
    \end{minipage}
    }
    \subfigure[Ours]{
    \begin{minipage}[a]{0.48\textwidth}
    \includegraphics[width=1\textwidth]{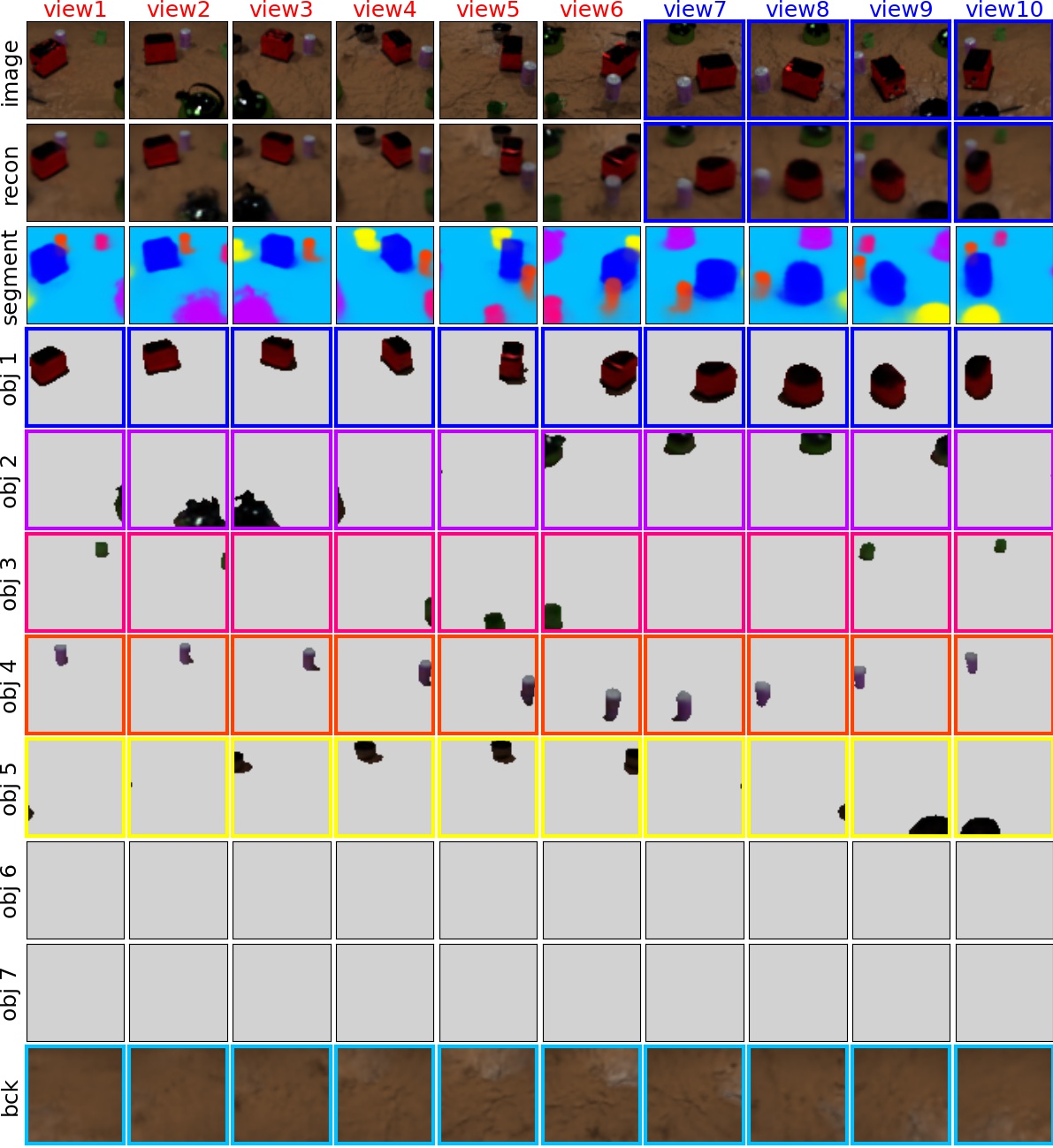}
    \end{minipage}
    }
    \caption{Qualitative comparison of prediction on the SHOP-COMPLEX dataset. The observed views are 6, test mode is 2, query views are 4.}
    \label{fig:shop_complex_m2_o6}
\end{figure}

We have fixed a number of viewpoints to make a fair comparison of prediction performance. Two tested mode called mode 1 and mode 2 are selected. In mode 1, the predicted viewpoints are inserted into the observed viewpoints. And in mode 2, the predicted viewpoints are completely out of the middle. For the two modes, we tested the prediction performance with 6/7/8/9 observed views. Figures \ref{fig:clevr_simple_m1_o6}, \ref{fig:clevr_complex_m1_o6}, \ref{fig:shop_simple_m1_o6} and \ref{fig:shop_complex_m1_o6} demonstrate the prediciton results of four datasets testing with mode 1. In addition to the prediction of novel viewpoints, our model can also recontrcution additional occlusion completion, which MulMON cannot do. Figures \ref{fig:clevr_simple_m2_o6}, \ref{fig:clevr_complex_m2_o6}, \ref{fig:shop_simple_m2_o6} and \ref{fig:shop_complex_m2_o6} are tested with mode 2. From the prediciton results, we can see that the farther away from the point of GP function, the worse the reconstruction performance will be.

We evaluate the qualitative results from different observed view numbers. Tables \ref{tab:predict_m1_o6}, \ref{tab:predict_m1_o7}, \ref{tab:predict_m1_o8}, \ref{tab:predict_m1_o9}, \ref{tab:predict_m2_o6}, \ref{tab:predict_m2_o7}, \ref{tab:predict_m2_o8} and \ref{tab:predict_m2_o9} show the qualitative results on multiple aspects. With more GP points, our model is getting better and better in fitting function. When the observd view number is only 6, our model is slightly worse than MulMON, while when the observed view number becomes more (such as up to 8), our method can make better predictions due to better function fitting, so it is better than MulMON in multiple metrics.

\begin{table*}[ht]
    \centering
    \scalebox{0.75}{
    \begin{tabular}{c|c|cccccccc} 
    \toprule[1.5pt]
    
    Dataset&Method&ARI-A$\uparrow$&AMI-A$\uparrow$&ARI-O$\uparrow$&AMI-O$\uparrow$&IoU$\uparrow$&F1$\uparrow$&OCA$\uparrow$&OOA$\uparrow$\\
    \midrule[0.5pt] \multirow{4}{*}{CLEVR-SIMPLE}&MulMON& 0.658$\pm$1e-3 & 0.603$\pm$1e-3 & 0.969$\pm$1e-3 & 0.956$\pm$1e-3 & 0.615$\pm$4e-3 & 0.741$\pm$4e-3 & 0.606$\pm$4e-2 & N/A\\ 
    ~&SIMONe & 0.086$\pm$5e-5 & 0.313$\pm$9e-5 &0.947$\pm$1e-4 & 0.924$\pm$2e-4 & 0.449$\pm$1e-4 & 0.601$\pm$2e-4 & 0.000$\pm$0e-0& N/A \\ 
    ~&OCLOC&0.541$\pm$2e-3& 0.512$\pm$2e-3 & 0.935$\pm$5e-3 & 0.930$\pm$4e-3 & 0.475$\pm$4e-3 & 0.629$\pm$4e-3 & 0.532$\pm$3e-2 & 0.955$\pm$1e-2\\
    ~&Ours& \bfseries0.830$\pm$3e-3 & \bfseries0.736$\pm$3e-3 & \bfseries0.973$\pm$7e-3 & \bfseries0.968$\pm$4e-3 &\bfseries 0.656$\pm$3e-3 &\bfseries 0.781$\pm$ 4e-3 & \bfseries0.704$\pm$3e-2 & \bfseries0.968$\pm$1e-2\\
    \midrule[0.5pt] \multirow{4}{*}{CLEVR-COMPLEX}&MulMON&0.552$\pm$9e-3& 0.533$\pm$4e-3 & 0.941$\pm$3e-3 & 0.923$\pm$2e-3 & 0.554$\pm$3e-3 & 0.689$\pm$4e-3 & 0.612$\pm$3e-2 & N/A\\ 
    ~&SIMONe & 0.073$\pm$3e-5 & 0.299$\pm$8e-5 & 0.939$\pm$2e-4 & 0.912$\pm$3e-4 & 0.396$\pm$5e-5 & 0.547$\pm$6e-5 & 0.000$\pm$0e-0 & N/A\\ 
    ~&OCLOC&0.396$\pm$1e-3 & 0.419$\pm$1e-3 & 0.915$\pm$4e-3 & 0.905$\pm$4e-3 & 0.375$\pm$3e-3 & 0.523$\pm$3e-3 & 0.676$\pm$2e-2 & 0.917$\pm$1e-2\\
    ~&Ours& \bfseries0.759$\pm$2e-3 & \bfseries0.657$\pm$3e-3 &\bfseries 0.963$\pm$4e-3 & \bfseries0.959$\pm$3e-3 & \bfseries0.569$\pm$6e-3 & \bfseries0.708$\pm$7e-3 & \bfseries0.694$\pm$2e-2 & \bfseries0.952$\pm$1e-2\\
    \midrule[0.5pt] \multirow{4}{*}{SHOP-SIMPLE}&MulMON& 0.435$\pm$2e-2 & 0.539$\pm$8e-3 & 0.894$\pm$5e-3 & 0.878$\pm$2e-3 & 0.596$\pm$9e-3 & 0.725$\pm$9e-3 & 0.148$\pm$4e-2 & N/A\\ 
    ~&SIMONe & 0.201$\pm$2e-4 & 0.437$\pm$2e-4 & 0.757$\pm$1e-4 & 0.805$\pm$1e-4 & 0.488$\pm$7e-5 & 0.633$\pm$7e-5 & 0.000$\pm$0e-0 & N/A\\  
    ~&OCLOC&0.650$\pm$4e-3 & 0.607$\pm$4e-3 & 0.918$\pm$6e-3 & 0.910$\pm$4e-3 & 0.609$\pm$4e-3 & 0.737$\pm$5e-3 & 0.448$\pm$5e-2 & 0.695$\pm$2e-2\\
    ~&Ours& \bfseries0.816$\pm$2e-3 & \bfseries0.739$\pm$2e-3 & \bfseries0.957$\pm$2e-3 & \bfseries0.954$\pm$1e-3 & \bfseries0.668$\pm$3e-3 & \bfseries0.780$\pm$3e-3 & \bfseries0.528$\pm$8e-2 & \bfseries0.790$\pm$2e-2\\
    \midrule[0.5pt] \multirow{4}{*}{SHOP-COMPLEX}&MulMON& 0.599$\pm$2e-2 & 0.595$\pm$6e-3 & 0.872$\pm$4e-3 & 0.863$\pm$2e-3 & 0.630$\pm$4e-3 & 0.751$\pm$4e-3 & 0.314$\pm$4e-2 & N/A\\ 
    ~&SIMONe& 0.185$\pm$6e-5 & 0.443$\pm$8e-5 & 0.796$\pm$7e-5 & 0.840$\pm$9e-5 & 0.535$\pm$1e-4 & 0.675$\pm$1e-4 & 0.000$\pm$0e-0 & N/A \\ 
    ~&OCLOC&0.342$\pm$1e-3 & 0.305$\pm$1e-3 & 0.380$\pm$5e-3 & 0.495$\pm$4e-3 & 0.249$\pm$3e-3 & 0.360$\pm$4e-3 & 0.160$\pm$4e-2 & 0.624$\pm$2e-2 \\
    ~&Ours&\bfseries0.796$\pm$4e-3 & \bfseries0.714$\pm$3e-3 & \bfseries0.946$\pm$7e-3 & \bfseries0.941$\pm$4e-3 & \bfseries0.654$\pm$4e-3 & \bfseries0.771$\pm$4e-3 &\bfseries0.518$\pm$2e-2 & \bfseries0.852$\pm$8e-3\\
    \bottomrule[1.5pt]
    \end{tabular} 
    }
    \caption{The comparison results of multiple aspects on test sets (training on 4 views and testing on 4 views). All test values are evaluated 5 times, recorded with mean and standard deviation.}
    \label{tab:segment_4views_test}
\end{table*}
\begin{table*}[ht]
    \centering
    \scalebox{0.75}{
    \begin{tabular}{c|c|cccccccc} 
    \toprule[1.5pt]
    
    Dataset&Method&ARI-A$\uparrow$&AMI-A$\uparrow$&ARI-O$\uparrow$&AMI-O$\uparrow$&IoU$\uparrow$&F1$\uparrow$&OCA$\uparrow$&OOA$\uparrow$\\
    \midrule[0.5pt] \multirow{4}{*}{CLEVR-SIMPLE}&MulMON& 0.584$\pm$8e-4 & 0.606$\pm$9e-4 & \bfseries0.939$\pm$2e-3 & \bfseries0.933$\pm$1e-3 & 0.542$\pm$3e-3 & 0.671$\pm$4e-3 & \bfseries0.440$\pm$5e-2 & N/A\\ 
    ~&SIMONe& 0.111$\pm$6e-5 & 0.409$\pm$1e-4 & 0.912$\pm$3e-4 & 0.885$\pm$3e-4 & 0.430$\pm$8e-5 & 0.573$\pm$8e-5 & 0.000$\pm$0e-0 & N/A\\ 
    ~&OCLOC&0.406$\pm$2e-3 & 0.489$\pm$3e-3 & 0.863$\pm$6e-3 & 0.872$\pm$4e-3 & 0.397$\pm$5e-3  & 0.541$\pm$7e-3 & 0.250$\pm$3e-2 & 0.897$\pm$8e-3 \\
    ~&Ours& \bfseries0.763$\pm$8e-4 & \bfseries0.706$\pm$1e-3 & 0.931$\pm$2e-3 & 0.931$\pm$1e-3 & \bfseries0.569$\pm$3e-3 & \bfseries0.691$\pm$3e-3 & 0.390$\pm$6e-2 & \bfseries0.936$\pm$8e-3\\
    \midrule[0.5pt] \multirow{4}{*}{CLEVR-COMPLEX}&MulMON& 0.477$\pm$3e-3 & 0.539$\pm$7e-4 & 0.906$\pm$2e-3 &  0.897$\pm$1e-3 & 0.469$\pm$1e-3 & 0.601$\pm$2e-3 & 0.326$\pm$4e-2 & N/A\\ 
    ~&SIMONe& 0.090$\pm$3e-5& 0.392$\pm$6e-5 & 0.914$\pm$2e-4 & 0.887$\pm$2e-4 & 0.387$\pm$5e-5 & 0.528$\pm$5e-5 & 0.000$\pm$0e-0 & N/A\\ 
    ~&OCLOC&0.187$\pm$9e-4 & 0.388$\pm$1e-3 & 0.829$\pm$6e-3 & 0.845$\pm$3e-3 & 0.290$\pm$8e-4 & 0.424$\pm$1e-3 & 0.316$\pm$2e-2 & 0.853$\pm$4e-3\\
    ~&Ours& \bfseries0.676$\pm$2e-3 & \bfseries0.630$\pm$3e-3 &\bfseries 0.917$\pm$6e-3 & \bfseries0.919$\pm$4e-3 & \bfseries0.496$\pm$6e-3 & \bfseries0.628$\pm$7e-3 & \bfseries0.390$\pm$4e-2 & \bfseries0.917$\pm$8e-3\\
    \midrule[0.5pt] \multirow{4}{*}{SHOP-SIMPLE}&MulMON& 0.509$\pm$9e-3  & 0.590$\pm$3e-3 & 0.871$\pm$3e-3 & 0.873$\pm$1e-3 & 0.565$\pm$4e-3 & \bfseries0.694$\pm$5e-3 & 0.316$\pm$5e-2 & N/A\\ 
    ~&SIMONe& 0.200$\pm$9e-5 & 0.454$\pm$8e-5 & 0.709$\pm$1e-4 & 0.763$\pm$1e-4 & 0.396$\pm$2e-4 & 0.527$\pm$2e-4 & 0.000$\pm$0e-0 & N/A\\ 
    ~&OCLOC&0.459$\pm$3e-3 & 0.525$\pm$3e-3 & 0.817$\pm$6e-3 & 0.838$\pm$4e-3 & 0.481$\pm$6e-3 & 0.612$\pm$7e-3 & 0.146$\pm$2e-2 & 0.636$\pm$2e-2\\
    ~&Ours& \bfseries0.737$\pm$2e-3 & \bfseries0.696$\pm$2e-3 & \bfseries0.921$\pm$4e-3 & \bfseries0.920$\pm$2e-3 & \bfseries0.570$\pm$5e-3 & 0.689$\pm$6e-3 & \bfseries0.336$\pm$3e-2 & \bfseries0.816$\pm$1e-2\\
    \midrule[0.5pt] \multirow{4}{*}{SHOP-COMPLEX}&MulMON&0.563$\pm$6e-3 & 0.594$\pm$2e-3 & 0.841$\pm$7e-3 & 0.850$\pm$3e-3 & \bfseries0.553$\pm$3e-3 & \bfseries0.677$\pm$3e-3 & 0.318$\pm$3e-2 & N/A\\ 
    ~&SIMONe& 0.196$\pm$3e-5 & 0.481$\pm$7e-5 & 0.785$\pm$1e-4 & 0.818$\pm$2e-4 & 0.481$\pm$1e-4 & 0.610$\pm$1e-4 & 0.004$\pm$5e-3 & N/A \\ 
    ~&OCLOC&0.230$\pm$3e-3 & 0.277$\pm$2e-3 & 0.301$\pm$2e-3 & 0.453$\pm$8e-4 & 0.179$\pm$1e-3 & 0.269$\pm$2e-3 & 0.172$\pm$1e-2 & 0.557$\pm$2e-2\\
    ~&Ours& \bfseries0.706$\pm$3e-3 & \bfseries0.666$\pm$3e-3 & \bfseries0.893$\pm$3e-3 & \bfseries0.893$\pm$3e-3 & 0.550$\pm$6e-3 & 0.670$\pm$6e-3 & \bfseries0.326$\pm$5e-2 & \bfseries0.808$\pm$6e-3\\
    \bottomrule[1.5pt]
    \end{tabular} 
    }
    \caption{The comparison results of multiple aspects on general sets (training on 4 views and testing on 4 views). All test values are evaluated 5 times, recorded with mean and standard deviation.}
    \label{tab:segment_4views_general}
\end{table*}
\begin{table*}
    \centering
    \scalebox{0.75}{
    \begin{tabular}{c|c|cccccccc} 
    \toprule[1.5pt]
    
    Dataset&Method&ARI-A$\uparrow$&AMI-A$\uparrow$&ARI-O$\uparrow$&AMI-O$\uparrow$&IoU$\uparrow$&F1$\uparrow$&OCA$\uparrow$&OOA$\uparrow$\\
    \midrule[0.5pt] \multirow{4}{*}{CLEVR-SIMPLE}&MulMON&0.632$\pm$1e-3 & 0.582$\pm$1e-3 & \bfseries 0.964$\pm$9e-4 & 0.949$\pm$7e-4 & \bfseries0.596$\pm$2e-3 &0.727$\pm$3e-3 & 0.564$\pm$2e-2 & N/A \\ 
    ~&SIMONe& 0.106$\pm$4e-5 & 0.310$\pm$3e-5 & 0.910$\pm$2e-4 & 0.887$\pm$2e-4 & 0.398$\pm$6e-5 & 0.555$\pm$6e-5 & 0.000$\pm$0e-0 & N/A\\ 
    ~&OCLOC&0.520$\pm$9e-4 & 0.492$\pm$1e-3 & 0.927$\pm$8e-3 & 0.917$\pm$4e-3 & 0.456$\pm$2e-3 & 0.615$\pm$3e-3 & 0.628$\pm$4e-2 & 0.936$\pm$1e-2\\
    ~&Ours& \bfseries0.772$\pm$2e-3 & \bfseries0.671$\pm$2e-3 & 0.959$\pm$3e-3  & \bfseries0.954$\pm$3e-3 & 0.595$\pm$5e-3  & \bfseries0.733$\pm$5e-3 & \bfseries0.594$\pm$5e-2 & \bfseries0.953$\pm$1e-2\\
    \midrule[0.5pt] \multirow{4}{*}{CLEVR-COMPLEX}&MulMON& 0.521$\pm$1e-2 & 0.509$\pm$6e-3 & 0.929$\pm$2e-3 & 0.908$\pm$2e-3 & \bfseries0.534$\pm$4e-3 & \bfseries0.672$\pm$4e-3 & \bfseries0.604$\pm$3e-2 & N/A\\
    ~&SIMONe& 0.092$\pm$1e-5 & 0.316$\pm$3e-5 & 0.914$\pm$3e-4 & 0.878$\pm$3e-4 & 0.423$\pm$2e-5 & 0.575$\pm$2e-5 & 0.000$\pm$0e-0 & N/A\\ 
    ~&OCLOC&0.366$\pm$1e-3 & 0.375$\pm$1e-3 & 0.827$\pm$8e-3 & 0.824$\pm$3e-3 & 0.351$\pm$2e-3 & 0.500$\pm$3e-3 & 0.168$\pm$5e-2 & 0.891$\pm$1e-2\\
    ~&Ours& \bfseries0.696$\pm$2e-3 & \bfseries0.592$\pm$2e-3 & \bfseries0.941$\pm$3e-3 &  \bfseries0.932$\pm$3e-3 & 0.509$\pm$4e-3 & 0.657$\pm$4e-3 & 0.550$\pm$8e-2 & \bfseries0.930$\pm$8e-3\\
    \midrule[0.5pt] \multirow{4}{*}{SHOP-SIMPLE}&MulMON& 0.435$\pm$1e-2 & 0.530$\pm$5e-3 & 0.883$\pm$6e-3 & 0.863$\pm$4e-3 & 0.587$\pm$4e-3 & 0.719$\pm$4e-3 & 0.160$\pm$4e-2 & N/A\\ 
    ~&SIMONe& 0.135$\pm$7e-5 & 0.321$\pm$1e-4 & 0.553$\pm$1e-4 & 0.581$\pm$2e-4 & 0.330$\pm$5e-5 & 0.462$\pm$6e-5 & 0.000$\pm$0e-0 & N/A\\ 
    ~&OCLOC& 0.663$\pm$3e-3 & 0.609$\pm$3e-3 & 0.913$\pm$4e-3 & 0.897$\pm$3e-3 & 0.619$\pm$6e-3 & 0.746$\pm$7e-3 & 0.388$\pm$2e-2 & 0.728$\pm$1e-2 \\
    ~&Ours& \bfseries0.803$\pm$7e-4 & \bfseries0.726$\pm$6e-4 & \bfseries0.958$\pm$1e-3 & \bfseries0.954$\pm$1e-3 & \bfseries0.656$\pm$6e-4 & \bfseries0.774$\pm$7e-4 & \bfseries0.528$\pm$5e-2 & \bfseries0.789$\pm$4e-3\\
    \midrule[0.5pt] \multirow{4}{*}{SHOP-COMPLEX}&MulMON& 0.585$\pm$1e-2 & 0.583$\pm$4e-3 & 0.871$\pm$2e-3 & 0.859$\pm$2e-3 & 0.625$\pm$5e-3 & 0.750$\pm$5e-3 & 0.330$\pm$5e-2 & N/A\\ 
    ~&SIMONe & 0.106$\pm$5e-5 & 0.234$\pm$2e-5 & 0.335$\pm$3e-4 & 0.388$\pm$2e-4 & 0.216$\pm$1e-4 & 0.329$\pm$2e-4 & 0.000$\pm$0e-0 & N/A\\ 
    ~&OCLOC&0.350$\pm$3e-3 & 0.273$\pm$2e-3 & 0.293$\pm$5e-3 & 0.408$\pm$6e-3 & 0.215$\pm$3e-3 & 0.321$\pm$4e-3 & 0.064$\pm$5e-3 & 0.579$\pm$1e-2\\
    ~&Ours& \bfseries0.786$\pm$4e-3 & \bfseries0.703$\pm$3e-3 & \bfseries0.949$\pm$4e-3 & \bfseries0.940$\pm$3e-3 & \bfseries0.662$\pm$6e-3  & \bfseries0.781$\pm$6e-3 & \bfseries0.434$\pm$3e-2 & \bfseries0.818$\pm$1e-2\\
    \bottomrule[1.5pt]
    \end{tabular} 
    }
    \caption{The comparison results of multiple aspects on test sets (training on 8 views and testing on 8 views). All test values are evaluated 5 times, recorded with mean and standard deviation.}
    \label{tab:segment_8views_test}
\end{table*}

\begin{table*}
    \centering
    \scalebox{0.75}{
    \begin{tabular}{c|c|cccccccc} 
    \toprule[1.5pt]
    
    Dataset&Method&ARI-A$\uparrow$&AMI-A$\uparrow$&ARI-O$\uparrow$&AMI-O$\uparrow$&IoU$\uparrow$&F1$\uparrow$&OCA$\uparrow$&OOA$\uparrow$\\
    \midrule[0.5pt] \multirow{4}{*}{CLEVR-SIMPLE}&MulMON& 0.554$\pm$1e-3 & 0.584$\pm$2e-3 & \bfseries0.924$\pm$3e-3 &\bfseries 0.920$\pm$3e-3 & \bfseries0.522$\pm$5e-3 & \bfseries0.656$\pm$6e-3 & \bfseries0.376$\pm$3e-2 & N/A\\ 
    ~&SIMONe & 0.132$\pm$4e-5 & 0.368$\pm$4e-5 & 0.787$\pm$7e-5 & 0.777$\pm$1e-4 & 0.326$\pm$3e-5 & 0.452$\pm$3e-5 & 0.020$\pm$1e-2 & N/A\\ 
    ~&OCLOC&0.393$\pm$2e-3 & 0.473$\pm$2e-3 & 0.844$\pm$6e-3 & 0.853$\pm$4e-3 & 0.386$\pm$4e-3 & 0.533$\pm$5e-3 & 0.254$\pm$3e-2 & 0.891$\pm$1e-2\\
    ~&Ours& \bfseries0.692$\pm$3e-3 & \bfseries0.635$\pm$4e-3 & 0.890$\pm$1e-2 & 0.897$\pm$6e-3 & 0.505$\pm$7e-3 & 0.637$\pm$8e-3 & 0.366$\pm$4e-2 & \bfseries0.913$\pm$1e-2\\
    \midrule[0.5pt] \multirow{4}{*}{CLEVR-COMPLEX}&MulMON& 0.458$\pm$4e-3 & 0.523$\pm$8e-4 & \bfseries0.893$\pm$2e-3 & \bfseries0.882$\pm$1e-3 & \bfseries0.459$\pm$1e-3 & \bfseries0.595$\pm$2e-3 & 0.322$\pm$7e-2 & N/A\\
    ~&SIMONe&0.109$\pm$1e-5 & 0.375$\pm$3e-5 & 0.814$\pm$6e-5  &0.795$\pm$9e-5 & 0.368$\pm$7e-5 & 0.500$\pm$1e-4 & 0.000$\pm$0e-0 & N/A\\ 
    ~&OCLOC& 0.173$\pm$2e-3 & 0.349$\pm$1e-3 & 0.730$\pm$5e-3 & 0.761$\pm$2e-3 & 0.267$\pm$2e-3 & 0.399$\pm$3e-3 & 0.084$\pm$3e-2 & 0.827$\pm$1e-2\\ 
    ~&Ours& \bfseries0.597$\pm$9e-4 & \bfseries0.560$\pm$6e-4 & 0.865$\pm$4e-3 & 0.873$\pm$3e-3 & 0.428$\pm$3e-3  & 0.569$\pm$5e-3 & \bfseries0.352$\pm$3e-2 & \bfseries0.874$\pm$4e-3 \\
    \midrule[0.5pt] \multirow{4}{*}{SHOP-SIMPLE}&MulMON&0.490$\pm$5e-3 & 0.575$\pm$1e-3 & 0.865$\pm$3e-3 & 0.859$\pm$1e-3 & 0.557$\pm$2e-3 & \bfseries0.688$\pm$3e-3 &\bfseries0.360$\pm$5e-2 & N/A\\ 
    ~&SIMONe & 0.105$\pm$5e-5 & 0.276$\pm$5e-5 & 0.418$\pm$1e-4 & 0.467$\pm$1e-4 & 0.193$\pm$5e-5 & 0.290$\pm$9e-5 & 0.002$\pm$4e-3 & N/A\\ 
    ~&OCLOC& 0.465$\pm$3e-3 & 0.513$\pm$3e-3 & 0.798$\pm$3e-3 & 0.812$\pm$3e-3 & 0.475$\pm$4e-3 & 0.605$\pm$4e-3 & 0.128$\pm$4e-2 & 0.614$\pm$5e-3 \\
    ~&Ours& \bfseries0.723$\pm$2e-3 & \bfseries0.683$\pm$2e-3 & \bfseries0.914$\pm$2e-3 & \bfseries0.915$\pm$1e-3 & \bfseries0.561$\pm$3e-3 & 0.685$\pm$3e-3 & 0.310$\pm$5e-2 & \bfseries0.827$\pm$1e-2\\
    \midrule[0.5pt] \multirow{4}{*}{SHOP-COMPLEX}&MulMON&0.561$\pm$7e-3 & 0.588$\pm$3e-3 & 0.839$\pm$3e-3 & 0.844$\pm$2e-3 & 0.560$\pm$4e-3 & 0.689$\pm$5e-3 & \bfseries0.374$\pm$5e-2 & N/A \\ 
    ~&SIMONe& 0.086$\pm$2e-5 & 0.203$\pm$4e-5 & 0.255$\pm$1e-4 & 0.323$\pm$9e-5 & 0.132$\pm$3e-5 & 0.213$\pm$6e-5 & 0.002$\pm$4e-3 & N/A\\ 
    ~&OCLOC&0.246$\pm$2e-3 & 0.243$\pm$2e-3 & 0.213$\pm$4e-3 & 0.371$\pm$3e-3 & 0.158$\pm$7e-4 & 0.246$\pm$1e-3 & 0.096$\pm$3e-2 & 0.570$\pm$2e-2 \\
    ~&Ours& \bfseries0.702$\pm$4e-3 & \bfseries0.660$\pm$3e-3 & \bfseries0.889$\pm$3e-3 & \bfseries0.889$\pm$3e-3 & \bfseries0.568$\pm$5e-3 & \bfseries0.690$\pm$6e-3 & 0.284$\pm$5e-2 & \bfseries0.823$\pm$1e-2\\
    \bottomrule[1.5pt]
    \end{tabular} 
    } 
    \caption{The comparison results of multiple aspects on general sets (training on 8 views and testing on 8 views). All test values are evaluated 5 times, recorded with mean and standard deviation.}
    \label{tab:segment_8views_general}
\end{table*}

\begin{table*}[ht]
    \renewcommand\arraystretch{1.5}
    \centering
    \scalebox{0.7}{
        \begin{tabular}{c|c|c| cccccccc} 
        \toprule[1.5pt]
        
        Dataset&Query&Method&ARI-A$\uparrow$&AMI-A$\uparrow$&ARI-O$\uparrow$&AMI-O$\uparrow$&IoU$\uparrow$&F1$\uparrow$&MSE$\downarrow$&OOA$\uparrow$ \\
        \midrule[0.5pt] \multirow{6}{*}{CLEVR-SIMPLE}&\multirow{2}{*}{1}& MulMON & 0.667$\pm$1e-3 & 0.606$\pm$1e-3 & \bfseries0.955$\pm$2e-3 & 0.948$\pm$2e-3 &\bfseries 0.619$\pm$2e-3 & \bfseries 0.741$\pm$3e-3&\bfseries 0.0017$\pm$3e-5 & N/A \\ 
        ~ & ~ & Ours &\bfseries0.789$\pm$8e-3 &\bfseries0.699$\pm$8e-3  & 0.948$\pm$8e-3  & \bfseries0.949$\pm$6e-3 & 0.593$\pm$5e-3  &0.720$\pm$5e-3 & 0.0020$\pm$1e-4 & 0.982$\pm$1e-2\\
        \Xcline{2-11}{0.75pt}
        ~ &\multirow{2}{*}{2}& MulMON & 0.632$\pm$1e-3 & 0.581$\pm$1e-3 & \bfseries0.959$\pm$2e-3 & \bfseries0.947$\pm$1e-3 &\bfseries0.592$\pm$2e-3  & \bfseries0.719$\pm$3e-3 &\bfseries0.0016$\pm$3e-5 &  N/A\\ 
        ~ &~& Ours &\bfseries 0.760$\pm$4e-3 & \bfseries0.662$\pm$5e-3 & 0.946$\pm$5e-3 & 0.945$\pm$4e-3 &0.572$\pm$3e-3  & 0.711$\pm$4e-3 &0.0020$\pm$6e-5 &  0.973$\pm$9e-3\\ 
        \Xcline{2-11}{0.75pt}
        ~&\multirow{2}{*}{4} &MulMON &0.618$\pm$9e-4 & 0.569$\pm$9e-4 & \bfseries0.955$\pm$1e-3 & \bfseries0.940$\pm$1e-3 & \bfseries0.580$\pm$2e-3 & \bfseries0.710$\pm$3e-3 & \bfseries0.0016$\pm$3e-5 &N/A \\
        ~&~& Ours & \bfseries0.753$\pm$4e-3& \bfseries0.648$\pm$4e-3 & 0.942$\pm$6e-3 & 0.937$\pm$4e-3 & 0.563$\pm$4e-3 &0.706$\pm$5e-3 & 0.0020$\pm$9e-5 & 0.964$\pm$8e-3\\
        \bottomrule[1.5pt]
        \multirow{6}{*}{CLEVR-COMPLEX}&\multirow{2}{*}{1}& MulMON & 0.564$\pm$1e-2 & 0.539$\pm$6e-3 & 0.941$\pm$4e-3 & 0.935$\pm$2e-3 & \bfseries0.549$\pm$5e-3 & \bfseries0.678$\pm$6e-3 &\bfseries0.0019$\pm$5e-5 & N/A \\ 
        ~ & ~ & Ours &\bfseries0.764$\pm$1e-2 &\bfseries0.668$\pm$1e-2  & \bfseries0.961$\pm$5e-3  & \bfseries0.959$\pm$4e-3 & 0.528$\pm$6e-3  &0.666$\pm$6e-3 & \bfseries0.0019$\pm$2e-4 & 0.973$\pm$2e-2\\
        \Xcline{2-11}{0.75pt}
        ~ &\multirow{2}{*}{2}& MulMON & 0.522$\pm$1e-2 & 0.506$\pm$5e-3 & 0.926$\pm$3e-3 & 0.910$\pm$2e-3 &\bfseries0.526$\pm$4e-3  & \bfseries0.661$\pm$5e-3 &\bfseries0.0021$\pm$3e-5 &  N/A\\ 
        ~ &~& Ours & \bfseries0.725$\pm$7e-3 & \bfseries0.625$\pm$7e-3 &\bfseries 0.956$\pm$5e-3 & \bfseries0.952$\pm$5e-3 &0.513$\pm$4e-3  & 0.660$\pm$3e-3 &\bfseries0.0021$\pm$1e-4 &  0.941$\pm$2e-2\\ 
        \Xcline{2-11}{0.75pt}
        ~&\multirow{2}{*}{4} &MulMON &0.519$\pm$1e-2 & 0.500$\pm$5e-3 & 0.917$\pm$4e-3 & 0.896$\pm$3e-3 & \bfseries0.521$\pm$5e-3 & \bfseries0.658$\pm$5e-3 & \bfseries0.0022$\pm$3e-5 &N/A \\
        ~&~& Ours &\bfseries 0.721$\pm$1e-3& \bfseries0.614$\pm$1e-3 & \bfseries0.949$\pm$4e-3 & \bfseries0.940$\pm$4e-3 & 0.511$\pm$2e-3 &0.660$\pm$3e-3  & 0.0023$\pm$5e-5 & 0.942$\pm$1e-2\\
        \bottomrule[1.5pt]
        \multirow{6}{*}{SHOP-SIMPLE}&\multirow{2}{*}{1}& MulMON & 0.457$\pm$1e-2 & 0.528$\pm$5e-3 & 0.867$\pm$3e-3 & 0.862$\pm$3e-3 & 0.555$\pm$4e-3 & 0.675$\pm$4e-3 &0.0050$\pm$6e-5 & N/A \\ 
        ~ & ~ & Ours &\bfseries0.785$\pm$1e-2 &\bfseries0.715$\pm$9e-3  &\bfseries 0.950$\pm$8e-3  & \bfseries0.953$\pm$7e-3 & \bfseries0.627$\pm$8e-3  &\bfseries0.739$\pm$9e-3  & \bfseries0.0035$\pm$4e-4 & 0.767$\pm$2e-2\\
        \Xcline{2-11}{0.75pt}
        ~ &\multirow{2}{*}{2}& MulMON & 0.425$\pm$1e-2 & 0.515$\pm$4e-3 & 0.868$\pm$3e-3 & 0.850$\pm$1e-3 &0.545$\pm$4e-3  & 0.668$\pm$4e-3 &0.0053$\pm$8e-5 &  N/A\\ 
        ~ &~& Ours & \bfseries0.768$\pm$9e-3 &\bfseries 0.693$\pm$9e-3 & \bfseries0.948$\pm$8e-3 & \bfseries0.948$\pm$7e-3 &\bfseries0.620$\pm$8e-3  & \bfseries0.743$\pm$9e-3 &\bfseries0.0037$\pm$3e-4 &  0.741$\pm$4e-2\\ 
        \Xcline{2-11}{0.75pt}
        ~&\multirow{2}{*}{4} &MulMON &0.410$\pm$1e-2 & 0.507$\pm$5e-3 & 0.864$\pm$3e-3 & 0.837$\pm$1e-3 & 0.544$\pm$4e-3 & 0.667$\pm$4e-3 & 0.0055$\pm$8e-5 &N/A \\
        ~&~& Ours & \bfseries0.758$\pm$7e-3& \bfseries0.679$\pm$7e-3 &\bfseries 0.943$\pm$5e-3 & \bfseries0.940$\pm$5e-3 &\bfseries 0.614$\pm$7e-3 &\bfseries0.739$\pm$8e-3  & \bfseries0.0040$\pm$3e-4 & 0.715$\pm$2e-2\\
        \bottomrule[1.5pt]
        \multirow{6}{*}{SHOP-COMPLEX}&\multirow{2}{*}{1}& MulMON & 0.612$\pm$1e-2 & 0.577$\pm$5e-3 & 0.835$\pm$4e-3 & 0.838$\pm$3e-3 & 0.571$\pm$2e-3 & 0.674$\pm$3e-3 &0.0043$\pm$9e-5 & N/A \\ 
        ~ & ~ & Ours &\bfseries0.732$\pm$1e-2 &\bfseries0.663$\pm$9e-3  &\bfseries 0.918$\pm$5e-3  & \bfseries0.923$\pm$3e-3 & \bfseries0.580$\pm$8e-3  &\bfseries0.698$\pm$9e-3  & \bfseries0.0038$\pm$3e-4 & 0.835$\pm$2e-2\\
        \Xcline{2-11}{0.75pt}
        ~ &\multirow{2}{*}{2}& MulMON & 0.607$\pm$1e-2 & 0.569$\pm$5e-3 & 0.836$\pm$4e-3 & 0.831$\pm$3e-3 &0.558$\pm$4e-3  & 0.666$\pm$4e-3 &0.0046$\pm$8e-5 &  N/A\\ 
        ~ &~& Ours & \bfseries0.734$\pm$1e-2 & \bfseries0.656$\pm$1e-2 &\bfseries 0.920$\pm$1e-2 & \bfseries0.921$\pm$6e-3 &\bfseries0.583$\pm$9e-3  &\bfseries 0.711$\pm$8e-3 &\bfseries0.0038$\pm$2e-4 &  0.796$\pm$2e-2\\ 
        \Xcline{2-11}{0.75pt}
        ~&\multirow{2}{*}{4} &MulMON &0.601$\pm$1e-2 & 0.561$\pm$5e-3 & 0.824$\pm$5e-3 & 0.819$\pm$4e-3 & 0.554$\pm$4e-3 & 0.663$\pm$5e-3 & 0.0049$\pm$9e-5 &N/A \\
        ~&~& Ours & \bfseries0.732$\pm$7e-3& \bfseries0.646$\pm$7e-3 & \bfseries0.911$\pm$6e-3 & \bfseries0.907$\pm$5e-3 & \bfseries0.579$\pm$6e-3 &\bfseries0.709$\pm$6e-3  & \bfseries0.0040$\pm$2e-4 & 0.749$\pm$2e-2\\
        \bottomrule[1.5pt]
        \end{tabular} 
    }
    \caption{The comparison results of prediction on test sets (the test mode is 1, the observed views are 6, and query views are 1, 2, 4). All test values are evaluated 5 times, recorded with mean and standard deviation.}\label{tab:predict_m1_o6}
\end{table*}

\begin{table*}[ht]
    \renewcommand\arraystretch{1.5}
    \centering
    \scalebox{0.7}{
        \begin{tabular}{c|c|c| cccccccc} 
        \toprule[1.5pt]
        
        Dataset&Query&Method&ARI-A$\uparrow$&AMI-A$\uparrow$&ARI-O$\uparrow$&AMI-O$\uparrow$&IoU$\uparrow$&F1$\uparrow$&MSE$\downarrow$&OOA$\uparrow$ \\
        \midrule[0.5pt] \multirow{6}{*}{CLEVR-SIMPLE}&\multirow{2}{*}{1}& MulMON & 0.604$\pm$6e-4 & 0.564$\pm$3e-4 &\bfseries0.962$\pm$2e-3  &\bfseries0.957$\pm$1e-3  &\bfseries0.579$\pm$3e-3  & \bfseries0.710$\pm$4e-3 &\bfseries0.0013$\pm$2e-5  & N/A \\ 
        ~ & ~ & Ours &\bfseries 0.725$\pm$8e-3 &\bfseries 0.631$\pm$6e-3 & 0.934$\pm$5e-3 & 0.938$\pm$4e-3 &0.544$\pm$5e-3  &0.679$\pm$5e-3  &0.0022$\pm$1e-4  &0.958$\pm$1e-2 \\
        \Xcline{2-11}{0.75pt}
        ~ &\multirow{2}{*}{2}& MulMON &0.621$\pm$7e-4  &0.576$\pm$8e-4  & \bfseries0.960$\pm$3e-3 &\bfseries 0.951$\pm$2e-3 & \bfseries0.588$\pm$3e-3 & \bfseries0.718$\pm$4e-3 &\bfseries0.0015$\pm$2e-5 &N/A  \\ 
        ~ &~& Ours & \bfseries0.733$\pm$8e-3 & \bfseries0.630$\pm$8e-3 & 0.924$\pm$8e-3 & 0.925$\pm$7e-3 & 0.554$\pm$5e-3 & 0.692$\pm$6e-3 & 0.0024$\pm$1e-4 & 0.958$\pm$1e-2 \\ 
        \Xcline{2-11}{0.75pt}
        ~&\multirow{2}{*}{4} &MulMON & 0.667$\pm$9e-4 &0.609$\pm$1e-3  & \bfseries0.965$\pm$3e-3 & \bfseries0.953$\pm$2e-3 &\bfseries 0.617$\pm$4e-3 & \bfseries0.742$\pm$4e-3 & \bfseries0.0014$\pm$2e-5 & N/A\\
        ~&~& Ours &\bfseries 0.774$\pm$9e-3 &\bfseries 0.662$\pm$9e-3 & 0.923$\pm$6e-3 & 0.918$\pm$6e-3 & 0.586$\pm$9e-3  & 0.723$\pm$8e-3 & 0.0022$\pm$2e-4 &0.963$\pm$5e-3 \\
        \bottomrule[1.5pt]
        \multirow{6}{*}{CLEVR-COMPLEX}&\multirow{2}{*}{1}& MulMON & 0.503$\pm$6e-3 & 0.497$\pm$2e-3 & 0.931$\pm$6e-3 & 0.929$\pm$3e-3 &\bfseries0.524$\pm$8e-4  & \bfseries0.664$\pm$1e-3 & \bfseries0.0019$\pm$1e-5 & N/A \\ 
        ~ & ~ & Ours & \bfseries0.694$\pm$9e-3 & \bfseries0.601$\pm$8e-3 &\bfseries 0.949$\pm$7e-3 &\bfseries0.951$\pm$5e-3  &0.495$\pm$8e-3  & 0.640$\pm$8e-3 & 0.0025$\pm$1e-4  & 0.932$\pm$4e-2\\
        \Xcline{2-11}{0.75pt}
        ~ &\multirow{2}{*}{2}& MulMON & 0.520$\pm$7e-3 &0.509$\pm$2e-3  & \bfseries0.930$\pm$7e-3 & \bfseries0.923$\pm$4e-3 & \bfseries0.531$\pm$1e-3 & \bfseries0.670$\pm$2e-3 & \bfseries0.0020$\pm$2e-5 & N/A  \\ 
        ~ &~& Ours & \bfseries0.681$\pm$2e-2 & \bfseries0.573$\pm$2e-2 & 0.886$\pm$3e-2 & 0.890$\pm$2e-2 & 0.473$\pm$2e-2 & 0.618$\pm$2e-2 & 0.0032$\pm$3e-4 & 0.879$\pm$3e-2 \\ 
        \Xcline{2-11}{0.75pt}
        ~&\multirow{2}{*}{4} &MulMON & 0.562$\pm$7e-3 &0.538$\pm$2e-3  &\bfseries 0.934$\pm$7e-3  &\bfseries 0.920$\pm$3e-3 & \bfseries0.558$\pm$9e-4 & \bfseries 0.694$\pm$1e-3 &\bfseries 0.0020$\pm$1e-5 & N/A \\
        ~&~& Ours &\bfseries 0.716$\pm$2e-2 & \bfseries0.600$\pm$2e-2 & 0.901$\pm$2e-2 & 0.897$\pm$2e-2 & 0.502$\pm$1e-2 &  0.648$\pm$1e-2 & 0.0031$\pm$2e-4 & 0.910$\pm$3e-2\\
        \bottomrule[1.5pt]
        \multirow{6}{*}{SHOP-SIMPLE}&\multirow{2}{*}{1}& MulMON & 0.435$\pm$1e-2 &0.519$\pm$5e-3  &0.864$\pm$9e-3  &0.851$\pm$5e-3  & 0.563$\pm$2e-3 & 0.692$\pm$3e-3 & 0.0047$\pm$3e-5 & N/A  \\ 
        ~ & ~ & Ours & \bfseries0.749$\pm$7e-3 & \bfseries0.676$\pm$6e-3 & \bfseries0.953$\pm$7e-3 & \bfseries0.954$\pm$5e-3 & \bfseries0.589$\pm$8e-3 & \bfseries0.710$\pm$9e-3 & \bfseries0.0038$\pm$2e-4 &0.846$\pm$5e-2 \\
        \Xcline{2-11}{0.75pt}
        ~ &\multirow{2}{*}{2}& MulMON & 0.430$\pm$1e-2 & 0.521$\pm$5e-3 & 0.867$\pm$7e-3 & 0.845$\pm$3e-3 & 0.564$\pm$2e-3 & 0.691$\pm$3e-3 &0.0048$\pm$4e-5 & N/A \\ 
        ~ &~& Ours &\bfseries0.746$\pm$5e-3  &\bfseries0.668$\pm$6e-3  & \bfseries0.939$\pm$1e-2 &\bfseries0.938$\pm$8e-3  & \bfseries0.584$\pm$7e-3 & \bfseries0.706$\pm$8e-3 & \bfseries0.0042$\pm$3e-4 & 0.814$\pm$2e-2 \\ 
        \Xcline{2-11}{0.75pt}
        ~&\multirow{2}{*}{4} &MulMON & 0.455$\pm$1e-2 & 0.545$\pm$5e-3 & 0.866$\pm$9e-3 & 0.846$\pm$3e-3 & 0.597$\pm$2e-3 & 0.722$\pm$3e-3 & 0.0049$\pm$7e-5 & N/A \\
        ~&~& Ours & \bfseries0.784$\pm$4e-3 & \bfseries0.701$\pm$5e-3 & \bfseries0.944$\pm$5e-3 & \bfseries0.939$\pm$5e-3 & \bfseries0.633$\pm$7e-3 &  \bfseries0.755$\pm$7e-3& \bfseries0.0041$\pm$2e-4 & \bfseries0.819$\pm$3e-2 \\
        \bottomrule[1.5pt]
        \multirow{6}{*}{SHOP-COMPLEX}&\multirow{2}{*}{1}& MulMON & 0.666$\pm$5e-3 & 0.619$\pm$2e-3 & 0.864$\pm$6e-3 &0.859$\pm$5e-3  & \bfseries0.631$\pm$2e-3 & \bfseries0.746$\pm$2e-3 & \bfseries0.0036$\pm$3e-5 & N/A \\ 
        ~ & ~ & Ours & \bfseries0.733$\pm$1e-2  & \bfseries0.658$\pm$1e-2 & \bfseries0.931$\pm$7e-3 & \bfseries0.933$\pm$5e-3 & 0.573$\pm$8e-3 & 0.696$\pm$8e-3 & 0.0037$\pm$3e-4 & 0.827$\pm$4e-2\\
        \Xcline{2-11}{0.75pt}
        ~ &\multirow{2}{*}{2}& MulMON & 0.669$\pm$5e-3 & 0.620$\pm$2e-3 & 0.870$\pm$6e-3 & 0.857$\pm$3e-3 & \bfseries0.631$\pm$2e-3 & \bfseries0.745$\pm$3e-3 & \bfseries0.0036$\pm$2e-5& N/A  \\ 
        ~ &~& Ours & \bfseries0.731$\pm$1e-2 &\bfseries 0.648$\pm$1e-2 & \bfseries0.911$\pm$1e-2 & \bfseries0.911$\pm$7e-3 & 0.561$\pm$9e-3 & 0.683$\pm$9e-3  & 0.0040$\pm$2e-4 &  0.773$\pm$2e-2\\ 
        \Xcline{2-11}{0.75pt}
        ~&\multirow{2}{*}{4} &MulMON &0.702$\pm$5e-3  & 0.645$\pm$2e-3 &0.866$\pm$7e-3  & 0.856$\pm$4e-3 & \bfseries0.663$\pm$2e-3 &\bfseries 0.775$\pm$3e-3 & \bfseries0.0037$\pm$1e-5 & N/A \\
        ~&~& Ours & \bfseries0.756$\pm$1e-2 & \bfseries0.668$\pm$1e-2 & \bfseries0.910$\pm$1e-2 & \bfseries0.908$\pm$9e-3 & 0.604$\pm$1e-2 & 0.729$\pm$9e-3 & 0.0041$\pm$2e-4 & 0.791$\pm$1e-2\\
        \bottomrule[1.5pt]
        \end{tabular} 
    }
    \caption{The comparison results of prediction on test sets (the test mode is 2, the observed views are 6, and query views are 1, 2, 4). All test values are evaluated 5 times, recorded with mean and standard deviation.}\label{tab:predict_m2_o6}
\end{table*}

\begin{table*}[ht]
    \renewcommand\arraystretch{1.5}
    \centering
    \scalebox{0.7}{
        \begin{tabular}{c|c|c| cccccccc} 
        \toprule[1.5pt]
        Dataset&Query&Method&ARI-A$\uparrow$&AMI-A$\uparrow$&ARI-O$\uparrow$&AMI-O$\uparrow$&IoU$\uparrow$&F1$\uparrow$&MSE$\downarrow$&OOA$\uparrow$ \\
        \midrule[0.5pt]
        \multirow{4}{*}{CLEVR-SIMPLE}&\multirow{2}{*}{1}& MulMON & 0.668$\pm$2e-3 & 0.607$\pm$2e-3 & 0.955$\pm$3e-3 & 0.947$\pm$2e-3 & \bfseries0.619$\pm$3e-3 & \bfseries0.741$\pm$4e-3 &\bfseries0.0017$\pm$3e-5  & N/A \\ 
        ~ & ~ & Ours & \bfseries0.805$\pm$1e-2 & \bfseries0.715$\pm$1e-2 & \bfseries0.964$\pm$9e-3 & \bfseries0.961$\pm$7e-3 & 0.606$\pm$1e-2 & 0.734$\pm$2e-2 & 0.0018$\pm$3e-4 & 0.977$\pm$2e-2\\
        \Xcline{2-11}{0.75pt}
        ~&\multirow{2}{*}{2} &MulMON & 0.629$\pm$2e-3 & 0.575$\pm$1e-3 &0.950$\pm$3e-3  & 0.939$\pm$2e-3 &\bfseries0.586$\pm$4e-3  & 0.715$\pm$5e-3 &\bfseries0.0016$\pm$4e-5  &N/A \\
        ~&~& Ours & \bfseries0.768$\pm$1e-2 & \bfseries0.669$\pm$1e-2 & \bfseries0.958$\pm$1e-2 & \bfseries0.954$\pm$1e-2 & 0.580$\pm$1e-2 & \bfseries0.719$\pm$1e-2 & 0.0018$\pm$2e-4 & 0.979$\pm$1e-2\\
        \bottomrule[1.5pt]
        \multirow{4}{*}{CLEVR-COMPLEX}&\multirow{2}{*}{1}& MulMON & 0.567$\pm$7e-3 & 0.542$\pm$3e-3 &0.944$\pm$2e-3  & 0.937$\pm$2e-3 & \bfseries0.554$\pm$4e-3 & \bfseries0.684$\pm$4e-3 & \bfseries0.0018$\pm$2e-5 & N/A  \\ 
        ~ & ~ & Ours & \bfseries0.765$\pm$9e-3 & \bfseries0.669$\pm$8e-3 & \bfseries0.959$\pm$1e-2 & \bfseries0.959$\pm$9e-3 & 0.524$\pm$1e-2 & 0.661$\pm$1e-2 & 0.0019$\pm$2e-4 & 0.950$\pm$9e-3\\
        \Xcline{2-11}{0.75pt}
        ~&\multirow{2}{*}{2} &MulMON & 0.526$\pm$8e-3 & 0.506$\pm$3e-3 & 0.925$\pm$2e-3 & 0.907$\pm$1e-3 & \bfseries0.528$\pm$3e-3 & \bfseries0.664$\pm$4e-3 & \bfseries0.0021$\pm$2e-5 & N/A\\
        ~&~& Ours &\bfseries 0.720$\pm$6e-3 & \bfseries0.619$\pm$6e-3 & \bfseries0.948$\pm$1e-2 & \bfseries0.947$\pm$8e-3 &0.505$\pm$7e-3  &  0.651$\pm$8e-3& 0.0023$\pm$8e-5 & 0.915$\pm$2e-2\\
        \bottomrule[1.5pt]
        \multirow{4}{*}{SHOP-SIMPLE}&\multirow{2}{*}{1}& MulMON &0.446$\pm$8e-3  & 0.525$\pm$4e-3  & 0.872$\pm$5e-3 &0.867$\pm$3e-3  & 0.557$\pm$6e-3 & 0.678$\pm$6e-3 & 0.0049$\pm$5e-5 & N/A \\ 
        ~ & ~ & Ours & \bfseries0.796$\pm$4e-3 & \bfseries0.725$\pm$4e-3 & \bfseries0.959$\pm$4e-3 & \bfseries0.961$\pm$3e-3 &\bfseries 0.635$\pm$6e-3 & \bfseries0.747$\pm$6e-3 & \bfseries0.0031$\pm$8e-5 & 0.788$\pm$2e-2\\
        \Xcline{2-11}{0.75pt}
        ~&\multirow{2}{*}{2} &MulMON & 0.406$\pm$9e-3 &0.508$\pm$5e-3  &0.868$\pm$7e-3  & 0.851$\pm$4e-3 & 0.543$\pm$5e-3 & 0.668$\pm$5e-3 & 0.0056$\pm$6e-5 & N/A \\
        ~&~& Ours & \bfseries0.776$\pm$4e-3 & \bfseries0.700$\pm$3e-3 & \bfseries0.958$\pm$4e-3 & \bfseries0.957$\pm$2e-3 & \bfseries0.624$\pm$6e-3 & \bfseries0.747$\pm$5e-3 & \bfseries0.0038$\pm$1e-4 & 0.820$\pm$1e-2\\
        \bottomrule[1.5pt]
        \multirow{4}{*}{SHOP-COMPLEX}&\multirow{2}{*}{1}& MulMON & 0.619$\pm$9e-3 &0.587$\pm$4e-3  & 0.849$\pm$4e-3 &0.851$\pm$1e-3  & 0.590$\pm$3e-3 & 0.696$\pm$3e-3 & 0.0039$\pm$4e-5 &  N/A\\ 
        ~ & ~ & Ours & \bfseries0.750$\pm$1e-2 &\bfseries 0.680$\pm$1e-2 & \bfseries0.923$\pm$6e-3  & \bfseries0.930$\pm$6e-3 & \bfseries0.594$\pm$1e-2 & \bfseries0.711$\pm$1e-2 & \bfseries0.0033$\pm$2e-4 & 0.852$\pm$1e-2\\
        \Xcline{2-11}{0.75pt}
        ~&\multirow{2}{*}{2} &MulMON & 0.613$\pm$8e-3 & 0.578$\pm$3e-3 & 0.843$\pm$4e-3 & 0.841$\pm$1e-3 & 0.580$\pm$4e-3 & 0.691$\pm$4e-3 &0.0044$\pm$7e-5  & N/A\\
        ~&~& Ours &\bfseries0.745$\pm$7e-3  & \bfseries0.668$\pm$6e-3 &\bfseries0.923$\pm$8e-3  & \bfseries0.926$\pm$7e-3 & \bfseries0.593$\pm$1e-2 &  \bfseries0.719$\pm$1e-2 & \bfseries0.0036$\pm$1e-4 & 0.844$\pm$1e-2 \\
        \bottomrule[1.5pt]
        \end{tabular}
    }
    \caption{The comparison results of prediction on test sets (the test mode is 1, the observed views are 7, and query views are 1, 2). All test values are evaluated 5 times, recorded with mean and standard deviation.}\label{tab:predict_m1_o7}
\end{table*}
\begin{table*}[ht]
    \renewcommand\arraystretch{1.5}
    \centering
    \scalebox{0.7}{
        \begin{tabular}{c|c|c| cccccccc} 
        \toprule[1.5pt]
        Dataset&Query&Method&ARI-A$\uparrow$&AMI-A$\uparrow$&ARI-O$\uparrow$&AMI-O$\uparrow$&IoU$\uparrow$&F1$\uparrow$&MSE$\downarrow$&OOA$\uparrow$ \\
        \midrule[0.5pt]
        \multirow{4}{*}{CLEVR-SIMPLE}&\multirow{2}{*}{1}& MulMON & 0.640$\pm$3e-3 & 0.594$\pm$1e-3 &\bfseries 0.960$\pm$3e-3 & 0.954$\pm$2e-3 & \bfseries0.606$\pm$3e-3 &\bfseries0.733$\pm$3e-3 & \bfseries0.0016$\pm$3e-5 & N/A \\ 
        ~ & ~ & Ours & \bfseries0.774$\pm$8e-3 & \bfseries0.683$\pm$7e-3 & 0.958$\pm$5e-3 & \bfseries0.959$\pm$5e-3 & 0.600$\pm$6e-3 & 0.731$\pm$7e-3 & 0.0021$\pm$1e-4 & 0.960$\pm$1e-2\\
        \Xcline{2-11}{0.75pt}
        ~&\multirow{2}{*}{2} &MulMON & 0.661$\pm$2e-3 & 0.609$\pm$2e-3 & \bfseries0.964$\pm$3e-3 & \bfseries0.955$\pm$2e-3 & \bfseries0.614$\pm$3e-3 & 0.739$\pm$3e-3 & \bfseries0.0015$\pm$2e-5 & N/A\\
        ~&~& Ours & \bfseries0.793$\pm$6e-3 & 0.\bfseries694$\pm$6e-3 & 0.954$\pm$6e-3 & 0.953$\pm$5e-3 & 0.613$\pm$5e-3 &  \bfseries0.744$\pm$5e-3 &0.0019$\pm$1e-4  & 0.972$\pm$8e-3\\
        \bottomrule[1.5pt]
        \multirow{4}{*}{CLEVR-COMPLEX}&\multirow{2}{*}{1}& MulMON & 0.543$\pm$2e-3 & 0.528$\pm$1e-3 & 0.931$\pm$3e-3 & 0.929$\pm$2e-3 & \bfseries0.542$\pm$3e-3 & \bfseries0.678$\pm$4e-3 & \bfseries0.0021$\pm$2e-5 & N/A \\ 
        ~ & ~ & Ours & \bfseries0.722$\pm$2e-2 & \bfseries0.628$\pm$2e-2 &  \bfseries0.934$\pm$1e-2  & \bfseries0.935$\pm$1e-2 & 0.511$\pm$2e-2 &0.651$\pm$2e-2 & 0.0030$\pm$4e-4  & 0.956$\pm$2e-2\\
        \Xcline{2-11}{0.75pt}
        ~&\multirow{2}{*}{2} &MulMON & 0.563$\pm$3e-3 & 0.543$\pm$2e-3 & 0.935$\pm$4e-3 & 0.927$\pm$2e-3 & \bfseries0.559$\pm$3e-3 &  \bfseries0.694$\pm$4e-3 & \bfseries0.0021$\pm$2e-5 & N/A\\
        ~&~& Ours & \bfseries0.743$\pm$2e-2 & \bfseries0.639$\pm$1e-2 &\bfseries0.936$\pm$2e-2  & \bfseries0.934$\pm$1e-2 & 0.527$\pm$1e-2 & 0.669$\pm$2e-2 & 0.0028$\pm$2e-4 & 0.938$\pm$2e-2\\
        \bottomrule[1.5pt]
        \multirow{4}{*}{SHOP-SIMPLE}&\multirow{2}{*}{1}& MulMON & 0.445$\pm$1e-2 & 0.536$\pm$7e-3 & 0.886$\pm$5e-3 &  0.870$\pm$3e-3 & 0.580$\pm$4e-3 & 0.705$\pm$3e-3  &  0.0048$\pm$5e-5 & N/A \\ 
        ~ & ~ & Ours & \bfseries0.755$\pm$1e-2 & \bfseries0.686$\pm$8e-3 & \bfseries0.954$\pm$6e-3 & \bfseries0.954$\pm$4e-3 &  \bfseries0.608$\pm$6e-3 & \bfseries0.724$\pm$5e-3 &\bfseries0.0043$\pm$6e-4  &0.766$\pm$1e-2 \\
        \Xcline{2-11}{0.75pt}
        ~&\multirow{2}{*}{2} &MulMON & 0.448$\pm$1e-2 & 0.545$\pm$6e-3 & 0.881$\pm$4e-3 & 0.864$\pm$2e-3 & 0.591$\pm$3e-3 & 0.716$\pm$2e-3 & 0.0048$\pm$6e-5 & N/A \\
        ~&~& Ours &\bfseries0.775$\pm$8e-3 & \bfseries0.700$\pm$8e-3 & \bfseries0.951$\pm$8e-3 & \bfseries0.948$\pm$7e-3 &  \bfseries0.629$\pm$3e-3 & \bfseries0.749$\pm$3e-3 &\bfseries0.0041$\pm$3e-4  &0.816$\pm$3e-2 \\
        \bottomrule[1.5pt]
        \multirow{4}{*}{SHOP-COMPLEX}&\multirow{2}{*}{1}& MulMON & 0.668$\pm$5e-3 & 0.626$\pm$2e-3 & 0.880$\pm$5e-3 & 0.871$\pm$4e-3 & \bfseries0.639$\pm$4e-3 & \bfseries0.750$\pm$4e-3  &\bfseries0.0036$\pm$3e-5  &  N/A\\ 
        ~ & ~ & Ours & \bfseries0.745$\pm$1e-2 & \bfseries0.672$\pm$1e-2 & \bfseries0.925$\pm$8e-3 & \bfseries0.928$\pm$6e-3 &  0.584$\pm$8e-3 & 0.704$\pm$9e-3 & 0.0038$\pm$2e-4 & 0.730$\pm$3e-2\\
        \Xcline{2-11}{0.75pt}
        ~&\multirow{2}{*}{2} &MulMON & 0.688$\pm$6e-3 & 0.641$\pm$3e-3 & 0.878$\pm$6e-3 & 0.871$\pm$4e-3 & \bfseries0.653$\pm$4e-3 & \bfseries0.765$\pm$5e-3 & \bfseries0.0037$\pm$5e-5 & N/A\\
        ~&~& Ours & \bfseries0.765$\pm$1e-2 & \bfseries0.683$\pm$1e-2 & \bfseries0.926$\pm$1e-2 & \bfseries0.926$\pm$1e-2 & 0.607$\pm$9e-3 & 0.731$\pm$8e-3 & 0.0038$\pm$3e-4 & 0.764$\pm$2e-2\\
        \bottomrule[1.5pt]
        \end{tabular}
    }
    \caption{The comparison results of prediction on test sets (the test mode is 2, the observed views are 7, and query views are 1, 2). All test values are evaluated 5 times, recorded with mean and standard deviation.}\label{tab:predict_m2_o7}
\end{table*}
    
\begin{table*}[ht]
    \renewcommand\arraystretch{1.5}
    \centering
    \scalebox{0.7}{
        \begin{tabular}{c|c|c| cccccccc} 
        \toprule[1.5pt]
        Dataset&Query&Method&ARI-A$\uparrow$&AMI-A$\uparrow$&ARI-O$\uparrow$&AMI-O$\uparrow$&IoU$\uparrow$&F1$\uparrow$&MSE$\downarrow$&OOA$\uparrow$ \\
        \midrule[0.5pt]
        \multirow{4}{*}{CLEVR-SIMPLE}&\multirow{2}{*}{1}& MulMON & 0.624$\pm$5e-4 & 0.581$\pm$6e-4 & \bfseries0.969$\pm$1e-3 & \bfseries0.961$\pm$1e-3 & \bfseries0.594$\pm$2e-3 &\bfseries0.722$\pm$3e-3 & \bfseries0.0014$\pm$2e-5 & N/A \\ 
        ~ & ~ & Ours & \bfseries0.756$\pm$1e-2 & \bfseries0.669$\pm$9e-3 & 0.961$\pm$7e-3 & \bfseries0.961$\pm$5e-3 &  0.570$\pm$8e-3& 0.705$\pm$8e-3 & 0.0018$\pm$2e-4 & 0.957$\pm$2e-2\\
        \Xcline{2-11}{0.75pt}
        ~&\multirow{2}{*}{2} &MulMON & 0.600$\pm$3e-4 & 0.559$\pm$6e-4 & \bfseries0.962$\pm$1e-3  & 0.950$\pm$2e-3 & \bfseries0.577$\pm$2e-3 & \bfseries0.711$\pm$3e-3 & \bfseries0.0014$\pm$2e-5 & N/A\\
        ~&~& Ours & \bfseries0.746$\pm$6e-3 & \bfseries0.650$\pm$7e-3 & 0.955$\pm$5e-3 & \bfseries0.953$\pm$4e-3 & 0.561$\pm$4e-3 & 0.703$\pm$4e-3 & 0.0018$\pm$9e-5 & 0.955$\pm$9e-3\\
        \bottomrule[1.5pt]
        \multirow{4}{*}{CLEVR-COMPLEX}&\multirow{2}{*}{1}& MulMON & 0.513$\pm$6e-3 & 0.513$\pm$3e-3 &0.934$\pm$3e-3  & 0.931$\pm$2e-3 &\bfseries0.534$\pm$3e-3  & \bfseries0.669$\pm$3e-3 & \bfseries0.0019$\pm$3e-5 & N/A \\ 
        ~ & ~ & Ours & \bfseries0.724$\pm$1e-2 & \bfseries0.633$\pm$9e-3 & \bfseries0.959$\pm$8e-3 & \bfseries0.957$\pm$8e-3 &  0.521$\pm$9e-3 & 0.666$\pm$1e-2 & 0.0020$\pm$2e-4 & 0.943$\pm$2e-2 \\
        \Xcline{2-11}{0.75pt}
        ~&\multirow{2}{*}{2} &MulMON & 0.501$\pm$6e-3 &0.494$\pm$3e-3  &0.915$\pm$3e-3  & 0.902$\pm$2e-3 & \bfseries0.517$\pm$3e-3 & \bfseries0.656$\pm$3e-3 & \bfseries0.0020$\pm$2e-5 &N/A \\
        ~&~& Ours & \bfseries0.710$\pm$9e-3 & \bfseries0.611$\pm$8e-3 & \bfseries0.955$\pm$9e-3 & \bfseries0.948$\pm$8e-3 & 0.506$\pm$7e-3 & 0.655$\pm$8e-3 & 0.0021$\pm$2e-4 & 0.938$\pm$2e-2\\
        \bottomrule[1.5pt]
        \multirow{4}{*}{SHOP-SIMPLE}&\multirow{2}{*}{1}& MulMON & 0.442$\pm$2e-2 & 0.526$\pm$7e-3 & 0.890$\pm$3e-3 & 0.877$\pm$3e-3 & 0.562$\pm$7e-3 & 0.690$\pm$7e-3 & 0.0046$\pm$1e-4 & N/A\\ 
        ~ & ~ & Ours &\bfseries0.767$\pm$2e-3 & \bfseries0.699$\pm$3e-3 & \bfseries0.960$\pm$1e-3 & \bfseries0.960$\pm$2e-3 & \bfseries0.605$\pm$4e-3 & \bfseries0.723$\pm$5e-3 & \bfseries0.0032$\pm$9e-5 & 0.856$\pm$3e-2\\
        \Xcline{2-11}{0.75pt}
        ~&\multirow{2}{*}{2} &MulMON & 0.415$\pm$2e-2 &0.519$\pm$7e-3  & 0.883$\pm$5e-3 & 0.864$\pm$4e-3 & 0.570$\pm$7e-3 & 0.701$\pm$6e-3 & 0.0049$\pm$8e-5 & N/A\\
        ~&~& Ours & \bfseries0.766$\pm$5e-3 & \bfseries0.693$\pm$5e-3 & \bfseries0.960$\pm$2e-3 & \bfseries0.957$\pm$2e-3 & \bfseries0.620$\pm$7e-3 & \bfseries0.744$\pm$8e-3 & \bfseries0.0034$\pm$1e-4 & 0.737$\pm$1e-2\\
        \bottomrule[1.5pt]
        \multirow{4}{*}{SHOP-COMPLEX}&\multirow{2}{*}{1}& MulMON & 0.633$\pm$1e-2 & 0.600$\pm$6e-3 & 0.876$\pm$6e-3 & 0.872$\pm$3e-3 & \bfseries0.606$\pm$6e-3 & \bfseries0.721$\pm$5e-3 & 0.0038$\pm$9e-5 &  N/A \\ 
        ~ & ~ & Ours & \bfseries0.737$\pm$9e-3 & \bfseries0.666$\pm$8e-3  &\bfseries0.932$\pm$6e-3  & \bfseries0.934$\pm$4e-3 &  0.572$\pm$4e-3 & 0.694$\pm$4e-3 & \bfseries0.0034$\pm$2e-4 & 0.855$\pm$3e-2\\
        \Xcline{2-11}{0.75pt}
        ~&\multirow{2}{*}{2} &MulMON &0.636$\pm$1e-2 &0.598$\pm$5e-3  & 0.869$\pm$7e-3 & 0.861$\pm$3e-3 &  \bfseries0.613$\pm$6e-3 & \bfseries0.731$\pm$5e-3 & 0.0040$\pm$6e-5 & N/A\\
        ~&~& Ours & \bfseries0.742$\pm$6e-3 & \bfseries0.663$\pm$6e-3 & \bfseries0.929$\pm$4e-3  &\bfseries 0.928$\pm$3e-3 & 0.588$\pm$6e-3 & 0.716$\pm$6e-3 & \bfseries0.0036$\pm$1e-4  & 0.767$\pm$1e-2\\
        \bottomrule[1.5pt]
        \end{tabular}
    }
    \caption{The comparison results of prediction on test sets (the test mode is 1, the observed views are 8, and query views are 1, 2). All test values are evaluated 5 times, recorded with mean and standard deviation.}\label{tab:predict_m1_o8}
\end{table*}
    
\begin{table*}[ht]
    \renewcommand\arraystretch{1.5}
    \centering
    \scalebox{0.7}{
        \begin{tabular}{c|c|c| cccccccc} 
        \toprule[1.5pt]
        Dataset&Query&Method&ARI-A$\uparrow$&AMI-A$\uparrow$&ARI-O$\uparrow$&AMI-O$\uparrow$&IoU$\uparrow$&F1$\uparrow$&MSE$\downarrow$&OOA$\uparrow$ \\
        \midrule[0.5pt]
        \multirow{4}{*}{CLEVR-SIMPLE}&\multirow{2}{*}{1}& MulMON & 0.681$\pm$1e-3 &  0.624$\pm$1e-3& \bfseries0.968$\pm$2e-3 & \bfseries 0.962$\pm$2e-3 & \bfseries0.630$\pm$3e-3 & \bfseries0.751$\pm$4e-3 & \bfseries0.0014$\pm$3e-5 & N/A \\ 
        ~ & ~ & Ours & \bfseries0.819$\pm$3e-3 & \bfseries0.728$\pm$2e-3 & 0.958$\pm$5e-3 & 0.960$\pm$4e-3 &  0.629$\pm$3e-3 & 0.754$\pm$3e-3 & 0.0017$\pm$8e-5 & 0.960$\pm$2e-2\\
        \Xcline{2-11}{0.75pt}
        ~&\multirow{2}{*}{2} &MulMON & 0.714$\pm$1e-3 & 0.648$\pm$1e-3 & \bfseries0.969$\pm$2e-3 & \bfseries0.960$\pm$1e-3 & \bfseries0.650$\pm$3e-3 & \bfseries0.767$\pm$3e-3 & \bfseries0.0013$\pm$2e-5 & N/A \\
        ~&~& Ours & \bfseries0.834$\pm$4e-3 & \bfseries0.739$\pm$6e-3 & 0.951$\pm$5e-3 &0.951$\pm$5e-3  & 0.641$\pm$7e-3 & 0.763$\pm$6e-3 & 0.0017$\pm$5e-5 & 0.958$\pm$2e-2\\
        \bottomrule[1.5pt]
        \multirow{4}{*}{CLEVR-COMPLEX}&\multirow{2}{*}{1}& MulMON &0.591$\pm$7e-3  & 0.564$\pm$3e-3 & 0.948$\pm$2e-3 & 0.938$\pm$1e-3 & \bfseries0.574$\pm$3e-3 & \bfseries0.703$\pm$3e-3 & \bfseries0.0020$\pm$1e-5  & N/A \\ 
        ~ & ~ & Ours & \bfseries0.769$\pm$1e-2 &\bfseries0.672$\pm$9e-3  &\bfseries0.951$\pm$5e-3  & \bfseries0.951$\pm$5e-3 &  0.542$\pm$7e-3 & 0.678$\pm$8e-3 & 0.0025$\pm$2e-4 & 0.944$\pm$2e-2\\
        \Xcline{2-11}{0.75pt}
        ~&\multirow{2}{*}{2} &MulMON &0.613$\pm$9e-3  & 0.577$\pm$4e-3 & 0.945$\pm$2e-3 &  0.932$\pm$9e-4 & \bfseries0.588$\pm$3e-3 & \bfseries0.716$\pm$3e-3 & \bfseries0.0020$\pm$2e-5 & N/A \\
        ~&~& Ours & \bfseries0.788$\pm$1e-2 & \bfseries0.682$\pm$1e-2 & \bfseries0.950$\pm$6e-3 & \bfseries0.947$\pm$7e-3 & 0.552$\pm$8e-3 & 0.689$\pm$8e-3 & 0.0024$\pm$1e-4 &0.930$\pm$4e-2 \\
        \bottomrule[1.5pt]
        \multirow{4}{*}{SHOP-SIMPLE}&\multirow{2}{*}{1}& MulMON & 0.485$\pm$1e-2 &  0.568$\pm$5e-3 & 0.880$\pm$6e-3 & 0.873$\pm$3e-3 & 0.607$\pm$5e-3 & 0.729$\pm$5e-3 &  0.0050$\pm$1e-4 & N/A \\ 
        ~ & ~ & Ours & \bfseries0.805$\pm$3e-3 & \bfseries0.735$\pm$3e-3 & \bfseries0.961$\pm$5e-3 & \bfseries0.959$\pm$4e-3 & \bfseries 0.656$\pm$2e-3 & \bfseries0.769$\pm$2e-3 & \bfseries0.0035$\pm$6e-5 & 0.866$\pm$3e-2\\
        \Xcline{2-11}{0.75pt}
        ~&\multirow{2}{*}{2} &MulMON & 0.502$\pm$1e-2 &0.581$\pm$5e-3  & 0.871$\pm$6e-3 &  0.862$\pm$3e-3 & 0.626$\pm$5e-3 & 0.745$\pm$5e-3 & 0.0050$\pm$1e-4 & N/A \\
        ~&~& Ours & \bfseries0.828$\pm$2e-3 & \bfseries0.751$\pm$2e-3 & \bfseries0.955$\pm$9e-4 & \bfseries0.951$\pm$2e-3 & \bfseries0.676$\pm$2e-3 &\bfseries 0.786$\pm$3e-3 & \bfseries0.0035$\pm$2e-5 & 0.851$\pm$1e-2\\
        \bottomrule[1.5pt]
        \multirow{4}{*}{SHOP-COMPLEX}&\multirow{2}{*}{1}& MulMON & 0.707$\pm$9e-3 &  0.660$\pm$4e-3 & 0.878$\pm$5e-3 & 0.875$\pm$3e-3 & \bfseries 0.670$\pm$2e-3 & \bfseries 0.776$\pm$2e-3 &  0.0037$\pm$3e-5 &  N/A\\ 
        ~ & ~ & Ours & \bfseries0.779$\pm$8e-3 & \bfseries0.705$\pm$7e-3 & \bfseries0.944$\pm$7e-3 & \bfseries0.941$\pm$7e-3 &  0.623$\pm$8e-3 & 0.739$\pm$8e-3 & \bfseries0.0036$\pm$1e-4 & 0.814$\pm$2e-2\\
        \Xcline{2-11}{0.75pt}
        ~&\multirow{2}{*}{2} &MulMON & 0.730$\pm$9e-3 & 0.672$\pm$4e-3 & 0.860$\pm$6e-3 & 0.858$\pm$3e-3 & \bfseries0.682$\pm$2e-4 &\bfseries0.787$\pm$6e-4  & \bfseries0.0038$\pm$3e-5 & N/A \\
        ~&~& Ours & \bfseries0.791$\pm$7e-3 & \bfseries0.710$\pm$7e-3 & \bfseries0.938$\pm$8e-3 & \bfseries0.932$\pm$7e-3 & 0.639$\pm$8e-3 & 0.756$\pm$9e-3 & \bfseries0.0038$\pm$1e-4 & 0.813$\pm$2e-2 \\
        \bottomrule[1.5pt]
        \end{tabular}
    }
    \caption{The comparison results of prediction on test sets (the test mode is 2, the observed views are 8, and query views are 1, 2). All test values are evaluated 5 times, recorded with mean and standard deviation.}\label{tab:predict_m2_o8}
\end{table*}
    
\begin{table*}[ht]
    \renewcommand\arraystretch{1.5}
    \centering
    \scalebox{0.7}{
        \begin{tabular}{c|c|c| cccccccc} 
        \toprule[1.5pt]
        Dataset&Query&Method&ARI-A$\uparrow$&AMI-A$\uparrow$&ARI-O$\uparrow$&AMI-O$\uparrow$&IoU$\uparrow$&F1$\uparrow$&MSE$\downarrow$&OOA$\uparrow$ \\
        \midrule[0.5pt]
        \multirow{2}{*}{CLEVR-SIMPLE}&\multirow{2}{*}{1}& MulMON &0.588$\pm$1e-3  &  0.549$\pm$1e-3& 0.954$\pm$1e-3 &0.949$\pm$1e-3  & \bfseries0.570$\pm$2e-3 &\bfseries0.704$\pm$2e-3 & \bfseries0.0015$\pm$2e-5 & N/A \\ 
        ~ & ~ & Ours & \bfseries0.727$\pm$8e-3 &\bfseries 0.638$\pm$7e-3 & \bfseries0.960$\pm$6e-3 & \bfseries0.960$\pm$5e-3 &  0.546$\pm$7e-3 & 0.687$\pm$9e-3 & 0.0019$\pm$2e-4 & 0.953$\pm$2e-2 \\
        \bottomrule[1.5pt]
        \multirow{2}{*}{CLEVR-COMPLEX}&\multirow{2}{*}{1}& MulMON & 0.480$\pm$1e-2 & 0.477$\pm$5e-3 & 0.910$\pm$2e-3 & 0.896$\pm$2e-3 & \bfseries0.512$\pm$3e-3 & \bfseries0.654$\pm$3e-3 & \bfseries0.0023$\pm$4e-5 & N/A \\ 
        ~ & ~ & Ours & \bfseries0.681$\pm$7e-3 &\bfseries 0.593$\pm$5e-3 & \bfseries0.951$\pm$4e-3 & \bfseries0.952$\pm$3e-3 &  0.491$\pm$3e-3 & 0.639$\pm$4e-3 & 0.0025$\pm$8e-5 & 0.890$\pm$2e-2 \\
        \bottomrule[1.5pt]
        \multirow{2}{*}{SHOP-SIMPLE}&\multirow{2}{*}{1}& MulMON & 0.413$\pm$6e-3 &  0.516$\pm$2e-3 &0.872$\pm$5e-3  & 0.862$\pm$2e-3 & 0.567$\pm$2e-3 & 0.702$\pm$2e-3 & 0.0058$\pm$1e-4 & N/A \\ 
        ~ & ~ & Ours & \bfseries0.752$\pm$6e-3 & \bfseries0.684$\pm$4e-3 & \bfseries0.956$\pm$2e-3  & \bfseries0.956$\pm$2e-3 & \bfseries0.608$\pm$3e-3 & \bfseries0.733$\pm$3e-3 & \bfseries0.0046$\pm$4e-4 & \bfseries0.823$\pm$2e-2\\
        \bottomrule[1.5pt]
        \multirow{2}{*}{SHOP-COMPLEX}&\multirow{2}{*}{1}& MulMON & 0.648$\pm$5e-3 & 0.609$\pm$3e-3 & 0.869$\pm$8e-3 & 0.866$\pm$4e-3 & \bfseries0.628$\pm$4e-3 & \bfseries0.746$\pm$3e-3 & 0.0042$\pm$5e-5 & N/A \\ 
        ~ & ~ & Ours & \bfseries0.728$\pm$1e-2 & \bfseries0.657$\pm$9e-3 & \bfseries0.930$\pm$7e-3 & \bfseries0.935$\pm$5e-3 &  0.574$\pm$8e-3 & 0.701$\pm$8e-3 & \bfseries0.0040$\pm$1e-4 & 0.811$\pm$2e-2\\
        \bottomrule[1.5pt]
        \end{tabular}
    }
    \caption{The comparison results of prediction on test sets (the test mode is 1, the observed views are 9, and query views are 1). All test values are evaluated 5 times, recorded with mean and standard deviation.}\label{tab:predict_m1_o9}
\end{table*}
    
\begin{table*}[ht]
    \renewcommand\arraystretch{1.5}
    \centering
    \scalebox{0.7}{
        \begin{tabular}{c|c|c| cccccccc} 
        \toprule[1.5pt]
        Dataset&Query&Method&ARI-A$\uparrow$&AMI-A$\uparrow$&ARI-O$\uparrow$&AMI-O$\uparrow$&IoU$\uparrow$&F1$\uparrow$&MSE$\downarrow$&OOA$\uparrow$ \\
        \midrule[0.5pt]
        \multirow{2}{*}{CLEVR-SIMPLE}&\multirow{2}{*}{1}& MulMON & 0.748$\pm$5e-4 &  0.676$\pm$7e-4 & \bfseries0.969$\pm$2e-3 & \bfseries0.962$\pm$1e-3 & \bfseries0.675$\pm$1e-3 & \bfseries0.785$\pm$2e-3 & \bfseries0.0011$\pm$6e-6 & N/A \\ 
        ~ & ~ & Ours & \bfseries0.858$\pm$1e-2 &\bfseries 0.779$\pm$1e-2 & 0.959$\pm$8e-3 & 0.960$\pm$7e-3 &  0.668$\pm$1e-2 & 0.778$\pm$1e-2 & 0.0016$\pm$2e-4 & 0.967$\pm$2e-2\\
        \bottomrule[1.5pt]
        \multirow{2}{*}{CLEVR-COMPLEX}&\multirow{2}{*}{1}& MulMON & 0.640$\pm$8e-3 & 0.597$\pm$4e-3 & 0.944$\pm$2e-3 & 0.936$\pm$2e-3 & \bfseries0.603$\pm$3e-3 & \bfseries0.728$\pm$3e-3 & \bfseries0.0020$\pm$6e-6 & N/A  \\ 
        ~ & ~ & Ours &\bfseries0.817$\pm$4e-3 &\bfseries 0.716$\pm$4e-3 & \bfseries0.963$\pm$4e-3 & \bfseries0.962$\pm$3e-3 &  0.571$\pm$3e-3 & 0.705$\pm$2e-3 & 0.0022$\pm$5e-5 & 0.955$\pm$2e-2 \\
        \bottomrule[1.5pt]
        \multirow{2}{*}{SHOP-SIMPLE}&\multirow{2}{*}{1}& MulMON & 0.551$\pm$2e-2 & 0.615$\pm$9e-3 & 0.870$\pm$4e-3 & 0.871$\pm$2e-3 & 0.648$\pm$5e-3 & 0.763$\pm$5e-3 & 0.0049$\pm$1e-4 & N/A  \\ 
        ~ & ~ & Ours & \bfseries0.848$\pm$7e-3 & \bfseries0.778$\pm$7e-3 & \bfseries0.952$\pm$5e-3 & \bfseries0.951$\pm$4e-3 &  \bfseries0.691$\pm$6e-3 & \bfseries0.795$\pm$5e-3 & \bfseries0.0038$\pm$3e-4 & 0.824$\pm$3e-2 \\
        \bottomrule[1.5pt]
        \multirow{2}{*}{SHOP-COMPLEX}&\multirow{2}{*}{1}& MulMON & 0.755$\pm$4e-3 & 0.694$\pm$1e-3 & 0.847$\pm$7e-3  & 0.853$\pm$4e-3 & \bfseries0.700$\pm$3e-3 & \bfseries0.800$\pm$4e-3 & \bfseries0.0039$\pm$6e-5 & N/A \\ 
        ~ & ~ & Ours & \bfseries0.804$\pm$2e-2 & \bfseries0.733$\pm$1e-2 & \bfseries0.938$\pm$9e-3 & \bfseries0.938$\pm$7e-3 &  0.655$\pm$1e-2 & 0.765$\pm$1e-2 & 0.0041$\pm$4e-4 & 0.812$\pm$7e-2\\
        \bottomrule[1.5pt]
        \end{tabular}
    }
    \caption{The comparison results of prediction on test sets (the test mode is 2, the observed views are 9, and query views are 1). All test values are evaluated 5 times, recorded with mean and standard deviation.}\label{tab:predict_m2_o9}
\end{table*}





\end{document}